\documentclass[runningheads,table]{llncs}

\usepackage{graphicx}
\usepackage{comment}
\usepackage{amsmath,amssymb}
\usepackage{color}
\usepackage{url}
\usepackage{hyperref}
\usepackage{subcaption,float}
\usepackage{xspace}
\usepackage{csquotes}
\usepackage{epsfig}
\usepackage{mwe}
\usepackage{enumitem}
\usepackage{cleveref}
\usepackage{highlight}
\usepackage{tikz}
\usepackage{comment}
\usepackage{amsmath,amssymb} 
\usepackage{xcolor}
\usepackage{enumitem}
\usepackage{csquotes}
\usepackage{booktabs}
\usepackage{hyperref}
\usepackage{subfiles}

\renewcommand{\opacity}{50}

\newcommand{\modelName}{VASTA\xspace}
\newcommand{\sos}{\texttt{[SOS]}\xspace}

\newif\ifreview
 \reviewfalse

\ifreview
	\usepackage{lineno}

	\linenumbers
\fi

\begin{document}


\def\SubNumber{36}


\title{Diverse Video Captioning by Adaptive Spatio-temporal Attention}
%


\ifreview
\else

	\author{Zohreh Ghaderi\inst{1} \and
	Leonard Salewski\inst{1}\and
	Hendrik P.\ A.\ Lensch\inst{1}}
	\authorrunning{Z. Ghaderi et al.}

	\institute{University of Tübingen, Tübingen, Germany \inst{1}\\
	\email{ \{zohreh.ghaderi, leonard.salewski, hendrik.lensch\}@uni-tuebingen.de}}
	
\fi
\maketitle

\begin{abstract}
To generate proper captions for videos, the inference needs to identify relevant concepts and pay attention to the spatial relationships between them as well as to the temporal development in the clip.
Our end-to-end encoder-decoder video captioning framework incorporates two transformer-based architectures, an adapted transformer for a single joint spatio-temporal video analysis as well as a self-attention-based decoder for advanced text generation. 
Furthermore, we introduce an adaptive frame selection scheme to reduce the number of required incoming frames while maintaining the relevant content when training both transformers.
Additionally, we estimate semantic concepts relevant for video captioning by aggregating all ground truth captions of each sample.
Our approach achieves state-of-the-art results on the MSVD, as well as on the large-scale MSR-VTT and the VATEX benchmark datasets considering multiple Natural Language Generation (NLG) metrics. 
Additional evaluations on diversity scores highlight the expressiveness and diversity in the structure of our generated captions.

\end{abstract}
\section{Introduction}
The interplay between visual and text information has recently captivated scientists in the field of computer vision research. Generating a caption for a short video is a simple task for most people, but a tough one for a machine. In particular, video captioning can be seen as a sequence-to-sequence task~\cite{sutskever2014sequence} similar to machine translation. Video captioning frameworks aim to learn a high-level understating of the video and then convert it into text.

Since deep learning has revolutionized almost all computer vision sub-fields, it also plays a notable role in video description generation. Usually, two main Deep Neural Networks are involved, the encoder analyses
visual content and the decoder generates text
~\cite{donahue2015long,sutskever2014sequence,venugopalan2014translating,yao2015describing,venugopalan2015sequence}.
The employed networks often are a variety of 2D-CNN and 3D-CNNs. They extract visual features and local motion information between successive frames.
Furthermore, a Faster RCNN object recognition (FRCNN)~\cite{ren2015faster} can be used to obtain fine-grained spatial information.

Attention mechanisms are adopted to let the model build relations between local and global temporal and spatio-temporal information. 
This information is subsequently fed to a recurrent neural network such as an LSTM or a GRU in order to produce grammatically correct sentences~\cite{aafaq2019spatio,yan2019stat,zhang2020object,pan2020spatio}.
The temporal processing is, however, somehow limited as it either involves global aggregation with no temporal resolution or is based on 3D-CNNs with a rather small temporal footprint. 
Transformer-based encoder-decoder architectures, on the other hand, can inherently establish relations between all components in a sequence independent of their positions~\cite{vaswani2017attention}.
After their breakthrough on language tasks, lately, transformers have been successfully applied to diverse vision applications, mainly for pure classification~\cite{arnab2021vivit,liu2021swin}.

In this work, we present \modelName, an end-to-end encoder-decoder framework for the task of video captioning where transformers perform detailed visual spatio-temporal analysis of the input as well as generating the caption output.

Our encoder architecture is adopted from the Video Swin Transformer~\cite{liu2021video}, which has been shown to be able to interpret non-local temporal dependencies in video-based action recognition.
The task of video captioning, however, requires even more than just spatio-temporal analysis. It needs to extract all semantically relevant concepts~\cite{shekhar2020domain,perez2021improving}, which will be key for the downstream text generation. We identify all relevant concepts in the captions of the training data sets and fine-tune the Swin Transformer to explicitly predict those before handing latent information to the BERT generator.

As end-to-end training of two transformers, in particular for video processing, is quite involved, we introduce an adaptive frame sampling (AFS) strategy that identifies informative keyframes for caption generation.

Our transformer-based encoder-decoder architecture harnesses the power of transformers for both the visual analysis as well as for the language generating part, rendering quite faithful descriptions to a broad range of videos. In summary, our contributions are:
\begin{enumerate}[label=\alph*)]
	\item a simple transformer-based video captioning approach, where a \emph{single} encoder extracts all necessary spatio-temporal information. Unlike other recent works we do not employ disjoint 2D analysis (e.g.\ object detection) and 3D analysis (e.g.\ 3D convolution).
	\item adaptive frame sampling for selecting more informative frames 
	\item visually grounded semantic context vectors derived from all captions of each sample provide high-quality semantic guidance to the decoder
	\item state-of-the-art results on three datasets (MSVD, MSR-VTT and VATEX)
    \item significantly increased diversity in the predicted captions.
\end{enumerate}

\section{Related Work}
Existing video captioning approaches can be grouped according to the techniques used for visual analysis and text generation. See Aafaq et al.\ \cite{aafaq2019survey} for a detailed survey.

\subsection{Classical Models}

Classical models mainly concentrate on detecting objects, actors and events in order to fill the SVO (SVOP) structure~\cite{thomason2014integrating}. Detection of objects and humans was accomplished by model-based shape matching including HOG, HOF and MbH,~\cite{dalal2006human,wang2009evaluation,wang2009hog}. 
The analysis part is typically weak on interpreting dynamics and the output of these models is limited due to their explicit sentence structures~\cite{kojima2002natural,khan2011human,lee2008save,guadarrama2013youtube2text,krishnamoorthy2013generating}. Classical models have recently been outperformed by models based on deep learning. 

\subsubsection{Spatio-temporal Analysis with CNNs}
On the encoder side, a more fine-grained analysis of the spatio-temporal aspects has been enabled with the advent of 3D-convolutions and the corresponding C3D network~\cite{tran2015learning}.
Li et al.~\cite{yao2015describing} present a 3D-CNN network for fine local motion analysis and employ soft-attention~\cite{bahdanau2014neural} for adaptive aggregation to obtain a global feature for video captioning. 
Methods like~\cite{aafaq2019spatio,cherian2020spatio,wang2019vatex,singh2020nits} combine both 2D and 3D-CNNs with attention mechanisms to obtain stronger spatial-temporal and global features.
As the video captions often reflect some temporal relation between specific objects, the use of explicit object detectors~\cite{pan2020spatio,zhang2019object,zheng2020syntax,zhang2020object} can improve the generated descriptions. 
Recently, the work on MGCMP~\cite{chen2021motion} and CoSB~\cite{vaidya2022co} illustrates that extracting fine-grained spatial information followed by propagation across time frames could provide visual features as good as other methods that use external object detector features as long as their relation is adequately realized in the temporal domain. 
While temporally resolved features are necessary to analyse the dynamics, 
global aggregates can provide the proper semantic context for generating captions.
In~\cite{shekhar2020domain,perez2021improving,chen2020semantics}, semantic attributes are learned inside a CNN-RNN framework. 
In contrast to the work of Gan et al.~\cite{gan2017semantic} 
we condition our decoder on the semantic context vector once instead of feeding it to the decoder at every step.
Still, the self-attention operation of our decoder allows it to be accessed whenever it is needed.
It has been shown that selecting informative frames aids in video action recognition~\cite{gowda2020smart} as this reduces the overall processing cost and lets the system focus on the relevant parts only. The frame selection could be trained to optimize the input for the downstream task as in \cite{chen2018less,gowda2020smart} but this would introduce further complexity to controlling the entire pipeline.
In contrast to \cite{chen2018less,gowda2020smart} our method is simpler and does not require to be learned.


\subsubsection{Transformer-based Models}
Following the success of transformers~\cite{vaswani2017attention} in text-related sequence-to-sequence tasks like translation, they have recently also been applied to vision tasks and in particular to video classification. 
A key ingredient of transformers is the multi-head self-attention mechanism where each head individually can attend and combine different parts of the sequence. This way, a transformer can explore both long-term and short-term dependencies in the same operation. 
For the task of action recognition, the ViViT transformer~\cite{arnab2021vivit} chops the video cube into separate spatio-temporal blocks, applying multi-head self-attention. To keep the complexity at bay, factorized attention alternates between relating temporal or spatially-aligned blocks. 
The video Swin Transformer~\cite{liu2021video} overcomes the problem of hard partition boundaries by shifting the block boundaries by half a block in every other attention layer. 
Instead for action recognition, we use the Swin transformer for video captioning.
On a different task, Zoha et al.\ \cite{zhou2018end} employ transformer in video dense captioning with long sequences and multiple events to be described. 
The concept of cross-attention can easily fuse the information coming in from different feature extractors. 
TVT~\cite{chen2018tvt} uses a transformer instead of a CNN-RNN network for the video captioning task. They use attentive-fusion blocks to integrate image and motion information.
The sparse boundary-aware transformer method~\cite{jin2020sbat} explicitly performs cross-attention between the image domain and extracted motion features. In addition, a specific scoring scheme tries to tune the multi-head attention to ignore redundant information in subsequent frames.
Unlike these works we do not require special fusion, as we use a single joint encoder of the input video.

\section{Model Architecture}

\begin{figure}[t]
\centerline{
\includegraphics[width=\linewidth]{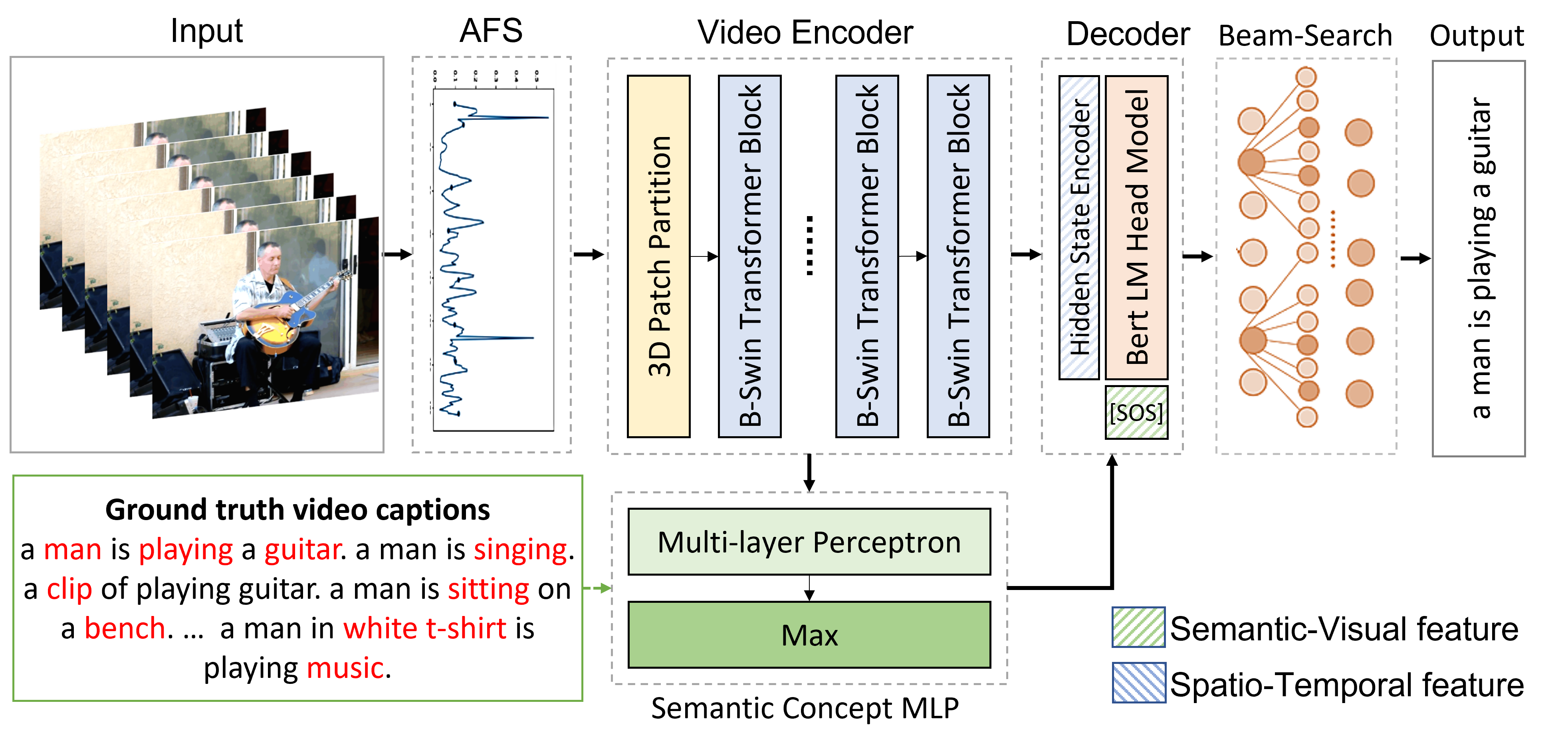}}
 \vspace{-0.3cm}
  \caption{
    \modelName (diverse Video captioning by Adaptive Spatio-Temporal Attention) The most informative 32 frames are selected by adaptive frame sampling (AFS) as input to a Swin transformer~\cite{liu2021video} for spatio-temporal video analysis. The output tokens of the encoder are used twice, once, to predict a semantic concept vector aggregating the entire sequence as start-of-sequence token for the BERT decoder, and, second, for cross-attention in the decoder. A beam search on the most likely words predicts the final caption. }
\label{fig:model_figure}
\end{figure}

The composed architecture of our \modelName model, visualized in~\Cref{fig:model_figure}, is based on an encoder-decoder transformer~\cite{vaswani2017attention}.
First, an adaptive selection method is used to find informative frames in the whole video length. Thereafter, the model encodes the selected video frames into a contextualised but temporally resolved embedding.
We modify the Swin Transformer block~\cite{liu2021video}, which was originally designed for action recognition and classification tasks, to interpret the input video. 
The last hidden layer of this encoder is passed to the decoder. Though compressed, the output of the encoder still contains a temporal sequence. This allows the BERT~\cite{devlin2018bert} decoder to cross-attend to that sequence when generating the output. 
Besides the direct encoder-decoder connection, the encoder output is further used to predict a globally aggregated semantic context vector that is to condition the language generator. 

\subsection{Adaptive Frame Selection}%
\label{sec:adaptive_frame_selection}
In videos, not all frames contribute the same information to the final caption. Some frames are rather similar to the previous ones while some contain dynamics or show new objects. In most video captioning approaches, frames are selected with fixed uniform intervals~\cite{cherian2020spatio,tran2015learning,zhang2020object}.

\begin{figure}[t]
    \centering
    \includegraphics[width=.65\linewidth]{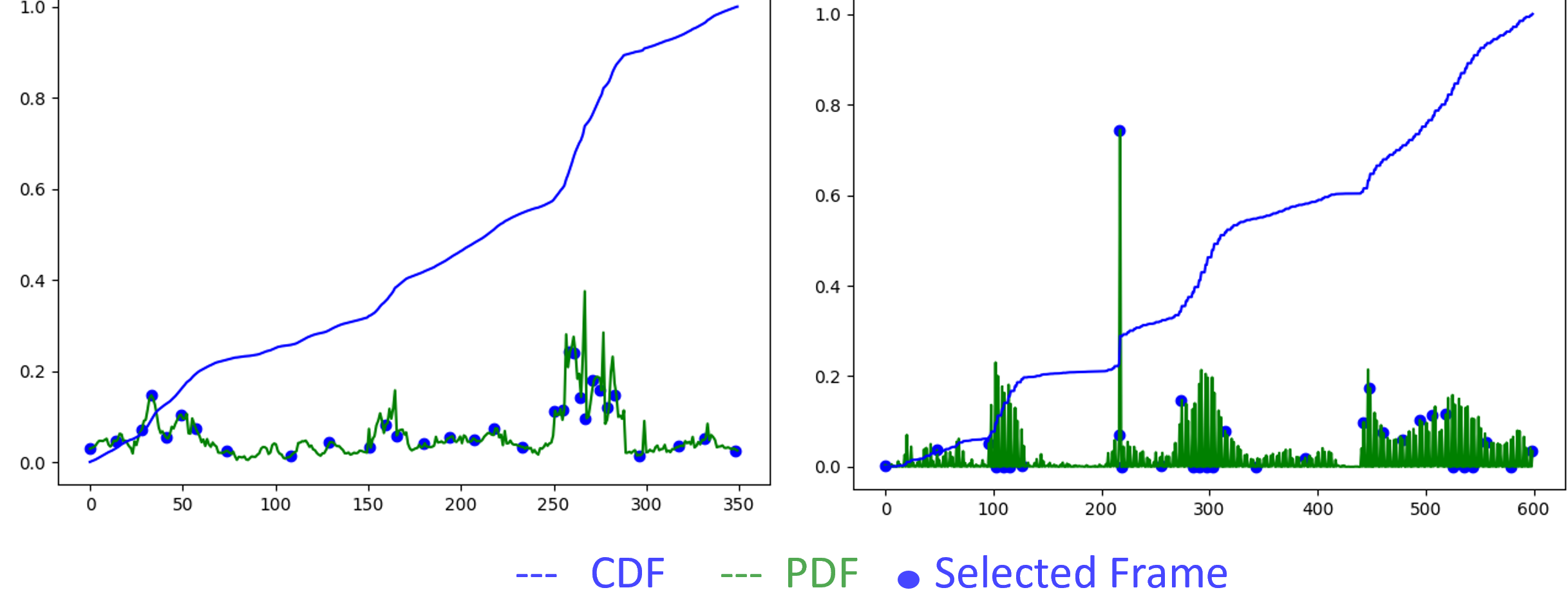}
     \vspace{-0.3cm}
    \caption{Adaptive frame selection visualized for two videos from the MSR-VTT. x-axis: video length, y-axis: PDF and CDF drived from LPIPS sampling.}%
    \label{fig:adaptive_frame_selection}
\end{figure}

Our adaptive frame selection (AFS) performs importance sampling on the input frames based on local frame similarity. First, the similarity for each pair of two consecutive frames is computed by the LPIPS score~\cite{zhang2018perceptual}.
As indicated in Figure~\ref{fig:adaptive_frame_selection}, we consider this similarity as a probability density function (PDF) $f$, normalizing it over all frames. Computing and inverting the cumulative density function (CDF) $F$ with 

\begin{equation}
F(x)=P(X\leq x)= \int_{-\infty}^{x} f(t) \mathrm{d}t,\text{ for }x \in \mathbb{R}
\end{equation}

\noindent one can sample $N$ frames $i$ according to $f$ starting with a uniform distribution $
j= \left\{ 0, \, \frac{1}{N},\,\frac{2}{N}, \, ... \,,\,\frac{N-1}{N} \right\}\label{eq:indexframe},  \;\; i = round( F^{-1}(j) )$.

We select the frame $i$ by rounding to the nearest integer.  
The resulting sequence forms the input to the 3D patch partition of the encoder.

\subsection{Encoder}
The Swin architecture is a hierarchical transformer that is able to act as general-purpose backbone in computer vision tasks~\cite{liu2021swin}. The vision input is initially split into non-overlapping patches, followed by \textbf{S}hifting these \textbf{Win}dows by half the patch size in every other self-attention step to avoid artifacts from the discrete partitioning. Consequently, the swin video transformer~\cite{liu2021video} operates on shifted 3D partitions to cover both the spatial and the temporal domain. The encoder backbone is the Swin-B variant with last the hidden layer size $(BS \times 16 \times 1024)$. 

\subsection{Semantic Concept Network}
The second step of our pipeline includes predicting a semantic concept vector of the entire video that is used as the start-of-sequence token in the decoder to condition its generation on it.
The training signal for this concept vector aggregates \emph{all} ground truth captions of a video to provide a high-quality signal for the caption generation. 
The concepts are defined by the $K$ most frequent words found in the captions of the entire data set.
Predicting the semantic concept vector of a video is learned as a binary multi-class classification task indicating which of the frequent words are relevant to describe the video. 
For generating the ground truth classification vectors, we first select nouns, verbs and adverbs\footnote{categorizing and POS tagging using NLTK (\url{https://www.nltk.org/})} from all captions of the training videos (see Figure~\ref{fig:model_figure}). 
Each video is labeled with the $K$-dimensional vector $L$ by:

$$
L_{k}=\begin{cases}
			1, & \text{if word }k \text{ occurs in any caption}\\
            0, & \text{otherwise}
		 \end{cases}
$$

An MLP acts separately with shared weights on each of the encoder outputs. 
A single max-pooling layer merges them to one token. 
Afterwards a two-layer MLP with RELU activation predicts the $K$ concepts. 
For training, the binary cross-entropy loss is minimized. In essence, the probability of each word in the concept dictionary is used as a semantic feature.
Introducing the semantic concept vector provides an aggregated constant signal for each video while the decoder is trained by generating individual captions. 

\subsection{Decoder}
Our decoder generates the language output word by word while self-attending to all already generated tokens. During training, masking ensures that BERT cannot access the future tokens while during inference it is auto-regressive. 

In our architecture, we pass the semantic feature vector as start-of-sequence token to the decoder for generating the first output word. The self-attention to the first input token conditions the predicted caption on the semantic concepts which have been identified by the semantics MLP.
In order to couple the hidden states of the decoder to the output of the encoder, cross-modal fusion is necessary. The necessary functionality is incorporated by extending the decoder with multi-head cross-attention~\cite{vaswani2017attention}. In $13$ layers, the architecture alternates between multi-head self-attention on the language tokens, cross-attention between the language tokens and the Swin output tokens, and a feed-forward layer followed by normalization. All these steps are bridged by a residual connection. 
A final linear layer projects the decoder internal hidden states to the size of the BERT vocabulary, followed by a softmax to produce word probabilities for each token. The sentence is finally generated by applying beam search~\cite{reddy1977speech} to obtain the most likely combination of words.

\section{Experiments}
To show the effectiveness of our proposed architecture, we train our model on three common video captioning data sets and achieve high-ranking results.
Details on the architecture, training and optimizer~\cite{loshchilov2017decoupled} settings are described in the supplementary.

\subsection{Datasets and Metrics}
Our model is trained on MSR-VTT~\cite{xu2016msr}, MSVD~\cite{chen2011collecting} and VATEX~\cite{wang2019vatex}.
\textbf{MSVD~\cite{chen2011collecting}} includes 1970 videos with an average length of 10 seconds. It provides up to 45 captions per video which are randomly sampled during training.  
\textbf{MSR-VTT~\cite{xu2016msr}}
contains a wide variety of open domain videos of 20 seconds average and 20 captions per video. 
\textbf{VATEX~\cite{wang2019vatex}}
is a large-scale video description dataset.
It is lexically richer, as each video has 10 unique sentences and every caption is unique in the whole corpus.
%

\noindent \textbf{Similarity Metrics} All captions in all datasets have been annotated by humans. We utilize the MS COCO Caption Evaluation~\cite{chen2015microsoft} protocol on both datasets and evaluate standard natural language generation metrics (NLG) as done by previous works in this field. These metrics include BLEU(4-gram)(B4)~\cite{papineni2002bleu}, METEOR(M)~\cite{banerjee2005meteor}, CIDEr(C)~\cite{vedantam2015cider} and ROUGE-L(R)~\cite{lin2004rouge}.
Additionally, we evaluate on BERTScore~\cite{bert-score}, a more modern evaluation metric, that has shown great alignment with human judgement. Notably, it does not share the same brittleness as the $n$-gram based metrics~\cite{papineni2002bleu}.

\noindent \textbf{Diversity Metrics} In contrast to previous work we further measure the diversity of the generated captions.
Zhu et al.\ \cite{zhu2018texygen} introduced Self-BLEU (SB) to evaluate the diversity of generated sentences.
It measures the BLEU value of each prediction wrt.\ the remaining predictions and averages the obtained BLEU values.
A lower value indicates higher diversity, as on average a prediction is less similar to all other predictions.
Furthermore, Dai et al.\ \cite{dai2018neural} proposed the concepts of Novel Captions (N), Unique Caption (U) and Vocab Usage (V) to evaluate diversity of the generated caption.
Novel Caption shows the percentage of generated captions which have not been seen in the training data;
Unique Caption denotes the percentage of distinct captions among all generated captions;
Vocab Usage indicates the percentage of words that are used to generate captions from the whole vocabulary.

\subsection{Quantitative Results}
We present quantitative results in~\Cref{tab:results-table,tab:results-table_vatex} and highlight that besides explanation quality also explanation diversity is important (\Cref{tab:diversity_result}). We ablate the components of our model in~\Cref{tab:ablations_afs} and discuss qualitative examples in~\Cref{example_result}.

\subsubsection{Comparison to Related Approaches}
On the very large VATEX data set our generated captions show significant performance improvements on all scores (see~\Cref{tab:results-table_vatex}). Similarly, on MSR-VTT and the even smaller MSVD we obtain high-ranking, most often top-scoring results with slightly less improvements (see \Cref{tab:results-table}). This indicates that fine tuning of the encoder and decoder transformers benefits from the additional training data. 

\begin{table}[t]
    \centering
    {
        \begin{tabular}{rcccccccc}
            \toprule
            Method  & Model & Year      & B4 $\uparrow$    & M $\uparrow$                  & C $\uparrow$         & R $\uparrow$   & BERT-S & $\uparrow$      \\
            \midrule
            Shared Enc-Dec~\cite{wang2019vatex} &A&2019& \gradientvb{28.4}& \gradientvm{21.7}&\gradientvc{45.1}&\gradientvr{47.0} & - \\
            NITS-VC~\cite{singh2020nits} &A&2020& \gradientvb{20.0}& \gradientvm{18.0}&\gradientvc{24.0}&\gradientvr{42.0} & - \\
            ORG-TRL~\cite{zhang2020object}     & A &2020  & \gradientvb{32.1}&\gradientvm{22.2}&\gradientvc{49.7}&\gradientvr{48.9} & - \\
            \modelName (Kinetics-backbone)      & T &2022&   \gradientvb{36.25} & \gradientvm{25.32} & \gradientvc{65.07} & \gradientvr{51.88} & \cellcolor{high!\tempa!low!\opacity}90.76 &  \\
            \bottomrule
        \end{tabular}
    }  
    \caption{Natural Language Generation (NLG) and BERT scores for VATEX.}
    \label{tab:results-table_vatex}
\end{table}
Thus, instead of just fine-tuning the full pipeline starting with the backbone trained on Kinetics~\cite{kay2017kinetics} for each individual data set, we trained once end-to-end on Vatex and then fine-tuned for MSVD and MSR-VTT. Through this transfer learning \modelName improves in general, particularly the CIDEr score  on MSR-VTT and MSVD by a big margin. 
The performance on METEOR and CIDEr is relevant as both consider semantic relatedness. METEOR excepts synonyms and it exhibits a higher correlation with human judgment on captions and explanations~\cite{kayser2021vil}. The CIDEr score has been particularly designed for measuring the quality of descriptions for visual content. While the NLG scores are all based on $n$-grams the BERTScore is more semantically robust and agrees even better with human assessments. 
On both data sets, our model achieves the highest BERTScore. 

\begin{table*}[]
    \centering
    \resizebox{\linewidth}{!}{
        \begin{tabular}{rcccccccccccc}
            \toprule
            & & & \multicolumn{5}{c}{MSR-VTT} & \multicolumn{5}{c}{MSVD}\\
            \cmidrule(r){4-8}\cmidrule(r){9-13}
            Method  & Model & Year      & B4     & M                   & C          & R     & BERT-S   & B4        & M        & C          & R   & BERT-S    \\
            \midrule
            Att-TVT~\cite{chen2018tvt}         & T&2018 &  \gradient{40.12}                       & \gradientmm{27.86}                    & \gradientmc{47.72}          & \gradientmr{59.63}    & -  & \gradientdb{53.21}          & \gradientdm{35.23}         & \gradientdc{86.76}   &-                  \\
            GRU-EVE~\cite{aafaq2019spatio}     & A &2019  & \gradient{38.3}                        & \gradientmm{28.4}                     & \gradientmc{48.1}           & \gradientmr{60.7}     & -    & \gradientdb{47.9}          & \gradientdm{35.0}          & \gradientdc{78.1}           & \gradientdr{71.5} &-         \\
            OA-BTG ~\cite{zhang2019object} & A &2019& \gradient{41.4} & \gradientmm{28.2}& \gradientmc{46.9}& -&-& &\gradientdb{36.9}  &\gradientdm{36.2}&\gradientdc{90.6}&-  \\
            STG~\cite{pan2020spatio}           & A &2020  & \gradient{40.5}                        & \gradientmm{28.3}                     & \gradientmc{47.10}          & \gradientmr{60.9}        & - & \gradientdb{52.2}          & \gradientdm{36.9}          & \gradientdc{93.0}           & \gradientdr{73.9}   &-       \\
            STATS~\cite{cherian2020spatio}     & A &2020  & \gradient{40.1}                        & \gradientmm{27.5}                     & \gradientmc{43.4}           & \gradientmr{60.4}          & -&\gradientdb{52.6}          & \gradientdm{33.5}          & \gradientdc{80.2}           & \gradientdr{69.5} &-        \\
            SAAT~\cite{zheng2020syntax}        & A &2020  & \gradient{40.5}                        & \gradientmm{28.2}                     & \gradientmc{49.1}         &   \gradientmr{60.9}          &\gradientmbs{82.50}&\gradientdb{46.5}          & \gradientdm{33.5}          & \gradientdc{81.0}           & \gradientdr{69.4}& -           \\
            ORG-TRL~\cite{zhang2020object}     & A &2020  & \gradient{43.6}                        & \gradientmm{28.8}                     & \gradientmc{50.9}           & \gradientmr{62.1}          & -&\gradientdb{54.3}          & \gradientdm{36.4}          & \gradientdc{95.2}           & \gradientdr{73.9} & - \\
            SAVCSS~\cite{chen2020semantics}    & A &2020  & \gradient{43.8}                        & \gradientmm{28.9}                     & \gradientmc{51.4}           & \gradientmr{62.4}         & \gradientmbs{90.00}&\gradientdb{61.8}          & \gradientdm{37.8}          & \gradientdc{103}            & \gradientdr{76.8}    &\gradientdbs{91.25}      \\
            DSD-3 DS-SEM~\cite{shekhar2020domain}    & A&2020   & \gradient{45.2}                        & \gradientmm{29.9}                     & \gradientmc{51.1}           & \gradientmr{64.2}          & -&\gradientdb{50.1}          & \gradientdm{34.7}          & \gradientdc{76}            & \gradientdr{73.1} &-         \\
            SBAT~\cite{jin2020sbat} &T& 2020& \gradient{42.9} & \gradientmm{28.9}& \gradientmc{51.6}& \gradientmr{61.5} &-&\gradientdb{53.1} &\gradientdm{35.3} & \gradientdc{89.5}& \gradientdr{72.3} &-
            \\
            SemSynAN~\cite{perez2021improving}$\dagger$ & A & 2021 & \gradient{46.4}               & \gradientmm{30.4}            & \gradientmc{51.9}           & \gradientmr{64.7}&  \gradientmbs{82.13} & \gradientdb{64.4} & \gradientdm{41.9} & \gradientdc{111.5} & \gradientdr{79.5}  & \gradientdbs{82.67}\\
        
            MGCMP~\cite{chen2021motion}& A&2021& \gradient{41.7} & \gradientmm{28.9} & \gradientmc{51.4}& \gradientmr{62.1} &- &\gradientdb{55.8} & \gradientdm{36.9} & \gradientdc{98.5} & \gradientdr{74.5} &- \\
            
            CoSB~\cite{vaidya2022co}& T &2022& \gradient{41.4} & \gradientmm{27.8} & \gradientmc{46.5}& \gradientmr{61.0} & -&\gradientdb{50.7} & \gradientdm{35.3} & \gradientdc{97.8} & \gradientdr{72.1} &- \\
            \bottomrule
            \modelName (Kinetics-backbone)      & T &2022&   \gradient{43.4}     &  \gradientmm{30.2} & \gradientmc{55.0}          & \gradientmr{62.5} & \gradientmbs{90.10} &\gradientdb{56.1}& \gradientdm{39.1} & \gradientdc{106.4} &\gradientdr{74.5} & \gradientdbs{92.00}          \\ 

            \modelName (Vatex-backbone)      & T &2022&   \gradient{44.21}     & \gradientmm{30.24} & \gradientmc{56.08}         & \gradientmr{62.9} & \gradientmbs{90.17} &\gradientdb{59.2}& \gradientdm{40.65} & \gradientdc{119.7} &\gradientdr{76.7} & \gradientdbs{92.21}\\
            
            \bottomrule
        \end{tabular}
    }
    \caption{Natural Language Generation (NLG) and BERT scores for the MSR-VTT and MSVD datasets (T: Transformer, A: Attention). Darker blue indicates higher scores. For both data sets our approach improves the BERTScore and produces high-ranking NLG scores. $\dagger$: BERT-score is computed on reproduced captions by the released code.} %
    \label{tab:results-table}
\end{table*}


\subsubsection{Caption Diversity}
Albeit their wide-spread use, NLG metrics only assess a few aspects of the generated captions. 
Maximising the scores on existing NLG metrics as presented in~\Cref{tab:results-table} can for example be achieved with focusing on the most prevalent sentence structure found in the ground truth captions.
However, we are interested in captions that are the most \enquote{human-like}.
Thus, we compute these diversity metrics on MSR-VTT, MSVD and VATEX and compare our model to those competitors where we have access to the generated captions for re-evaluation.
As seen in \Cref{tab:diversity_result}, our model not only predicts highly accurate captions but also manages to predict a highly diverse set of captions.
\modelName generates the most distinct captions 
and does not overfit to the training data, i.e.\ generates novel captions for some of the test videos.
Our model by far exploits most of the training vocabulary. 
Further analysis on the diversity of sentence structures is given in the supplementary.

Analyzing the performance of SemSynAN~\cite{perez2021improving} (a model which has strong similarity metrics) where the number of captions per video is limited to just five to train a predictor for the most common syntactic POS structures (see Supplementary) reveals that 
\begin{table*}[tb]
    \resizebox{\linewidth}{!}{
        \begin{tabular}{rcccccccccccc}
            \toprule
            &\multicolumn{4}{c}{MSR-VTT} & \multicolumn{4}{c}{MSVD}   & \multicolumn{4}{c}{VATEX}\\
            \cmidrule(r){2-5}\cmidrule(r){6-9}\cmidrule(r){10-13}
            Method                                   &  SB $\downarrow$    & N $\uparrow$             &  U $\uparrow$                & V $\uparrow$           & SB $\downarrow$ & N $\uparrow$     & U $\uparrow$      & V $\uparrow$  & SB $\downarrow$ & N $\uparrow$     & U $\uparrow$      & V $\uparrow$      
            \\
            SAVCSS~\cite{chen2020semantics}  & 95.19 & 44.61                       & 33.44                   & 1.88               &   84.32& \textbf{51.34}          & 42.08        &  2.07 & - & - & - & -     \\
             SAAT~\cite{zheng2020syntax}        & 99.99 & 40.46                       & 20.06   & 1.33  &- & - & - &-     & - & - & - & - \\
            SemSynAN~\cite{perez2021improving}    & 96.47     & 42.84                       & 18.92                   & 1.57  &83.00&47.16 & 37.61 & 2.19 & - & - & - & - \\
            \midrule
            \modelName (Kinetics-backbone)  & 92.94 & \textbf{45.98}                       & \textbf{34.74}                   &2.93 & 81.88&30.49 & 42.89 & 3.48& 86.18 & 97.29 & 85.80 & 7.04  \\
            \modelName (Vatex-backbone)  &\textbf{92.70}& 45.51                       & 34.21                   &\textbf{ 3.00} & \textbf{76.90}&42.75 & \textbf{52.16} & \textbf{3.94} & - & - & - & -  \\
            \bottomrule
   
        \end{tabular}
    }
    \caption{Diversity of the generated captions.}
    \label{tab:diversity_result}
\end{table*}
this comes at the cost of reduced caption diversity, sentence quality and video-caption match. Thus, we found that its diversity is much lower.


\begin{table*}[tb]
    \resizebox{\linewidth}{!}{
        \begin{tabular}{rcccccccccccccccc}
            \toprule
                          &    \multicolumn{8}{c}{MSR-VTT} &  \multicolumn{8}{c}{VATEX}                        
                     \\
            \cmidrule(r){2-9}\cmidrule(r){10-17}

            Method   & B4 $\uparrow$                      & M $\uparrow$                   & C $\uparrow$          & R $\uparrow$    &  SB $\downarrow$                      & N $\uparrow$                   & U $\uparrow$         & V $\uparrow$ & B4 $\uparrow$                      & M $\uparrow$                   & C $\uparrow$          & R $\uparrow$   &   SB $\downarrow$                      & N $\uparrow$                   & U $\uparrow$         & V $\uparrow$    \\
            \midrule
             UFS-SB & 43.21                       & 29.55                    & 52.91& 62.1 & 96.48                       & 37.09                    & 19.23 & 1.90  &35.31&25.05&63.82&51.27&87.73 & \textbf{97.62} & 81.41 & 6.70 
            \\
            AFS-SB          & 43.07                       & 29.72                    & \textbf{55.08} & 62.02 & 93.93 & 38.29 & 27.95 & 2.46 & 35.64& \textbf{25.43}&64.98&51.53&88.57 & 97.51 & 83.33 & 6.50
            \\
            UFS-SBS  & \textbf{43.51}                       & 29.75                    & 53.59 & 62.27 &94.82 & 42.44 & 26.48 & 2.36&35.68&25.42&\textbf{65.63}&51.58&88.40 & 97.35 & 83.38 & 6.43 
            \\AFS-SBS & 43.43                       & \textbf{30.24}                    & 55.00          & \textbf{62.54}   &   \textbf{92.94} & \textbf{45.98} & \textbf{34.74} & \textbf{2.93}    &\textbf{36.25}&25.32&65.04&\textbf{51.88}  &\textbf{86.18} & 97.29 & \textbf{85.80} & \textbf{7.04}    \\

          \bottomrule
        \end{tabular}
    }
    \caption{Influence of the individual components in \modelName. Applying both, AFS and semantic vectors, yields favorable scores. \small{UFS: uniform frame selection, AFS: Adaptive frame selection, SB: Swin BERT, SBS: Swin BERT Semantics.} }%
    \label{tab:ablations_afs}
\end{table*}

\subsection{Ablation Study}
The results of \Cref{tab:results-table} have been achieved by carefully designing our adaptive spatio-temporal encoder-decoder framework. 
As demonstrated by the ablation results in \Cref{tab:ablations_afs}, introducing adaptive frame sampling (AFS) helps improve the image-description related CIDEr score while adding the semantic concept vector further improves on the more translation-related scores (BLEU-4, METEOR, ROUGE-L, CIDEr).
Similarly, both AFS and the semantic concept prediction improves the diversity score. Thus, more informative and more precise encoder predictions support higher quality and more diverse language output. In the supplemental we demonstrate how the results depend on the chosen decoder model, by replacing Bert by GPTNeo Causal LM~\cite{black2021gpt,radford2018improving}. There, we also study different inference methods (beam search, greedy, top-k, top-p).



\begin{figure*}[ht]
    \centering

    {\centering 
    \includegraphics[width=\linewidth]{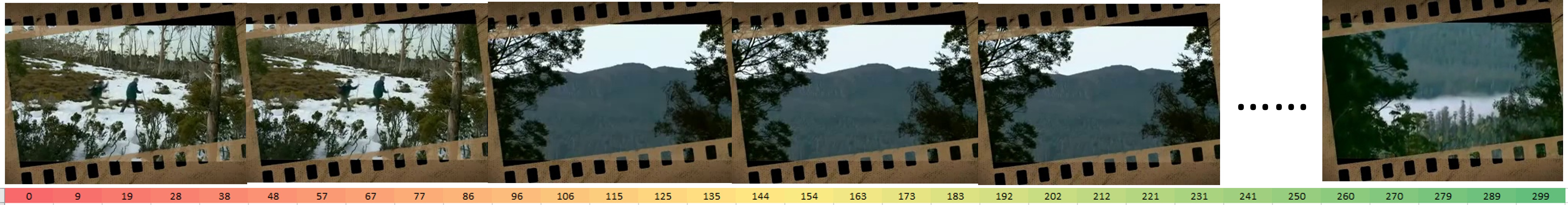} 
    }
    \scriptsize{ 
        \vspace{-.4cm}
        \textbf{Uniform sampling}: there is a car moving on the road} \\
     {\centering 
     \vspace{0.4cm}
    \includegraphics[width=\linewidth]{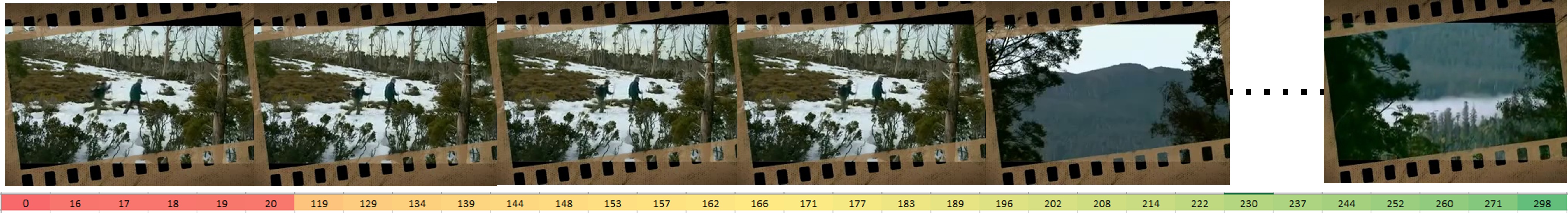} 
    }
    \scriptsize{ 
          \vspace{-.4cm}
        \textbf{Adaptive sampling}: a man is walking through the woods}\\
        
    \vspace{1ex}
    \caption{Adaptive Frame Selection (AFS). Uniform sampling (top) keeps frames with repetitive non-informative content (cf.\ frames with foliage).
    In contrast, adaptive sampling enhances the diversity of input frames by selecting those with activity (cf.\ frames with people walking). Ground truth: \enquote{Two men walking around a forest by a lake.}
    }%
    \label{fig:AFSEffect}
\end{figure*}

\subsubsection{AFS Results}
Figure~\ref{fig:AFSEffect} exemplifies the effect of our adaptive frame selection approach. The video transformer can only take in 32 frames.
Driven by the frame differences, the adaptive frame selection samples more diverse frames than simple uniform subsampling, increasing the chance of selecting informative time steps. 
An irregular temporal sampling pattern on the other hand leads to a non-uniform play-back speed.  
Still, AFS consistently improves the CIDEr result (\Cref{tab:ablations_afs}), indicating that the gain in input information has a more positive effect than potential introducing temporal disturbance in the Swin transformer inference.

\subsubsection{Semantic Concept Vectors}
The accuracy (BCE score) of the multi-class extraction task for the semantic concept vectors is 0.88 (0.12) on the training set and 0.85 (0.15) on the test set. This indicates, that this training step generalizes well. Introducing the semantic concept vectors improves the overall performance, as one can see by the strong correlation between classification accuracy and resulting NLG scores in \Cref{tab:ablations_sman}.
Bad examples most often occur in conjunction with misclassification of the main actors or concepts in the video. 
In these cases, often the content of the video is not well represented by the most common $768$ concepts.

\begin{table*}[tb]
\centering
        \begin{tabular}{rcccc}
  \toprule
      BCE                                  & B4 $\uparrow$                      & M $\uparrow$                  & C $\uparrow$          & R $\uparrow$   
                  \\
            \midrule
        $10\%$-best & 55.67 & 41.25& 109.4& 75.44      \\
        
         $10\%$-worst & 31.50 & 21.97& 22.76& 49.09
        \\
          \bottomrule
        \end{tabular}
    \caption{Dependency on the quality of the predicted semantic vector. Sorting all test samples of the MSR-VTT wrt.\ the classification accuracy of the proposed semantic vector, a strong correlation with the evaluation scores is revealed.}%
    \label{tab:ablations_sman}
\end{table*}



\begin{figure*}[ht]
  { \centering
    \parbox{.90\linewidth}{
          \parbox{0.25\linewidth}{
    \includegraphics[width=\linewidth]{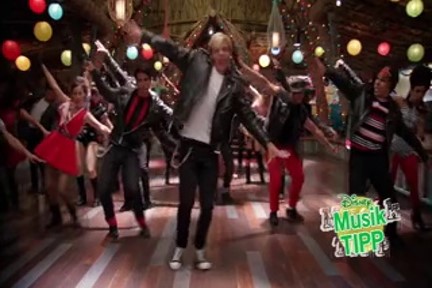}} 
      \parbox{0.65\linewidth}{ 
      \scriptsize{\textbf{Reference:}
      a group of people are dancing and singing}\\
      \scriptsize{\textbf{Our:}
      a group of people are dancing and singing}\\
       \scriptsize{\textbf{SymsynAN:}
      a group of people are dancing}}}
      \\
      \parbox{.90\linewidth}{
    \parbox{0.25\linewidth}{
    \includegraphics[width=\linewidth]{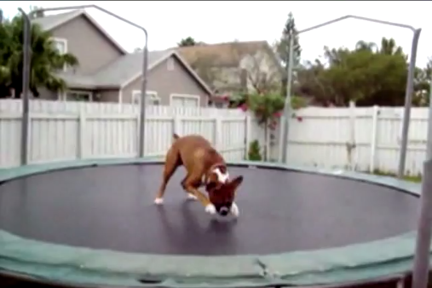} }
      \parbox{0.65\linewidth}{ 
      \scriptsize{\textbf{Reference:}
      a dog is playing on a trampoline}\\
      \scriptsize{\textbf{Our:}
     a dog is playing on a trampoline}\\
       \scriptsize{\textbf{SymsynAN:}
     a dog is playing with a dog}}}
   \\
    \parbox{.90\linewidth}{
        \parbox{0.25\linewidth}{
    \includegraphics[width=\linewidth]{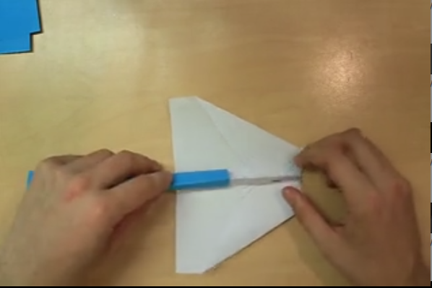}} 
      \parbox{0.65\linewidth}{ 
      \scriptsize{\textbf{Reference:}
      a person is making a paper aircraft}\\
      \scriptsize{\textbf{Our:}
     a person is making a paper airplane}\\
       \scriptsize{\textbf{SymsynAN:}
     a person is folding paper}}}
    
   }
 \vspace{-0.3cm}




    \caption{Examples for the top-performing videos in the test set.
    }%
    \label{fig:resultImages}
\end{figure*}

\subsection{Qualitative Results}\label{example_result}
\label{sec:exampleResults}
In Figure~\ref{fig:resultImages}, representative videos and the generated captions are shown. 
We list examples of the top $1\%$-percentile on the METEOR score.
For positive examples, the content of the video is fully recognized leading to a description that matches one of the reference captions almost exactly. In the bad examples (see supplementary) the content is often misinterpreted or the video is so abstract that there could be many diverse explanations. 


\subsubsection{Spatio-Temporal Attention}
\begin{figure*}
    {\centering 
    \includegraphics[width=\linewidth]{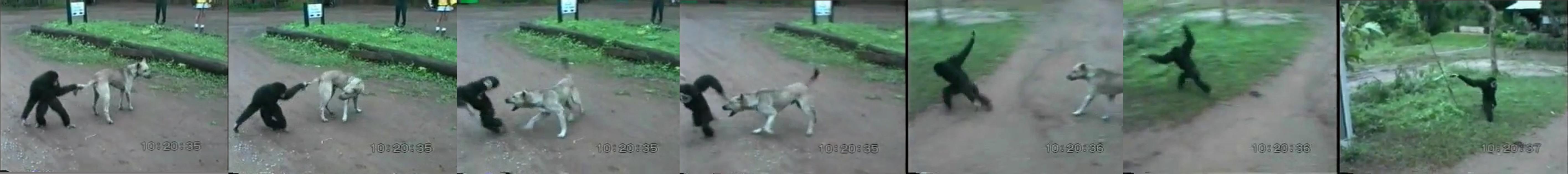} 
    }
    \scriptsize{ 
        \textbf{Reference:} A monkey pulling on a dog 's tail and running away
        \\
        \textbf{Generated text:} a monkey grabs a dog's tail and runs away\\[1ex] 
    }
    {\centering
    \includegraphics[width=\linewidth]{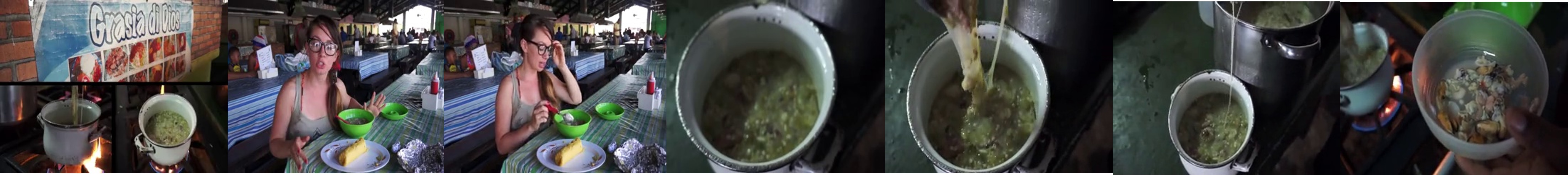} 
    }
    \scriptsize{
        \textbf{Reference:} a girl is talking about some food she is eating\\
        \textbf{Generated text:} a woman is cooking in a pot and talking about it 
    }
 \vspace{-0.3cm}

    \caption{Spatio-temporal inference gathers information from different segments.
    }%
    \label{fig:spatioTemporalSaliency}
\end{figure*}
The video in Figure~\ref{fig:spatioTemporalSaliency} on the top features a complex temporal interaction between two actors ({\it monkey and dog}). The generated caption correctly reflects both spatial detail ({\it dog's tail}) as well as multiple temporal stages ({\it grabbing the tail} and {\it running away}). 
Similarly, the temporal domain is also respected in the second example. Different parts of the video contribute to different sections in the generated captions ({\it woman talking about food} and {\it cooking in pot}). 
These examples demonstrate that high-quality detection and tracking from the Swin transformer across multiple frames goes hand-in-hand with the powerful language skills of the fine-tuned generator in our proposed framework. 

\section{Limitations and Discussion}
The introduced \modelName architecture performs quite well according to the commonly used evaluation metrics.
Even though our model has the best diversity scores, looking at some samples of generated captions, they are often rather general and might miss some important detail about the video. This suggests, further research in the diversity aspect is important.


As indicated in~\Cref{sec:exampleResults}, the current extraction of concepts for video captioning might need further improvement, potentially by the use of a larger training data set. Compared to the very good performance of the produced language, the visual analysis is not yet on par. 
The training and the evaluation are done on three data sets MSR-VTT, MSVD and VATEX, 
which come with their own distributions of scenes, people, objects, and actions. Any marginalization of specific social groups present in the data sets will likely also be present in our trained encoder-decoder framework. 
In general, our approach to automated video analysis and captioning might furthermore be trained and applied in other contexts. While we think that the application to the presented data is not problematic, ethical issues can quickly arise in surveillance or military applications.

\section{Conclusion}
We presented \modelName, a video captioning encoder-decoder approach which incorporates the processed visual tokens by multi-layer multi-head cross-attention. While the Swin tokens extracts separate spatio-temporal information, we introduce also a globally aggregating semantic concept vector that initializes the sentence generation in the BERT module. 
By proposing a content-based adaptive frame selection sampling, we can assure that the most informative frames are selected while maintaining efficient training.

This transformer-based video captioning framework introduces a new architecture that produces plausible captions that are state-of-the-art on the MSR-VTT, MSVD and VATEX benchmark data sets. Our evaluation highlights that the commonly used NLG metrics only address some of the aspects necessary to fully assess the quality of video descriptions. We demonstrate that our method produces highly diverse captions. We hope that this work will inspire further research for a better, broader assessment of the performance of caption generation algorithms.

\section{Acknowledgements} This work has been supported by the German Research Foundation: EXC 2064/1 – Project number 390727645, the CRC 1233 - Project number 276693517, as well as by the German Federal Ministry of Education and Research (BMBF): Tübingen AI Center, FKZ: 01IS18039A. The authors thank the International Max Planck Research School for Intelligent Systems (IMPRS-IS) for supporting Zohreh Ghaderi and Leonard Salewski. 

\bibliographystyle{splncs04}
\bibliography{main}

%
%

%
%
%





\newpage
\section{Supplementary: Diverse Video Captioning by Adaptive Spatio-temporal Attention.}
%


This supplementary presents model details in~\Cref{sec:architecture}, and provides additional ablation studies on our decoder model in~\Cref{tab:ablation}. 
Also, an analysis on the diversity of generated captions is shown in~\Cref{sec:diversity} and demonstrates additional qualitative results in~\Cref{sec:qualExamples}.
Training detail of our model is explained in~\Cref{tab:training}.

\subsection{Architecture}%
\label{sec:architecture}
\label{sec:architecture}
In this section, we show technical sketches of the three subnetworks in our \modelName{} framework and present additional details of the network structures and configurations.  Also, our code is available at \url{https://github.com/zohrehghaderi/VASTA}.
\subsubsection{Data Preparation}
The input videos are read with the mmcv library\footnote{\url{https://github.com/open-mmlab/mmcv}} choosing the NCTHW input shape format. Each input is resized and cropped to $224\times224$ resolution followed by normalization with 
mean of $[123.675,116.28,103.53]$ and standard deviation of $[58.395, 57.12,57.375]$. 

The adaptive frame selection (AFS, see Section 3.1 of the main paper) extracts the 32 most informative frames based on LPIPS~\cite{zhang2018perceptual}  similarity scores and passes the video with shape $32\times 224\times 224\times 3$ to the Swin encoder.

\subsubsection{Data set splits}

\textbf{MSVD~\cite{chen2011collecting}.} Following~\cite{chen2011collecting,yao2015describing} the data set is split into 1200 samples for training, 100 samples for validation and the remaining 670 samples for testing.
\textbf{{MSR-VTT}~\cite{xu2016msr}.}
Following the official setting~\cite{xu2016msr}, the data set is split as follows: 6513, 497 and 2990 videos for training, validation and test.
\textbf{VATEX~\cite{wang2019vatex}.}
This data set officially includes 25991, 3000, 6000 videos for train, validation and test.
Unfortunately, roughly 10\% of the original set are no longer available for download. 
Thus, our evaluation in paper is on 23303 videos for training, 2690 videos for validation and 5398 videos for test (see Table \cref{tab:dataset-splits}).


\begin{table*}
   \centering {
        \begin{tabular}{rccc}
            \toprule
            & MSR-VTT~\cite{xu2016msr} & MSVD~\cite{chen2011collecting} & VATEX~\cite{wang2019vatex}\\
            \midrule
            Train & 6513 & 1200 & 23303\\
            Val & 497 & 100 & 2690\\
            Test & 2990 & 670 & 5398\\
            \midrule
            Total &10000 &1970 & 31391 \\
            \bottomrule
        \end{tabular}
    }
    \caption{Train, validation and test splits of the utilized data sets. 
    }%
    \label{tab:dataset-splits}
\end{table*}

\subsubsection{Swin Encoder}
\begin{figure}
    \centering
    {
        \hfill \includegraphics[width=0.48\linewidth]{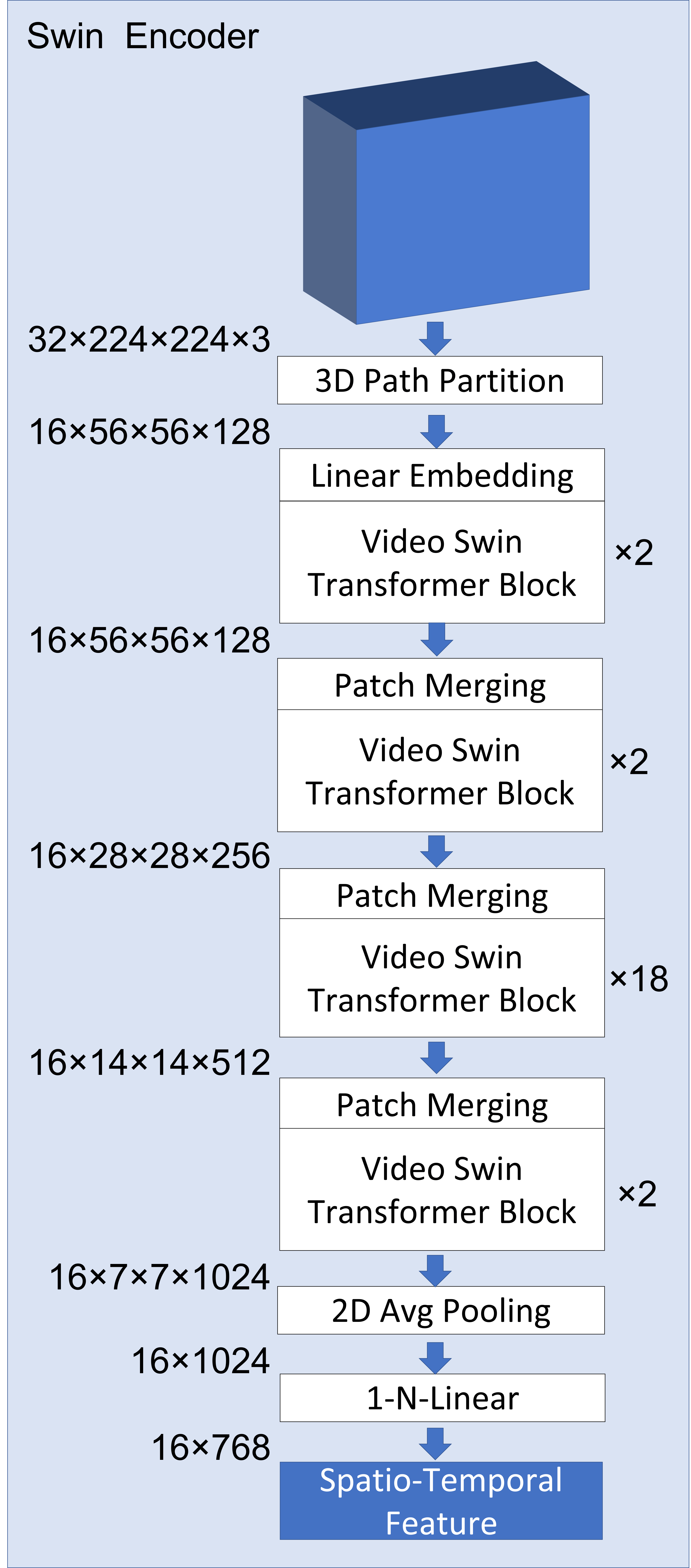} \hfill
        \hfill  \includegraphics[width=.48\linewidth]{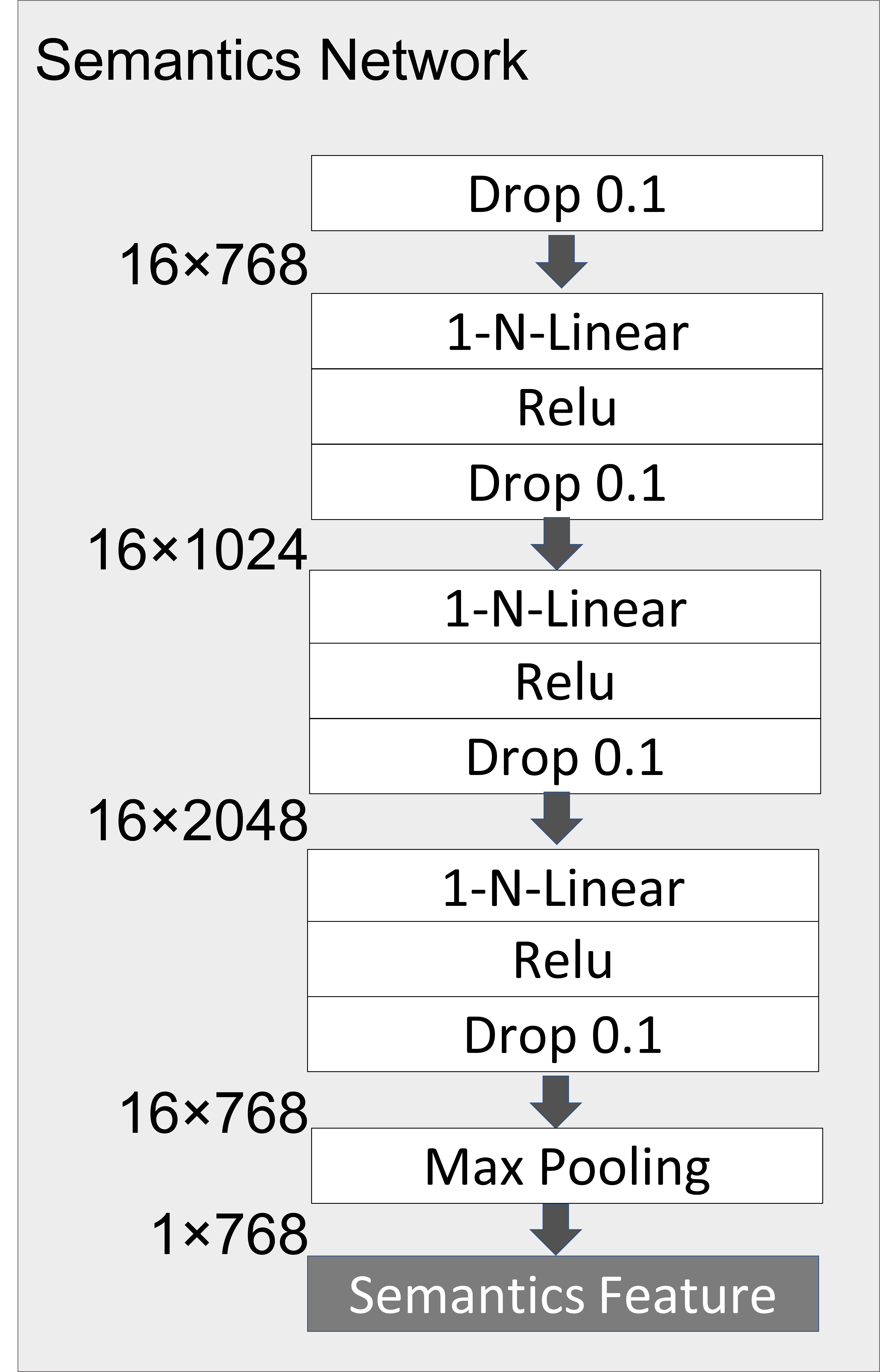} \hfill
        \\
        \parbox{0.48\linewidth}{\centering
        \small{(a)}
        }
        \parbox{0.48\linewidth}{\centering
        \small{(b)}
        }
    }
    
    \caption{a) Detailed architecture of our encoder model based on the Video Swin Transformer~\cite{liu2021video}. b) Detailed architecture of our semantics model. Each token of the Swin transformer output is processed individually by a three-layer MLP before fusing all 16 tokens by a max layer.}%
    \label{fig:encoder}
\end{figure}

The encoder backbone to our \modelName{} model is the Swin network, more precisely the Swin-B variant~\cite{liu2021swin}. While the original pipeline is designed for action classification in our context, video description, we modify the network as shown in Figure~\ref{fig:encoder}.a, generating 16 output tokens that are reshaped to fit the expected input size of the BERT decoder.
More details on the specific configuration and parameters of the Swin transformer~\cite{liu2021swin} are listed in \Cref{tab:framework}.

\subsubsection{Semantic Context Network}

The network to extract the semantic context vector is three-layer MLP that is shown in Figure~\ref{fig:encoder}.b. The MLP operates on each Swin token individually and then joins them using max-pooling. The resulting vector yields the probability of the most frequent $768$ words.

The predicted semantic feature is passed as the start-of-sequence (\texttt{[SOS]}) token to the BERT decoder where it replaces the BERT-embedding layer for the first time step.


\subsubsection{BERT Decoder}

\begin{figure}
    \centering{
    \includegraphics[width=.45\linewidth]{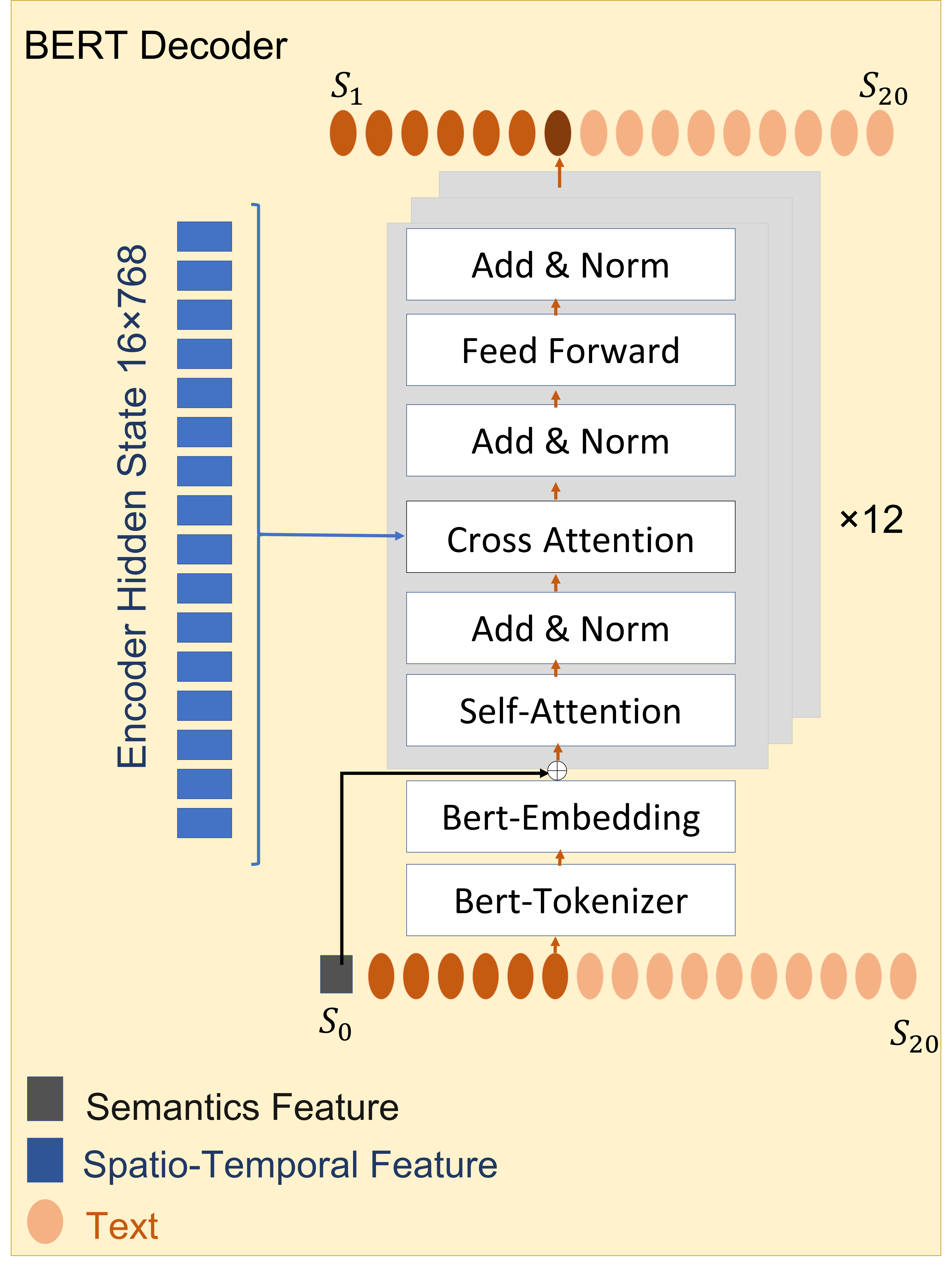} 
    }
    
    \caption{Detailed architecture of our BERT-based decoder model. The decoder considers the visual input once by using the semantic vector as \sos token and second by cross-attending to it in the 12 transformer layers. Note, that the conditioning on the semantic vector bypasses the embedding step.}%
    \label{fig:decoder}
\end{figure}

The architecture of the BERT decoder is shown in~\Cref{fig:decoder}.
It follows the traditional BERT architecture~\cite{devlin2018bert} with an additional cross-attention layer in the 12 transformer blocks which relates the BERT inference to the encoder output sequence. The specific configuration of the layers and attention heads is given in~\Cref{tab:framework}.

\subsection{Ablation Study}

\subsubsection{Different Inference Methods.}
As illustrated in ~\Cref{tab:ablations_inference}, we compare frequently used inference methods applied to our \modelName{} model. Beam search with 3 beams achieves the best results on the MSVD and MSR-VTT data sets.

\begin{table*}
    \centering{
        \begin{tabular}{rcccccccc}
            \toprule
             & \multicolumn{4}{c}{MSR-VTT} & \multicolumn{4}{c}{MSVD} \\
            \cmidrule(r){2-5}\cmidrule(r){6-9}
            Method & B4 $\uparrow$  & M $\uparrow$                   & C $\uparrow$            & R $\uparrow$       & B4 $\uparrow$        & M $\uparrow$        & C $\uparrow$          & R $\uparrow$       \\
             \midrule
            Top-p p=0.95   &     42.28                 &        29.53            & 53.00 & 62.28 &  54.56   & 38.52  & 102.5  & 74.15 \\
            Top-k k=20   & 19.03                       & 22.18                    & 28.50 & 46.39 & 30.96  & 29.58  & 57.50  & 61.48 \\
            Top-k k=3   &         29.59               &           25.85          & 39.12 & 54.16 & 41.78  & 33.79  & 74.73  & 67.04  \\
            Greedy          &       42.28                 &       29.53              &  53.00& 62.28& 54.56 & 38.52 & 102.5 & 74.15 \\
            \midrule
            Beam Search b=3 &\textbf{ 43.43}                       & \textbf{30.24}                    & \textbf{55.00}          & \textbf{62.54}         & \textbf{56.14}         & \textbf{39.09}         & \textbf{106.3}         & \textbf{74.47}   \\
            \bottomrule
        \end{tabular}
    }
    \caption{Influence of the difference inference method on \modelName{} model.}%
    \label{tab:ablations_inference}
\end{table*}


\subsubsection{Ablation on the decoder model.}%
\label{tab:ablation}
To study the effect of different pre-trainings of our language decoder we try two different decoder models.
GPT~\cite{radford2018improving} is an auto-regressive language model whose aim is to predict the next word based on all of previous words.
Huge amounts of data are used to train the large number of parameters.
Additionally, the size of the GPT model with its 125M parameters limits the ability to combine it with a large Swin network.
Thus we use the GPTNeo Causal Language Modelling~\cite{black2021gpt} model, which is similar to GPT2 except that GPTNeo uses local attention in every other layer.
We compare the performance when replacing the BERT decoder by GPTNeo in~\Cref{tab:BERTGPT}.

\begin{table*}
    \resizebox{\linewidth}{!}{
        \begin{tabular}{rcccccccccccc}
            \toprule
            
            & \multicolumn{4}{c}{MSR-VTT} & \multicolumn{4}{c}{MSVD} & \multicolumn{4}{c}{VATEX} \\
            \cmidrule(r){2-5}\cmidrule(r){6-9}\cmidrule(r){10-13}

            Decoder & B4 $\uparrow$                      & M $\uparrow$                   & C $\uparrow$          & R $\uparrow$       & B4 $\uparrow$        & M $\uparrow$        & C $\uparrow$          & R $\uparrow$     & B4 $\uparrow$        & M $\uparrow$        & C $\uparrow$          & R $\uparrow$        \\
            \midrule
            BERT         & \textbf{43.43}                       & \textbf{30.24}                    & \textbf{55.00}& \textbf{62.54} &  \textbf{ 56.14}                       & \textbf{39.09}                    & \textbf{106.3} & \textbf{74.47}& \textbf{34.96} & \textbf{25.46} & \textbf{51.33} & \textbf{64.33} \\
            GPT-Neo      & 40.13 & 28.09 & 46.91 & 57.25 & 50.83 & 33.80 & 81.40 & 65.89 & 27.25 & 24.50 & 47.01 & 38.79            \\
           \bottomrule
        \end{tabular}
    }
    \caption{Comparison of BERT and GPTNeo~\cite{black2021gpt} as decoders for our model \modelName. BERT clearly outperforms GPTNeo as a decoder.}%
    \label{tab:BERTGPT}
\end{table*}

We hypothesize, that BERT is trained on mask filling whose aim is to predict a masked word bases on previous words and upcoming words in the sentences.
Thus, it is leading to improved sentence understanding which is related to video understanding for caption generation.

\begin{table*}
    \resizebox{\linewidth}{!}{
        \begin{tabular}{lclclc}
            \toprule
            \multicolumn{2}{c}{Swin Encoder} & 
            \multicolumn{2}{c}{Semantics Network} &
            \multicolumn{2}{c}{BERT Decoder} \\
            \cmidrule(r){1-2}\cmidrule(r){3-4}\cmidrule(r){5-6}
            Type & Size & Type & Size & Type & Size      \\
            \midrule
            Patch-size & (2,4,4) & Linear & 1024 & Hidden-size & 768   \\
            Depths & [2,2,18,2] & Activation & RELU & Num-hidden-layers & 12 \\
            Embed-dim & 128 & Drop out train & 0.5 & Num-attention-heads & 12 \\
            Num-heads & [4,8,16,32] & Linear & 2048 & Intermediate-size & 2872 \\
            Window-size & (8,7,7)& Activation & RELU & Hidden-act & Gelu \\
            MLP-ratio & 4 & Drop out train & 0.5 & Hidden-dropout-prob & 0.3 \\
            qkv-bias & True & Max pooling & 1D & Attention-probs-drop-out-prob & 0.3 \\ 
            qk-scale & None & Drop out fine tunig & 0.1 &Max-position-embeddings & 512 \\
            Drop-rate & 0 &&& Type-vocab-size&2\\
            Attn-drop-rate & 0 & & &  Initializer-rang & 0.02\\
            Drop-path-rate & 0.2  & & & Layer-norm-eps & 1e-12 \\
            Path-norm & True & & & Vocab size & 30522\\
            &  & & & Position-embedding-type & absolute \\
            &  & & & Pad-token-id  & 0 \\
            \bottomrule
        \end{tabular}
    }
    \caption{Parameter sets and configurations for the three subnetworks in our pipeline: the Swin encoder, the semantics network and the BERT decoder.}%
    \label{tab:framework}
\end{table*}

\newpage
\subsection{Qualitative Examples}%
\begin{figure*}
    {
    \includegraphics[width=\linewidth]{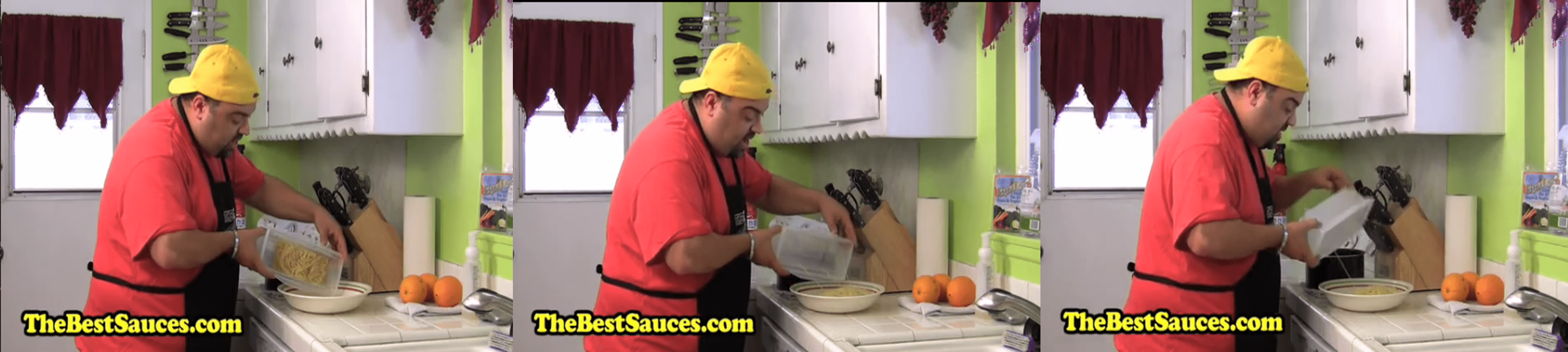} 
    }
 
        \small{
        \begin{tabular}{lll}
        \textbf{Video
        \texttt{hJFBXHtxKIc-286-291}}\\
        \textbf{Reference}: a man pours cooked pasta from a plastic container into a bowl& CIDEr \\
        \textbf{AFS-Swin-Bert}: a man cooking his kichen & 35.04 \\
        \textbf{AFS-Swin-Bert-Semantics}: {a man is pouring pasta out of a plastic container} & 75.86 \\
        \end{tabular}
        }
       \newline
    {
    \includegraphics[width=\linewidth]{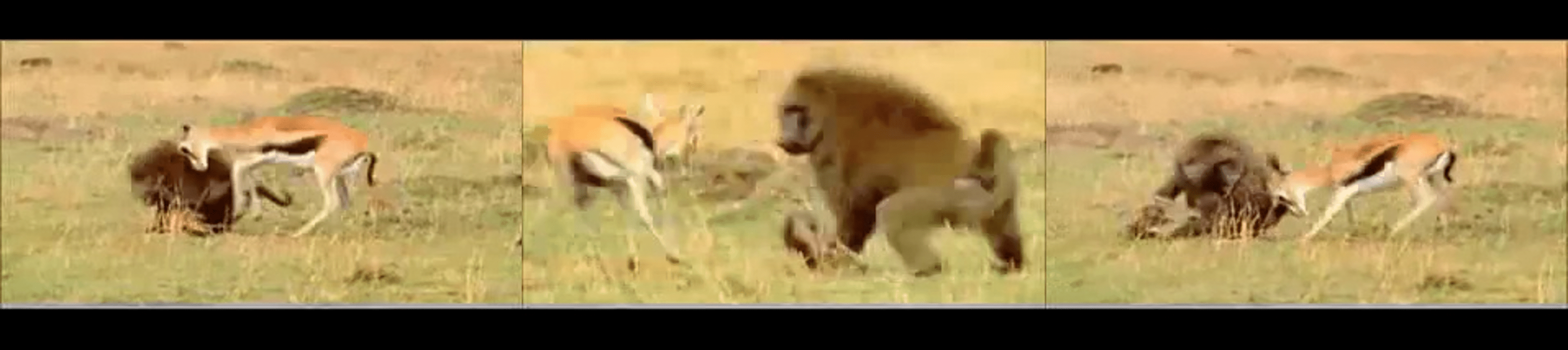} 
    }
        \small{
        \begin{tabular}{lll}
        \textbf{Video
        \texttt{qypmR4O1Gwk-0-10}}\\
        \textbf{Reference}:  a gazelle is fighting with a baboon & CIDEr \\
        \textbf{AFS-Swin-Bert}:  two zebras are fighting & 23.74 \\
        \textbf{AFS-Swin-Bert-Semantics}:  a cheetah is chasing a gazelle & 57.89 \\
        \end{tabular}
        }
       
    {
    \includegraphics[width=\linewidth]{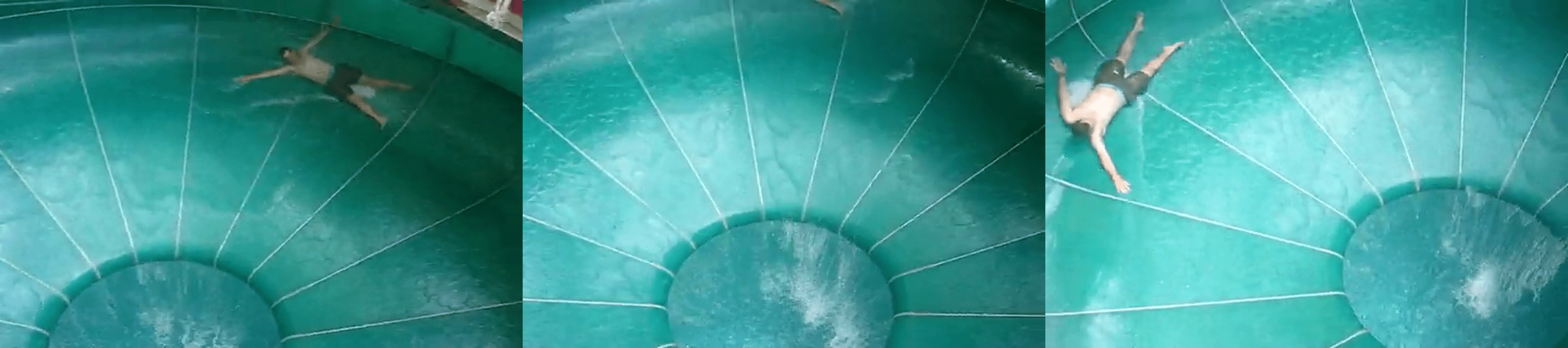} 
    }
        \small{
        \begin{tabular}{lll}
        \textbf{Video
        \texttt{idXJu0BQRvo-2-6}}\\
        \textbf{Reference}: a boy is sliding around a water slide & CIDEr \\
        \textbf{AFS-Swin-Bert}: a dog is swimming in a pool & 26.64 \\
        \textbf{AFS-Swin-Bert-Semantics}: a man is sliding down a water slide & 208.0 \\
        \end{tabular}
        }
        
    {
    \includegraphics[width=\linewidth]{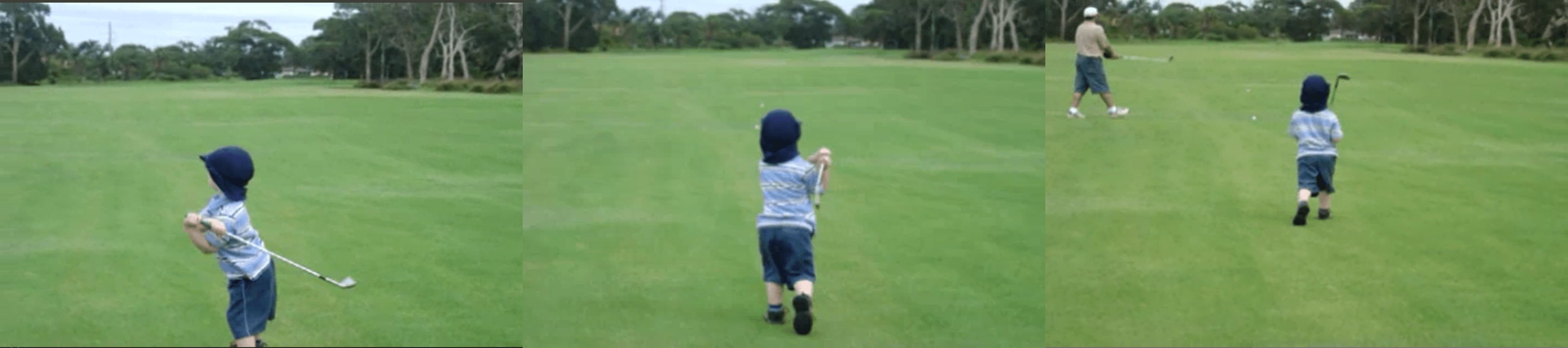} 
    }

        \small{
        \begin{tabular}{lll}
        \textbf{Video \texttt{zpgW7m7-LZw-2-15}}\\
        \textbf{Reference}: a little boy is playing golf & CIDEr \\
        \textbf{AFS-Swin-Bert}: a man is riding a bicycle & 0.005 \\
        \textbf{AFS-Swin-Bert-Semantics}: a boy is playing & 157.5 \\ 
        \end{tabular}
        }
      
    \caption{Effect of considering the semantic context vector in caption generation. In these videos from the MSVD collection the CIDEr score has been effectively increased.
    }%
    \label{fig:msvd-output}
\end{figure*}

\begin{figure*}[ht]
   
    {
    \includegraphics[width=\linewidth]{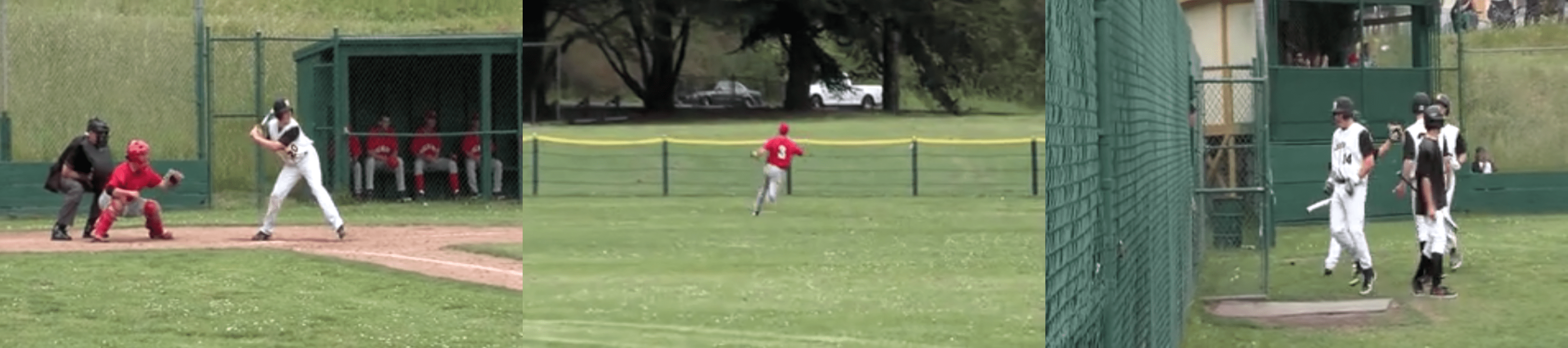} 
    }
        \small{
        \begin{tabular}{lll}
        \textbf{Video
        \texttt{7021}}\\
        \textbf{Reference}: a baseball batter hits the ball & CIDEr \\
        \textbf{AFS-Swin-Bert}: there is a man is playing with a ball & 29.45 \\
        \textbf{AFS-Swin-Bert-Semantics}: a man is playing baseball in a field & 115.1 \\
        \end{tabular}
        }
       \newline
    {
    \includegraphics[width=\linewidth]{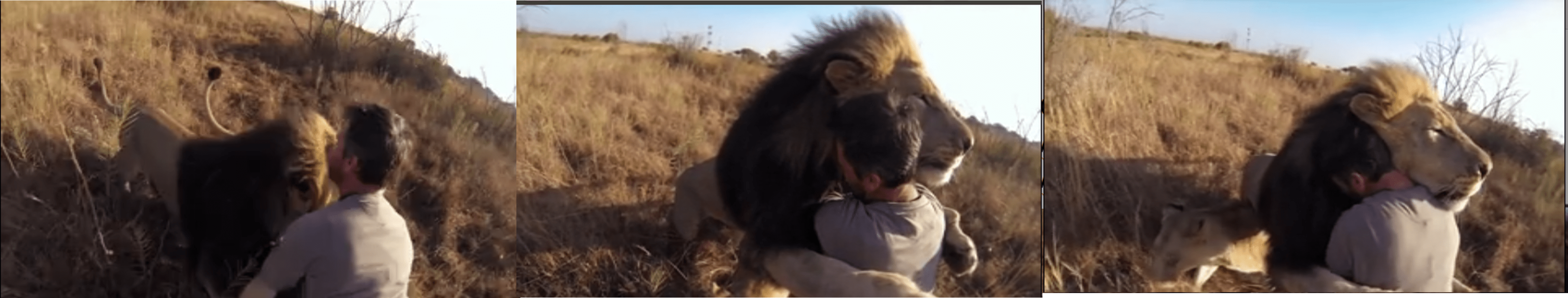} 
    }
   
        \small{
        \begin{tabular}{lll}
        \textbf{Video
        \texttt{7053}}\\
        \textbf{Reference}: a man playing with a lion & CIDEr \\
        \textbf{AFS-Swin-Bert}: a lion is playing with a lion & 155.6 \\
        \textbf{AFS-Swin-Bert-Semantics}: a man is playing with a lion & 300.9 \\
        \end{tabular}
        }
    \newline
    {
    \includegraphics[width=\linewidth]{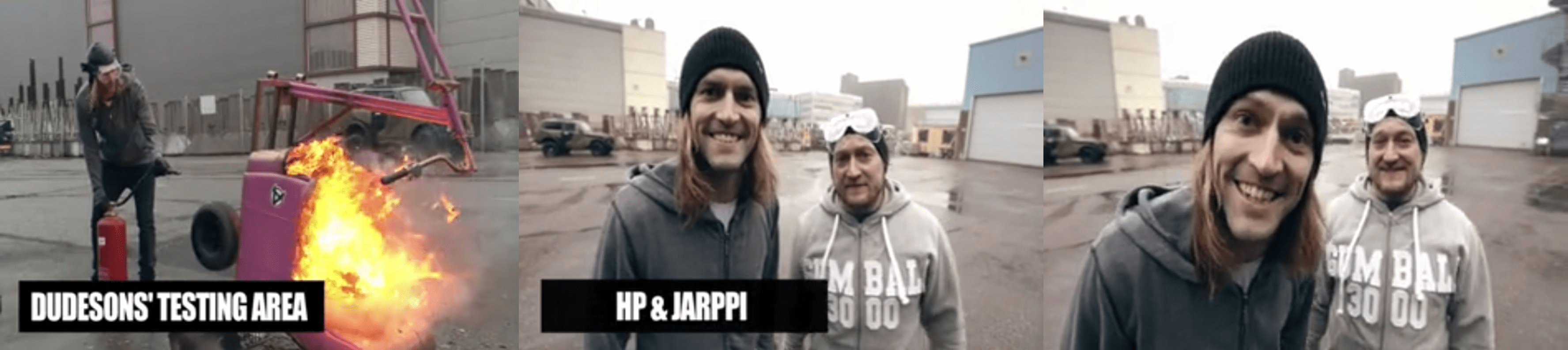} 
    }
        \small{
        \begin{tabular}{lll}
        \textbf{Video
        \texttt{7060}}\\
        \textbf{Reference}: two men are giving an introduction & CIDEr \\
         \textbf{AFS-Swin-Bert}: a man is talking to another man & 3.20 \\
        \textbf{AFS-Swin-Bert-Semantics}: a person is explaining something & 101.1\\
        \end{tabular}
        }
        \newline
    {
    \includegraphics[width=\linewidth]{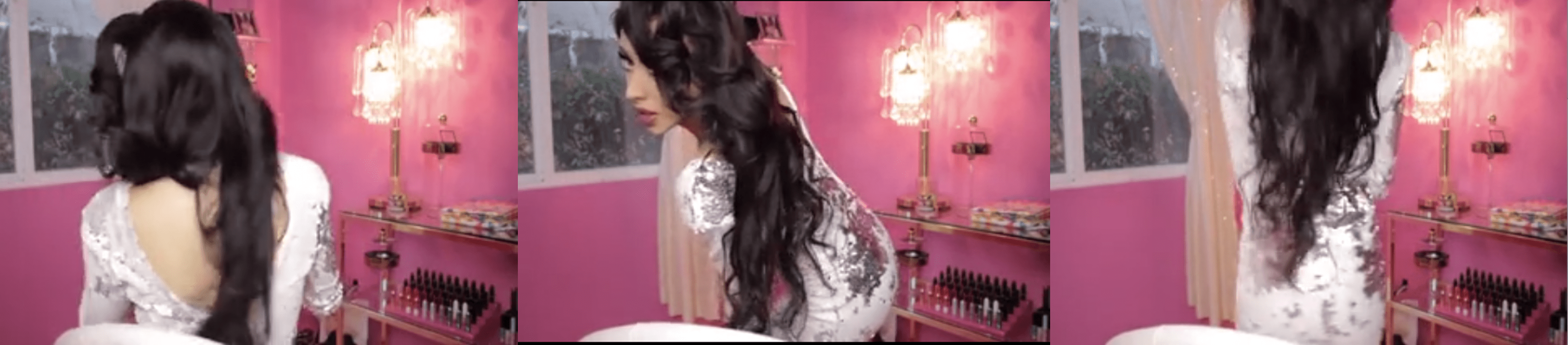} 
    }
    \small{
        \begin{tabular}{lll}
        \textbf{Video
        \texttt{7265}}\\
        \textbf{Reference}: a girl in her room shows off her new hairstyle & CIDEr \\
        \textbf{AFS-Swin-Bert}: a woman is showing how to wash her hair & 31.37 \\
        \textbf{AFS-Swin-Bert-Semantics}: a woman is showing off her hair & 73.79 \\
        
        \end{tabular}
    }
      
    \caption{Effect of considering the semantic context vector in caption generation. In these videos from the MSR-VTT collection the CIDEr score has been effectively increased. 
    }%
    \label{fig:msrvtt-output}
\end{figure*}

\begin{figure*}
    \centering

    \includegraphics[width=0.46\linewidth, height=6cm]{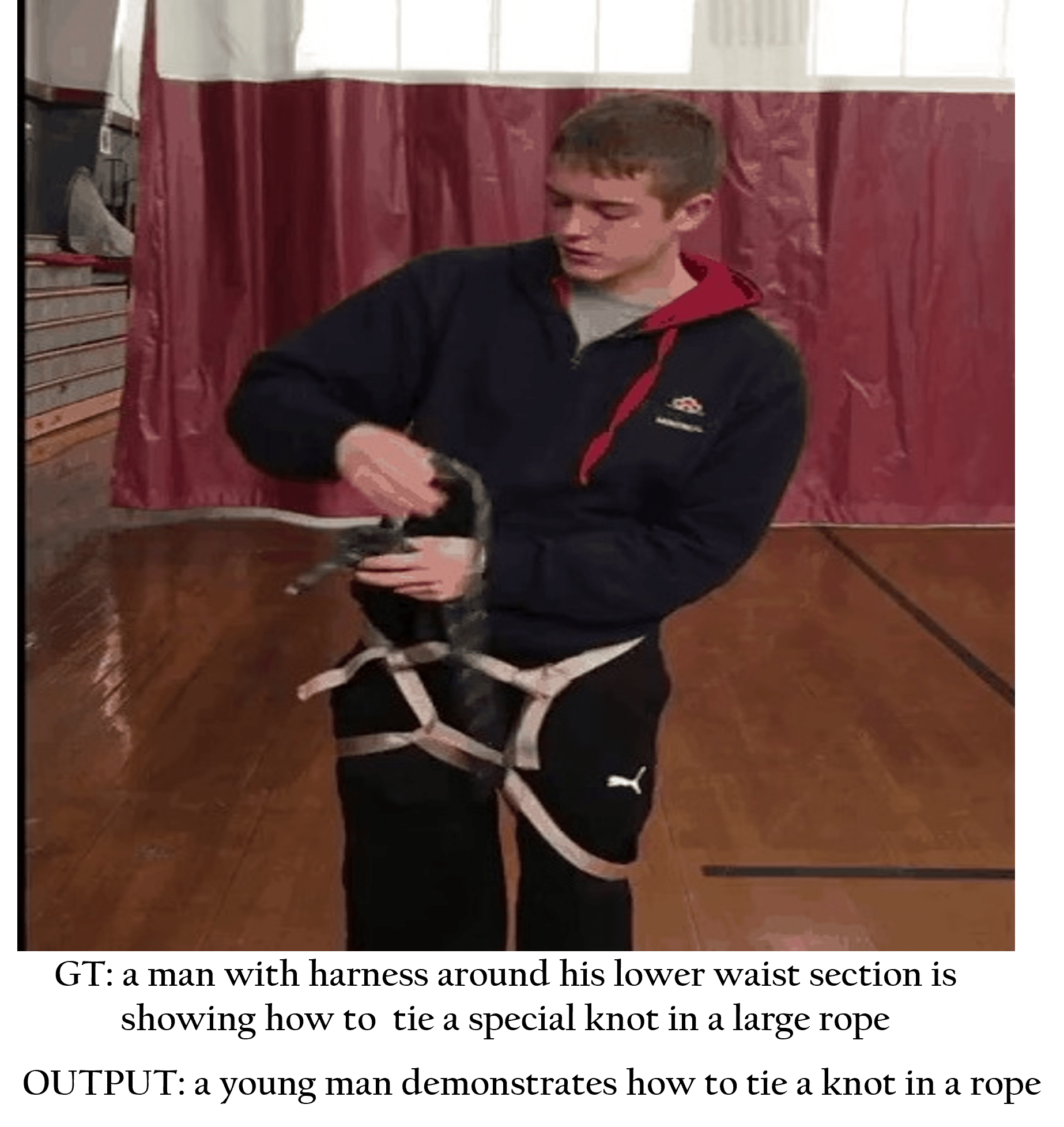}
    \includegraphics[width=0.46\linewidth, height=6cm]{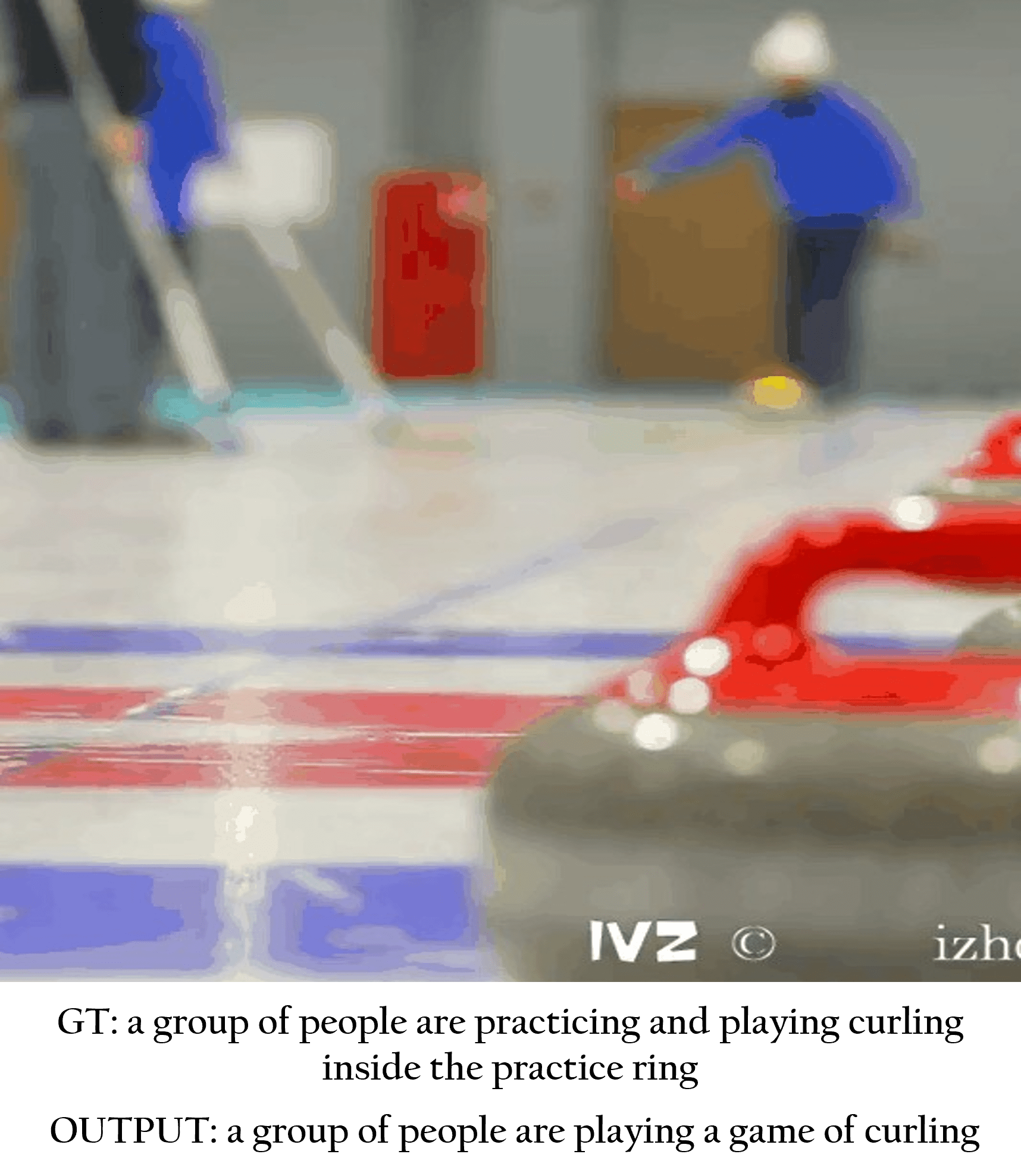}
    \\
    \includegraphics[width=0.46\linewidth, height=6cm]{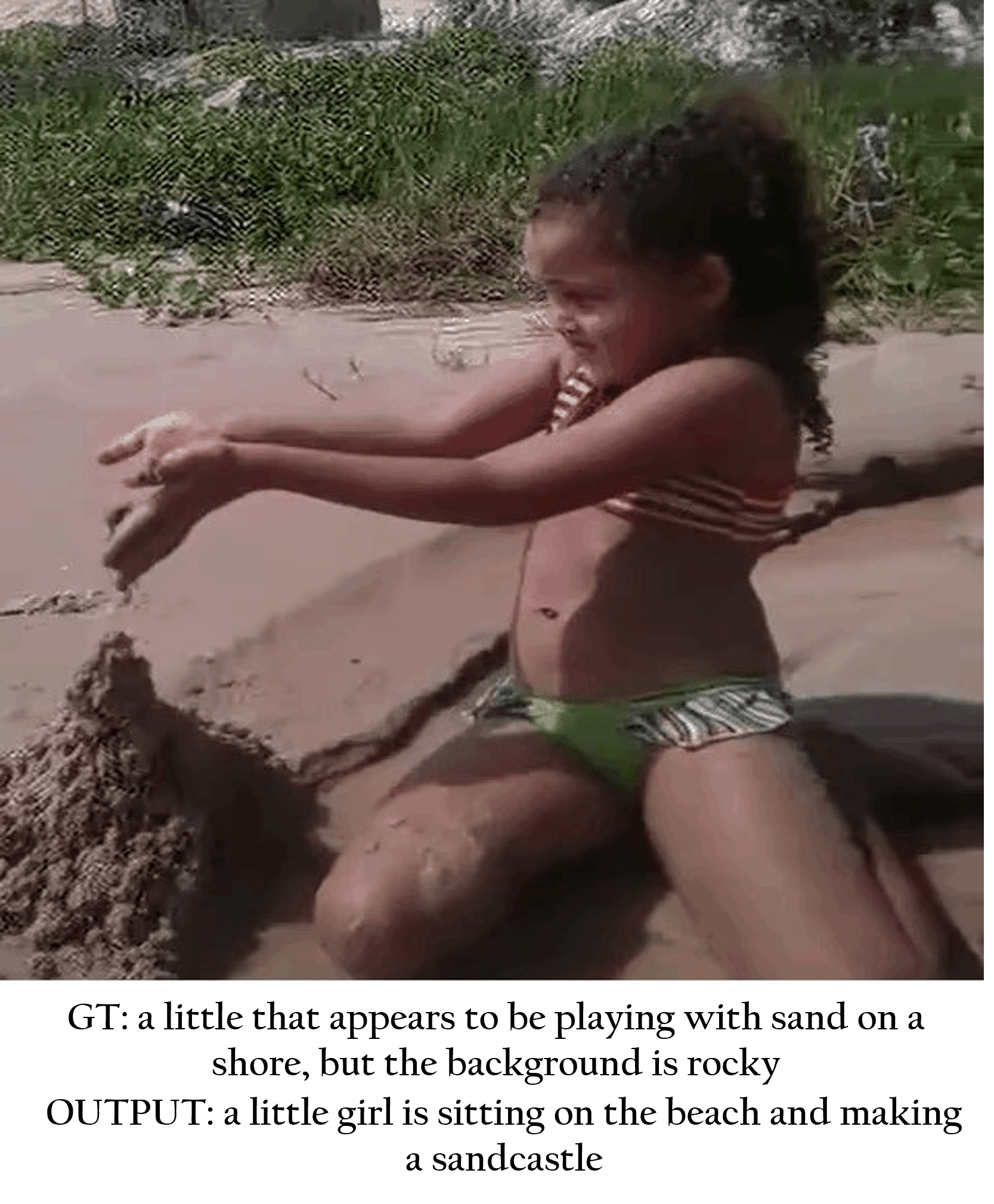}
    \includegraphics[width=0.46\linewidth, height=6cm]{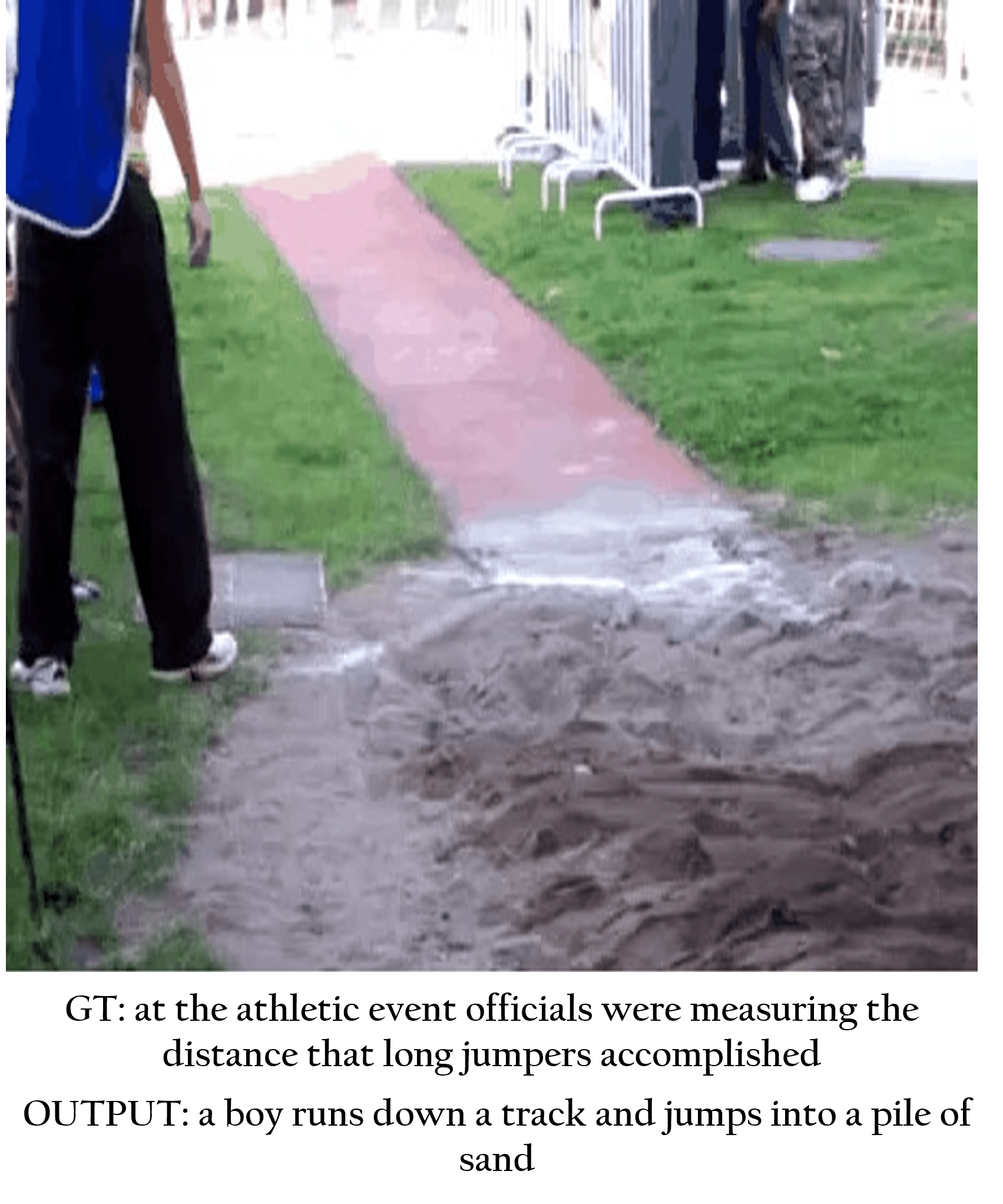}
    \\
    \includegraphics[width=0.46\linewidth, height=6cm]{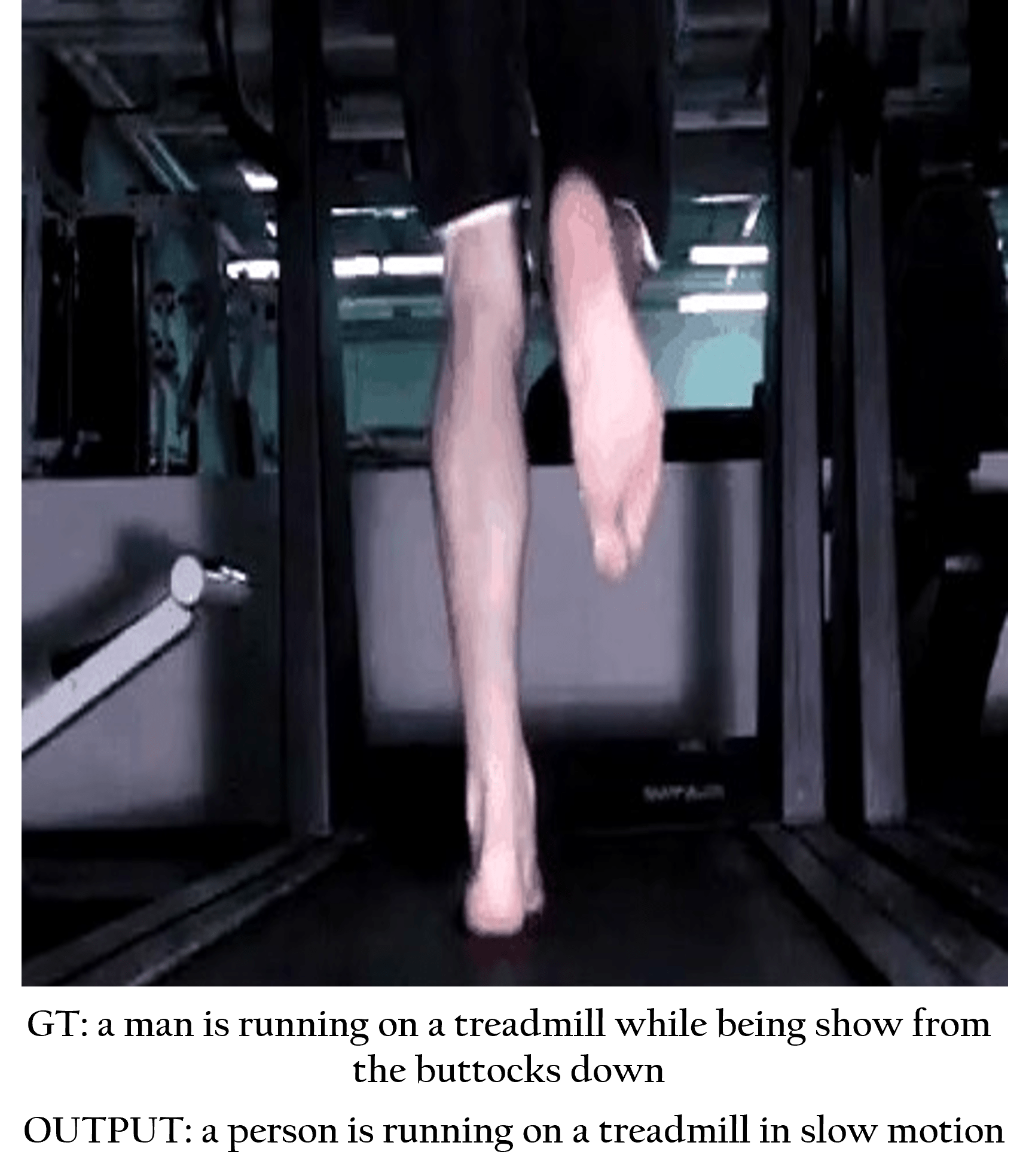}
    \includegraphics[width=0.46\linewidth, height=6cm]{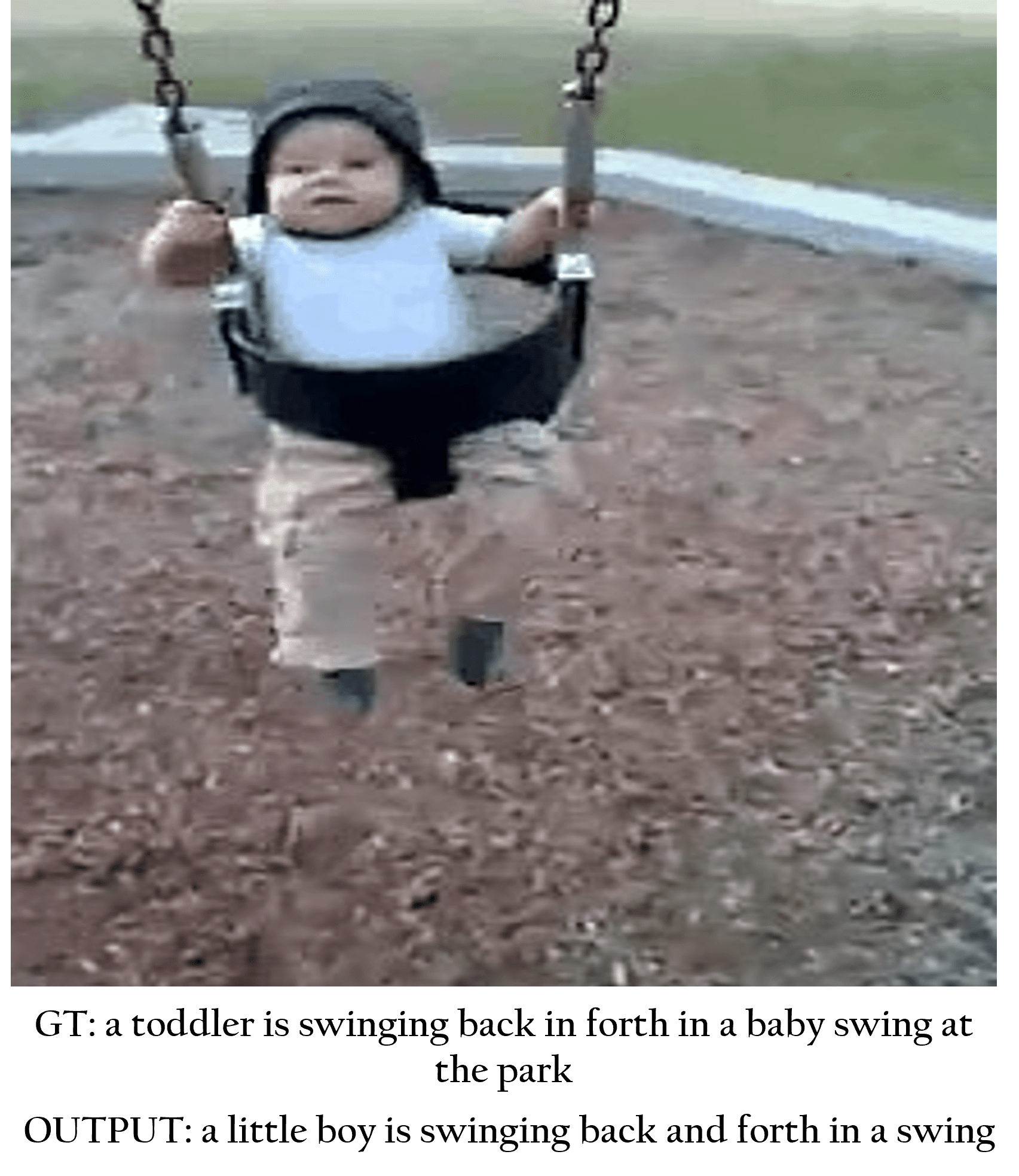}
    \\
      
    \caption{Based on METEOR and BLEU-4 we present a selection of the 10 percent best-performing videos in the  VATEX data set. GT refers to the ground truth reference.
   }%
    \label{fig:best-outvatex}
\end{figure*}

\label{sec:qualExamples}
To better assess the quality of the captions generated by our model, we present a few more generated examples. \Cref{fig:msvd-output}, \Cref{fig:msrvtt-output} and \Cref{fig:best-outvatex} feature some videos of the MSVD and the MSR-VTT data sets, respectively, where introducing the semantic context vector into our pipeline (see Section 4.3) improved the CIDEr score significantly compared to the version of our model without the semantic vector.

\begin{figure*}

    \includegraphics[width=0.31\linewidth]{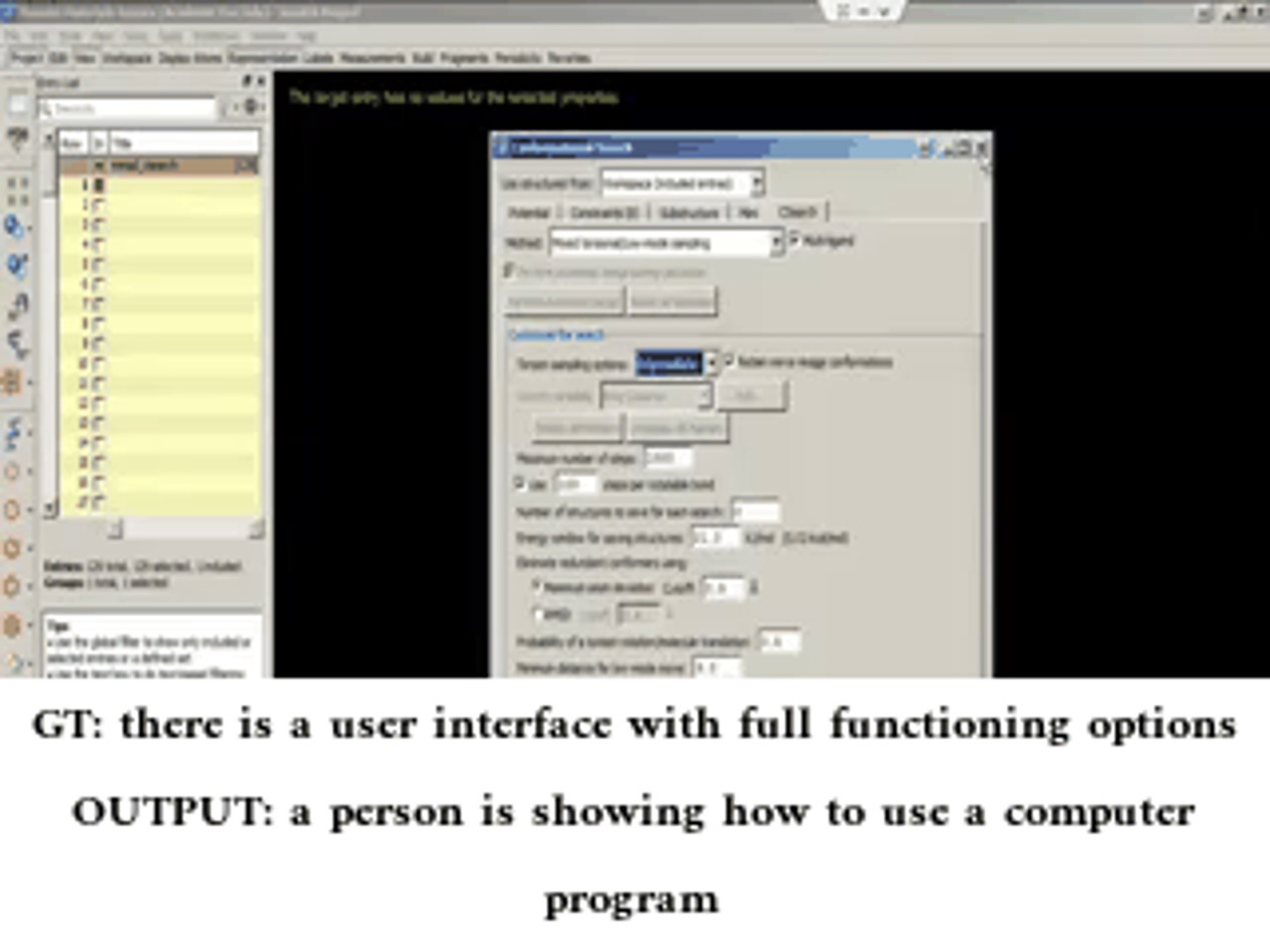}
    \includegraphics[width=0.31\linewidth]{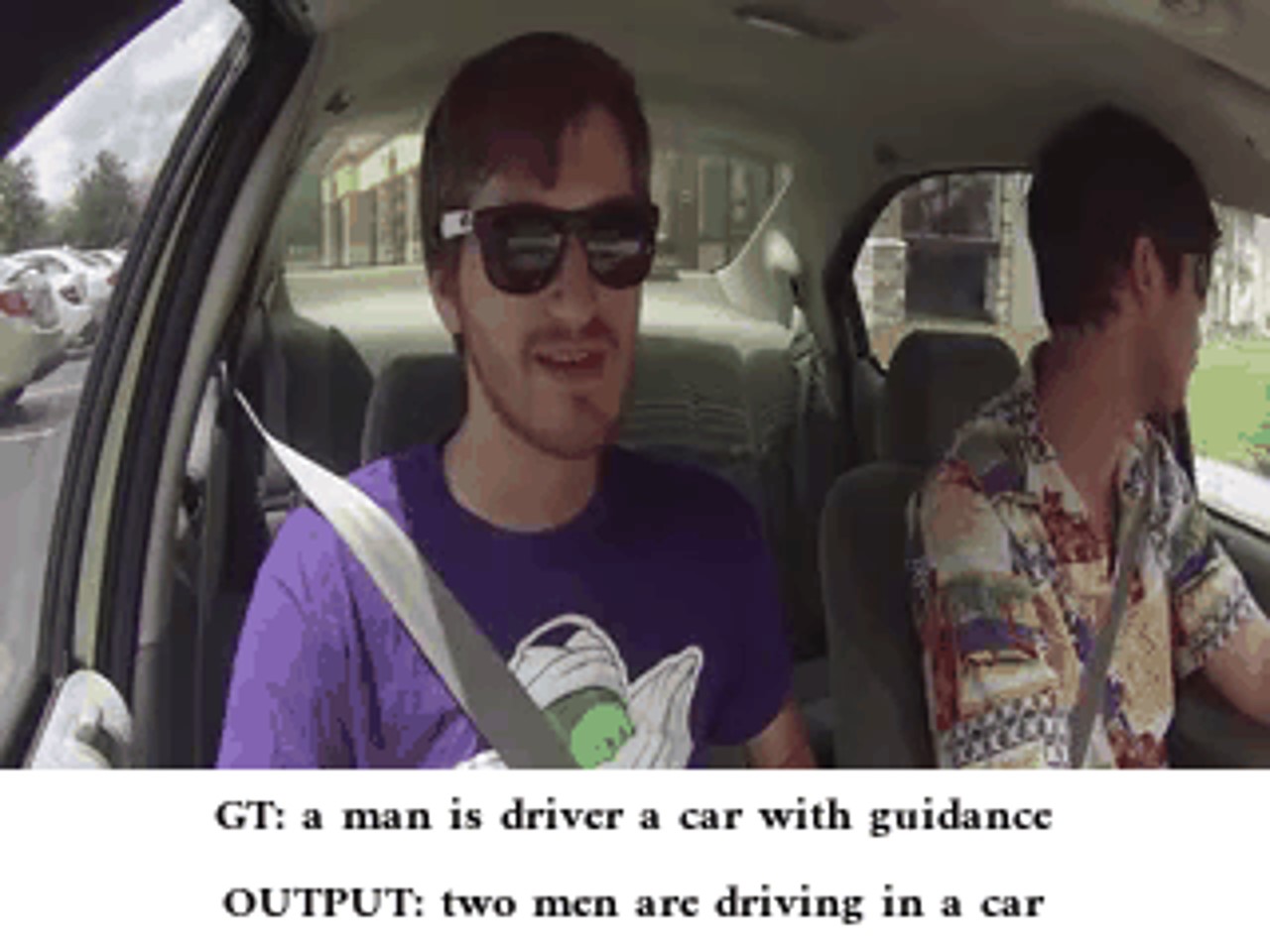}
    \includegraphics[width=0.31\linewidth]{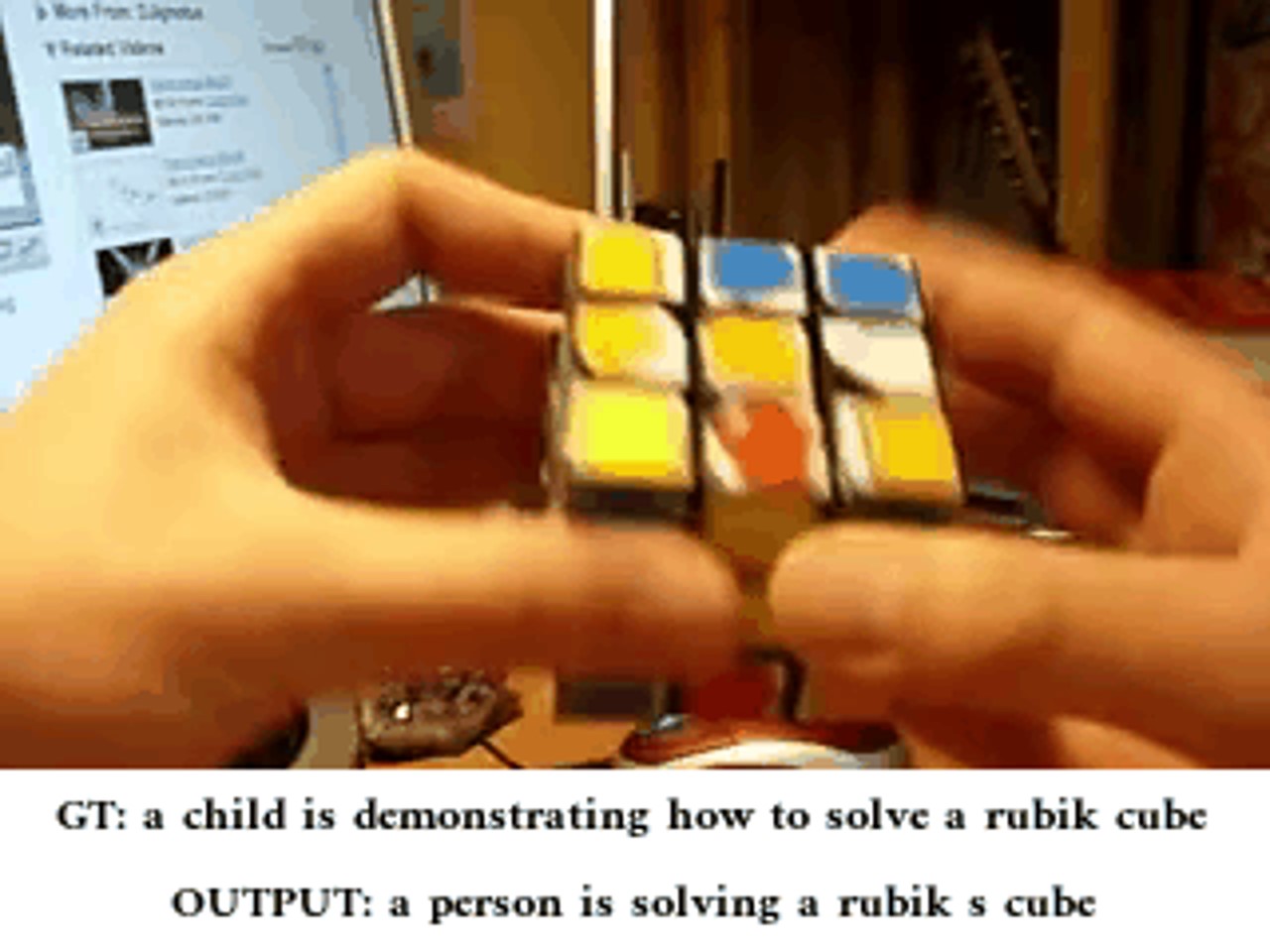}
    \\
    \includegraphics[width=0.31\linewidth]{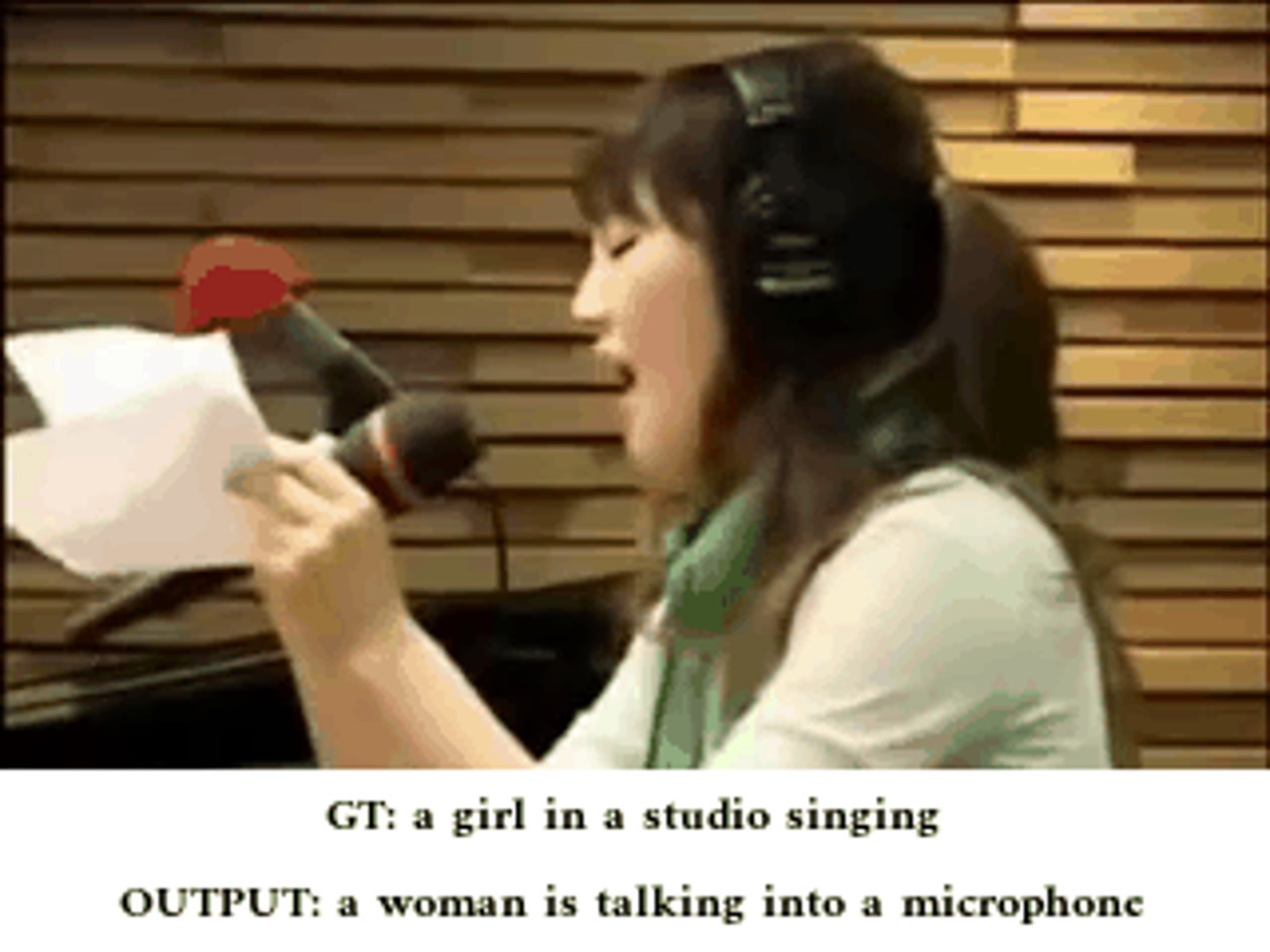}
    \includegraphics[width=0.31\linewidth]{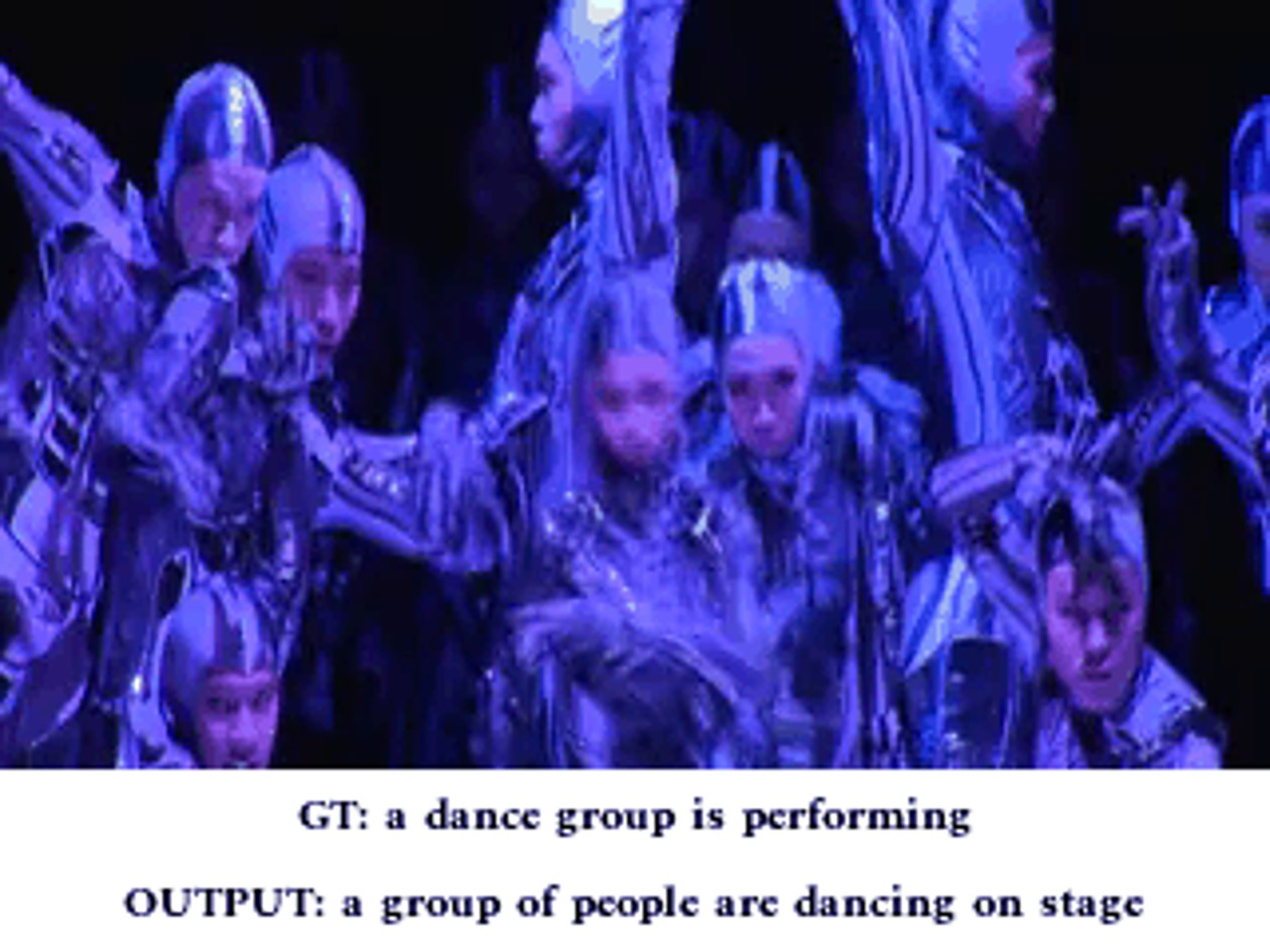}
    \includegraphics[width=0.31\linewidth]{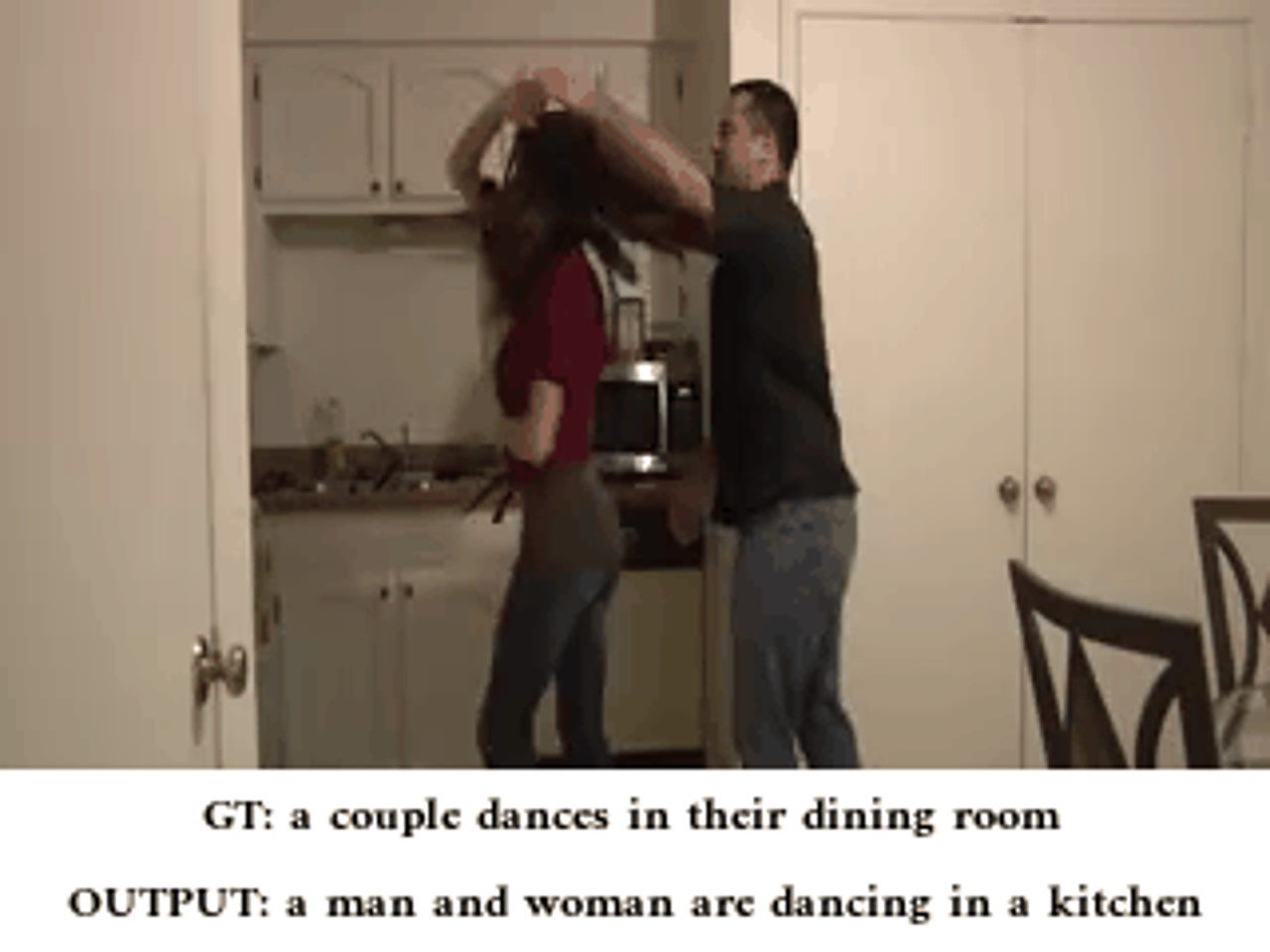}
    \\
    \includegraphics[width=0.31\linewidth]{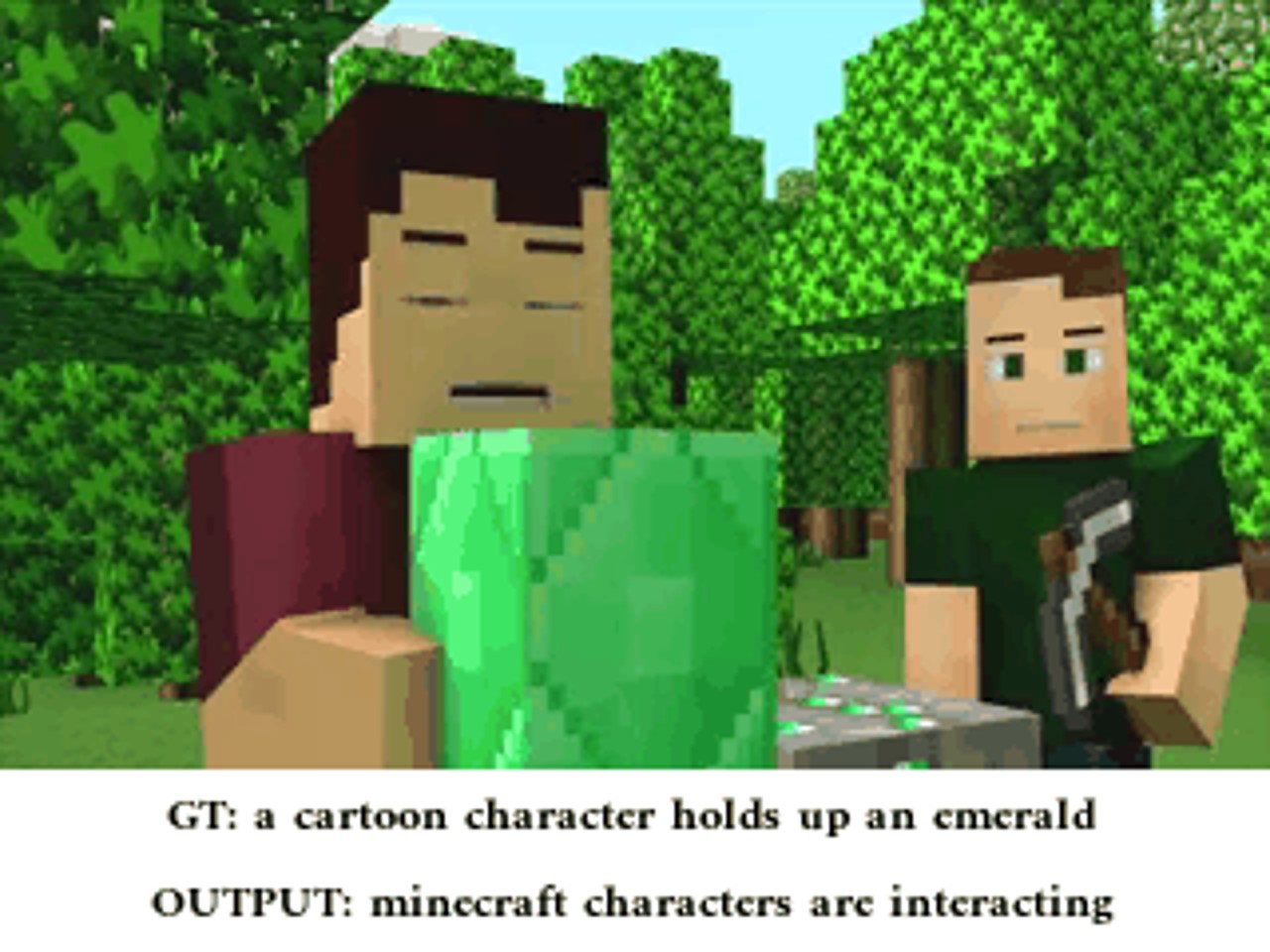}
    \includegraphics[width=0.31\linewidth]{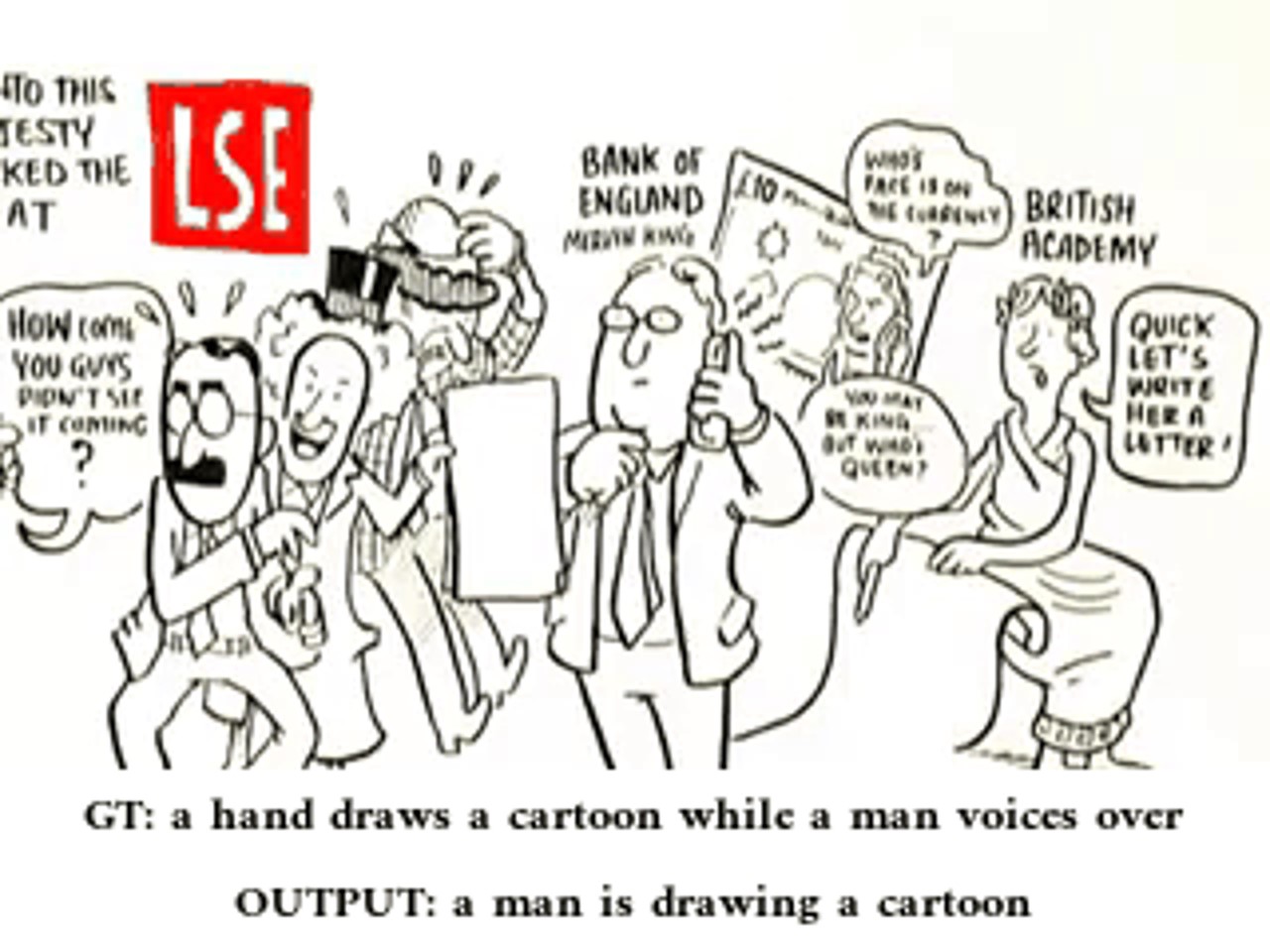}
    \includegraphics[width=0.31\linewidth]{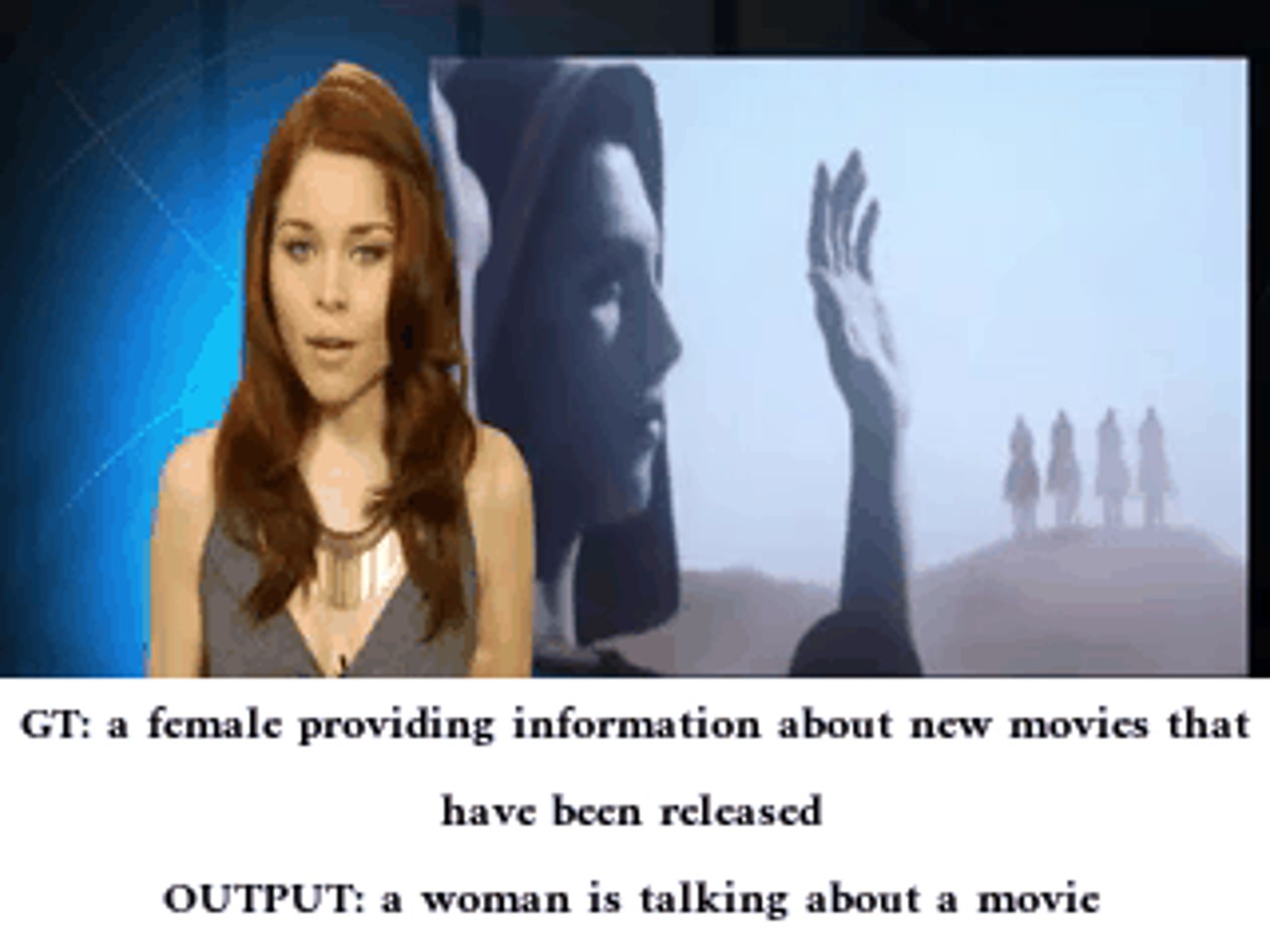}
    \\
    \includegraphics[width=0.31\linewidth]{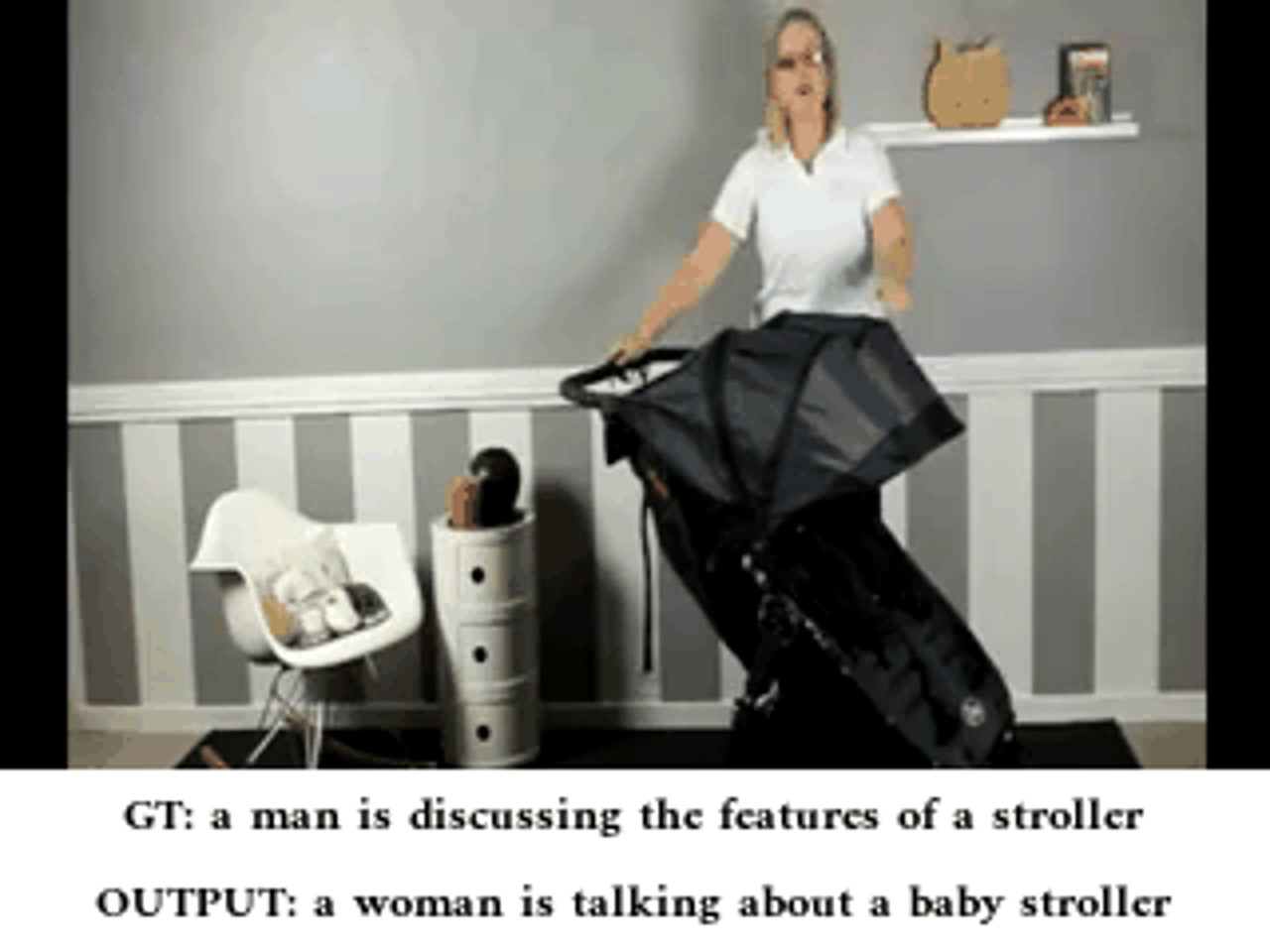}
    \includegraphics[width=0.31\linewidth]{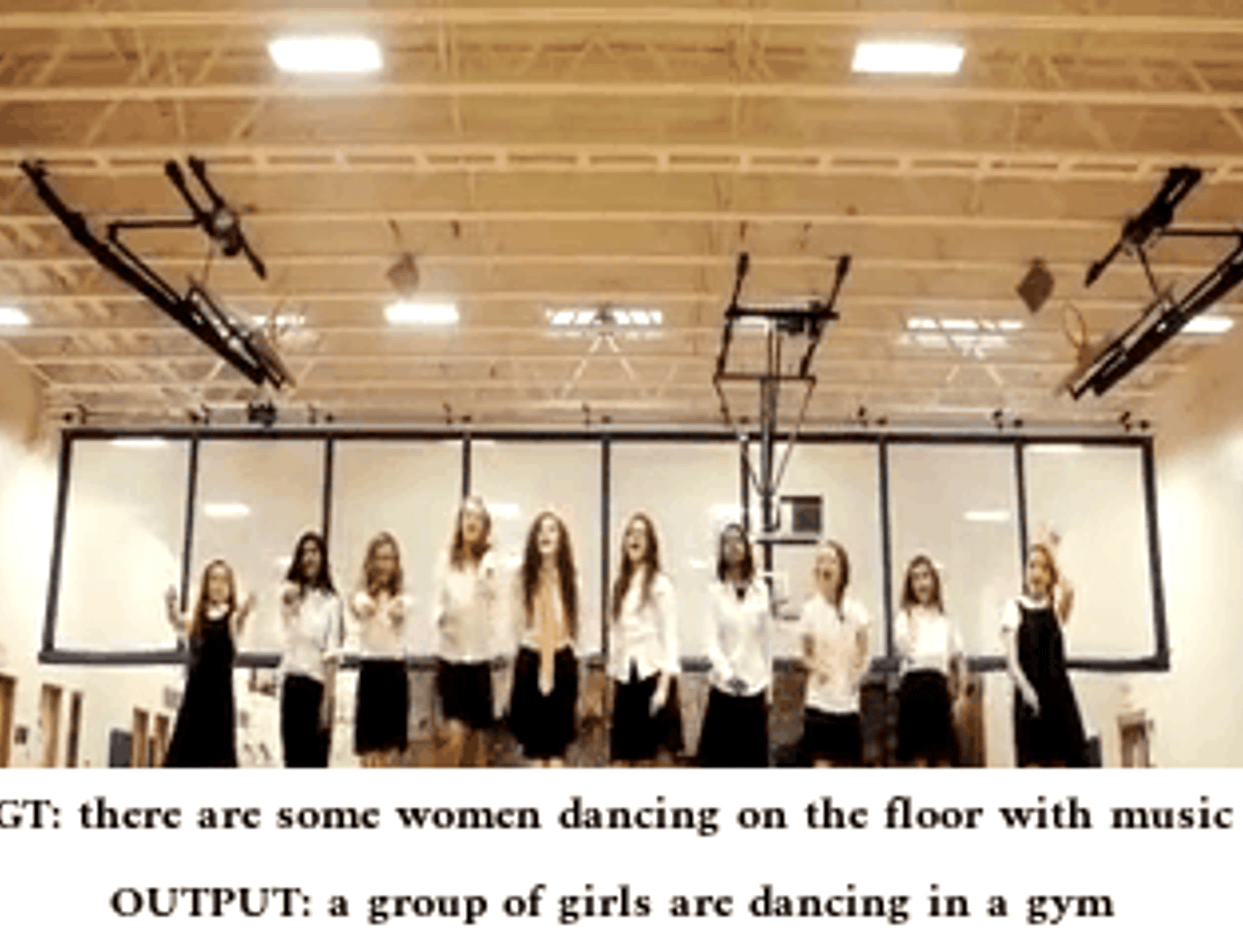}
    \includegraphics[width=0.31\linewidth]{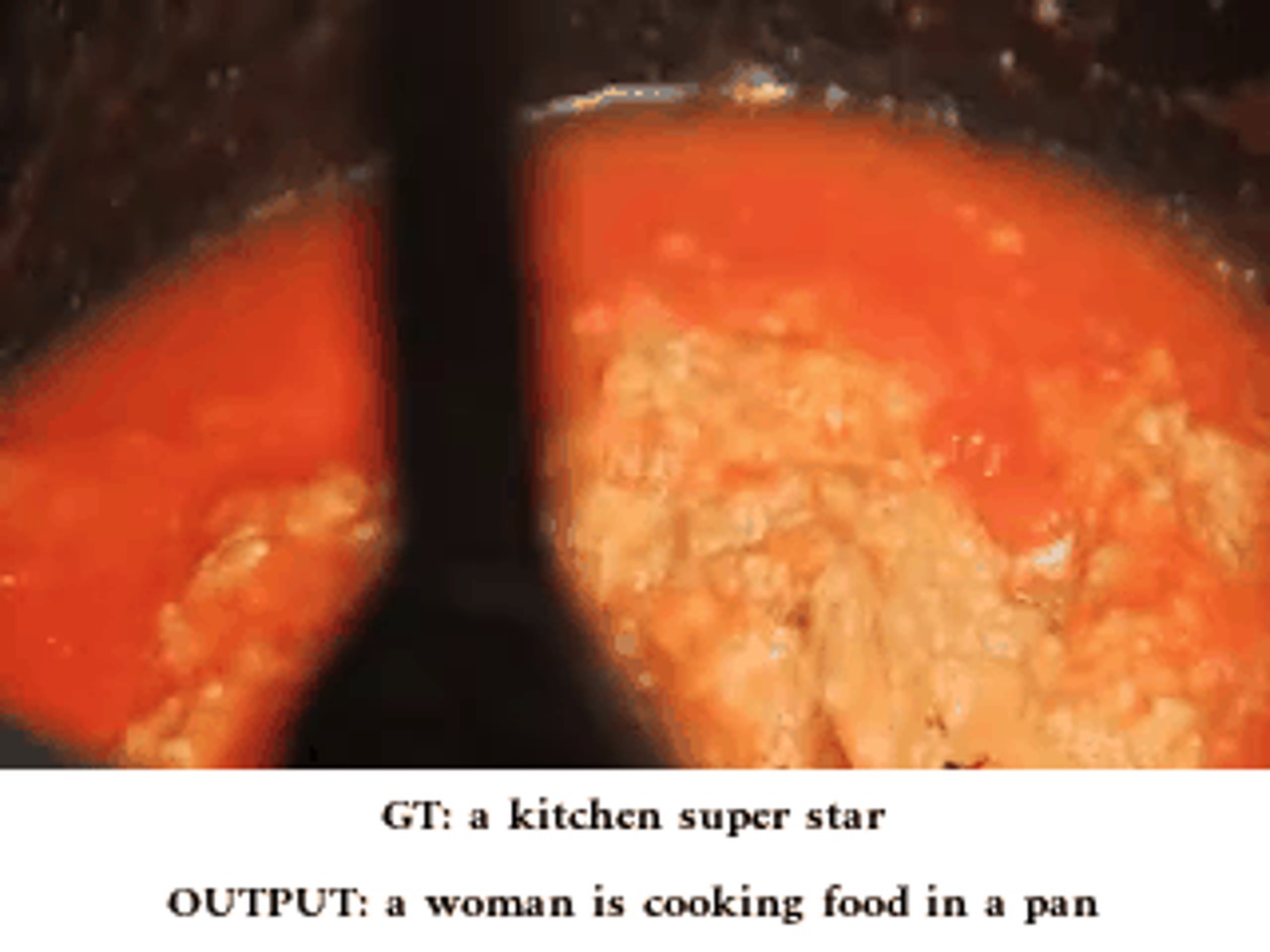}
    \\
      
    \caption{Based on METEOR and BLEU-4 we present a selection of the 10 percent best-performing videos in the  MSR-VTT data set. GT refers to the ground truth reference.}%
    \label{fig:best-out}
\end{figure*}

\begin{figure*}
    \centering

    \includegraphics[width=0.31\linewidth, height=3cm]{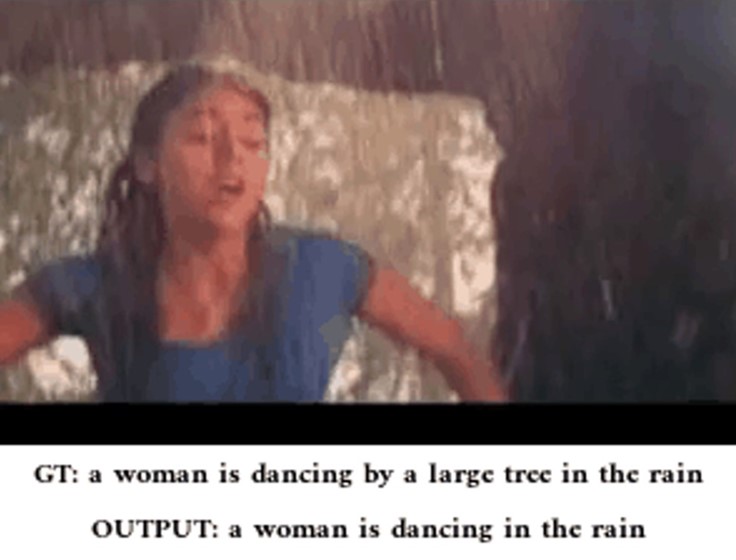}
    \includegraphics[width=0.31\linewidth, height=3cm]{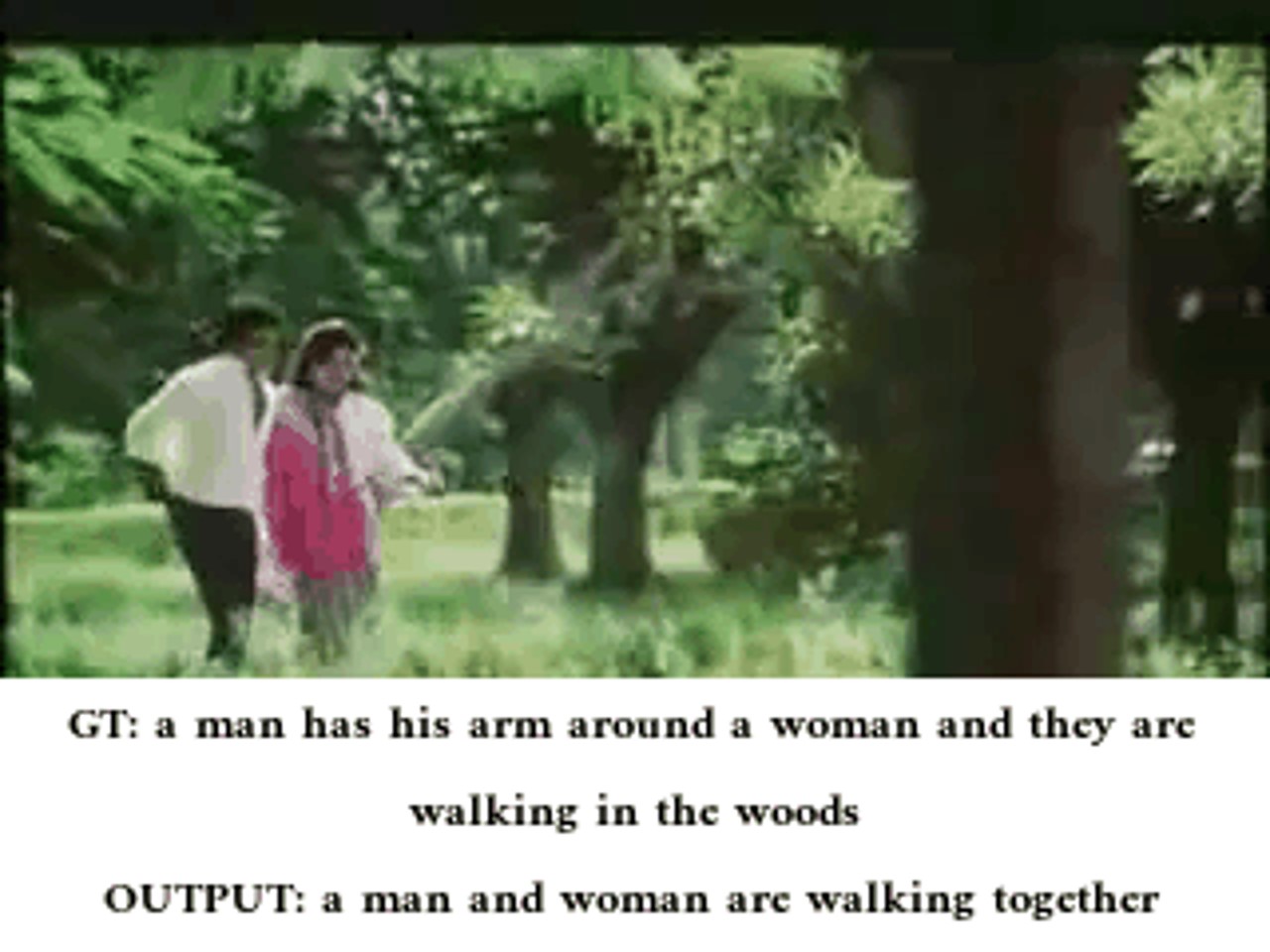}
    \includegraphics[width=0.31\linewidth, height=3cm]{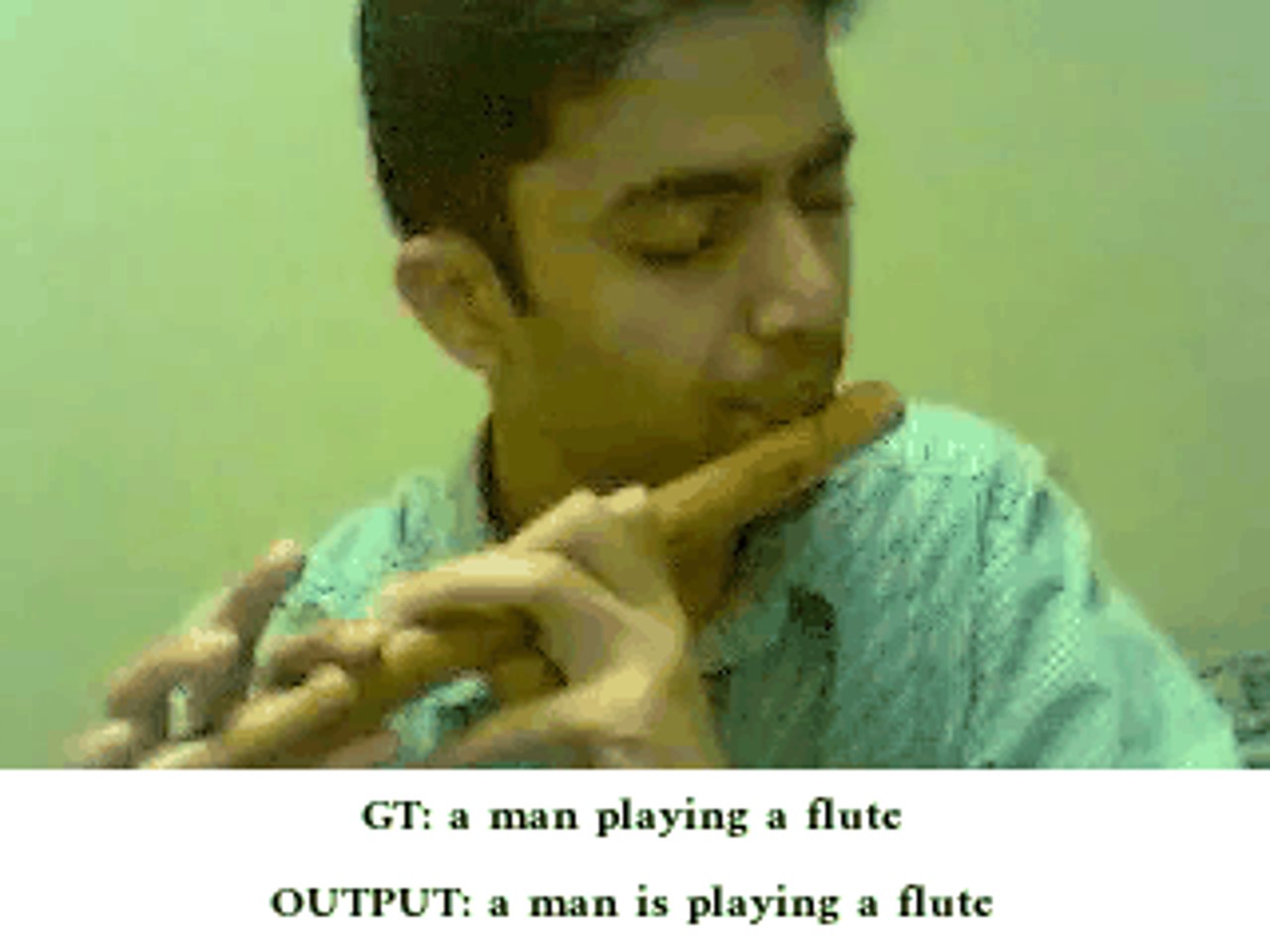}
    \\
    \includegraphics[width=0.31\linewidth, height=3cm]{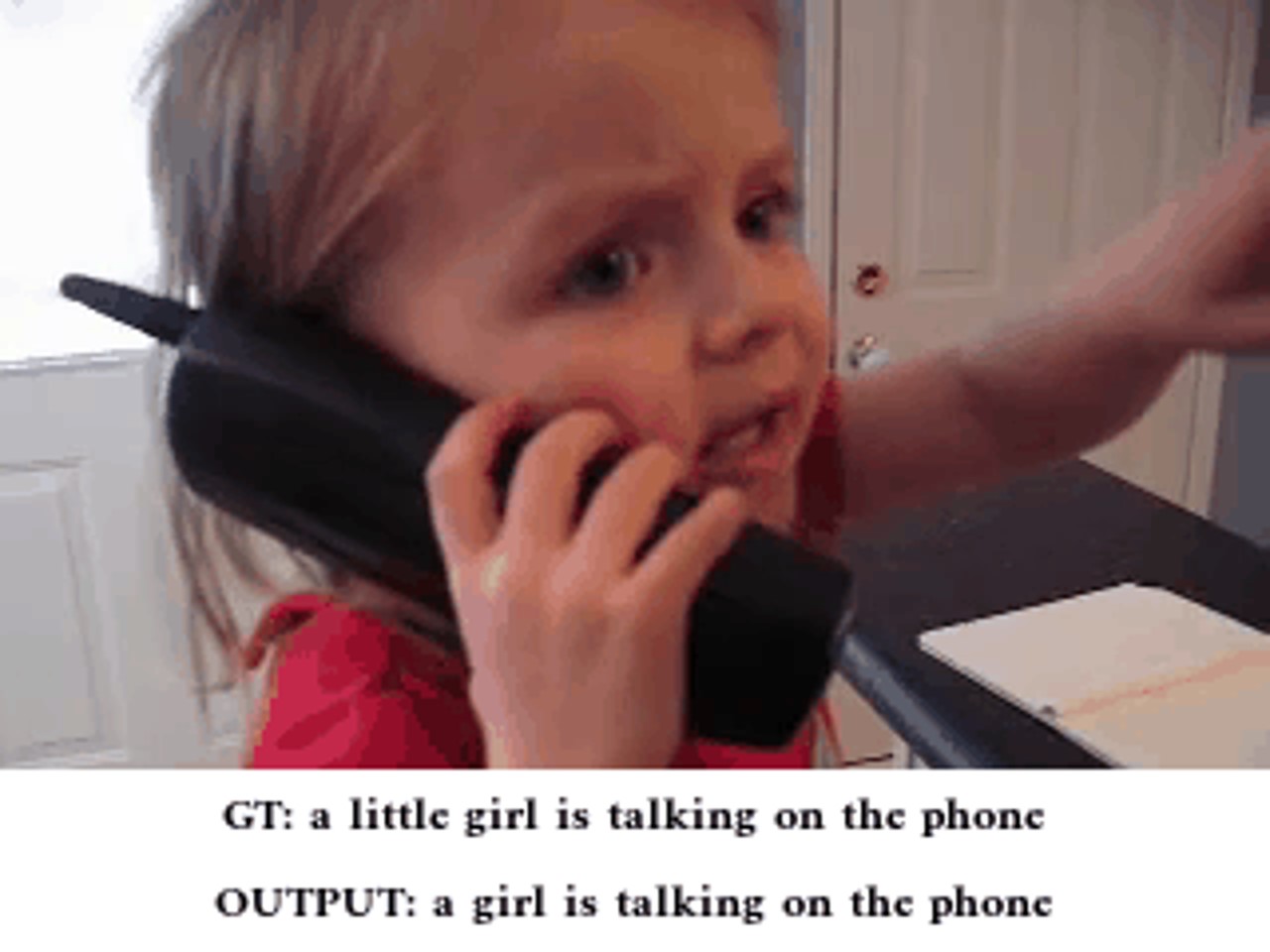}
    \includegraphics[width=0.31\linewidth, height=3cm]{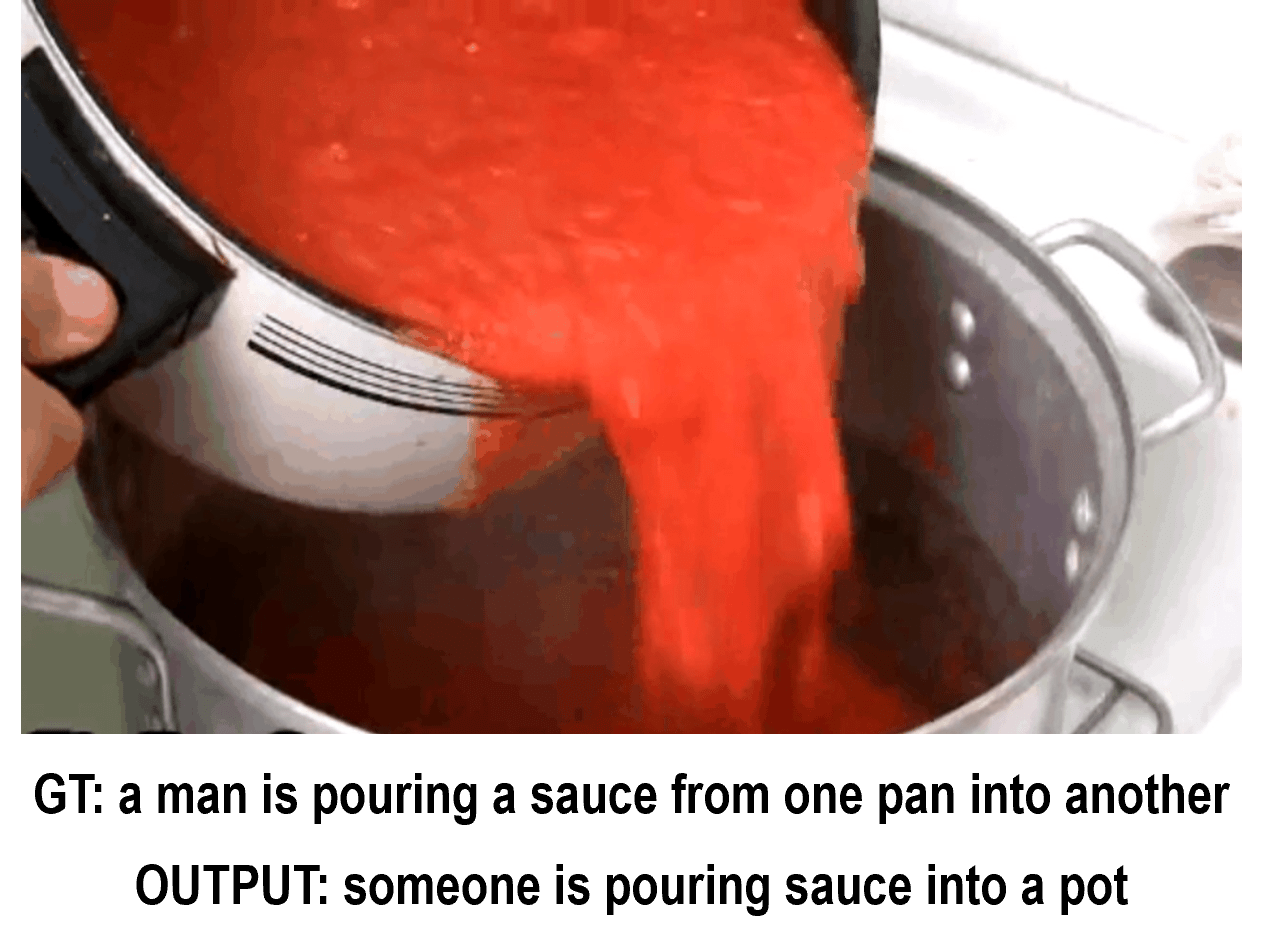}
    \includegraphics[width=0.31\linewidth, height=3cm]{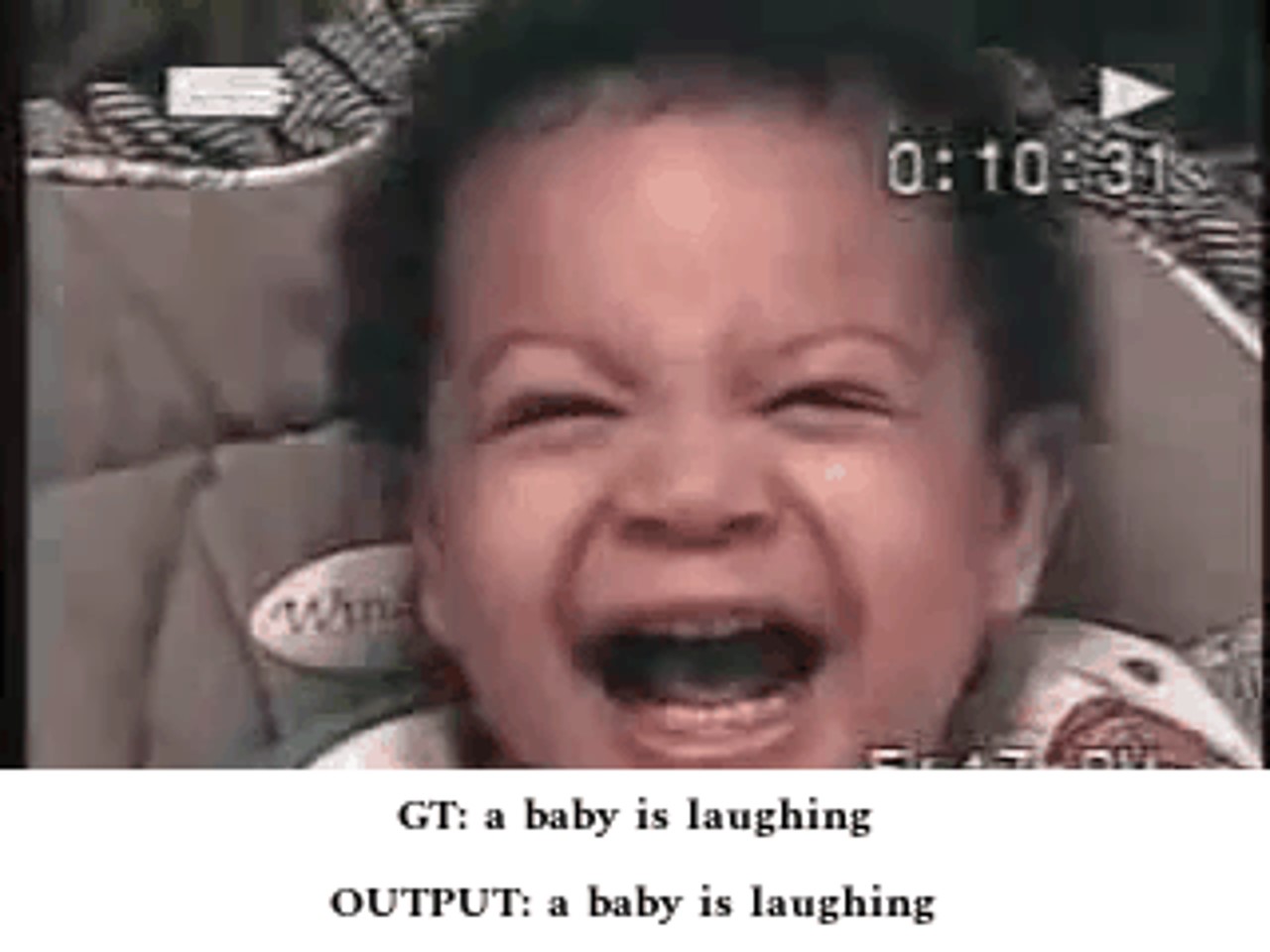}
    \\
    \includegraphics[width=0.31\linewidth, height=3cm]{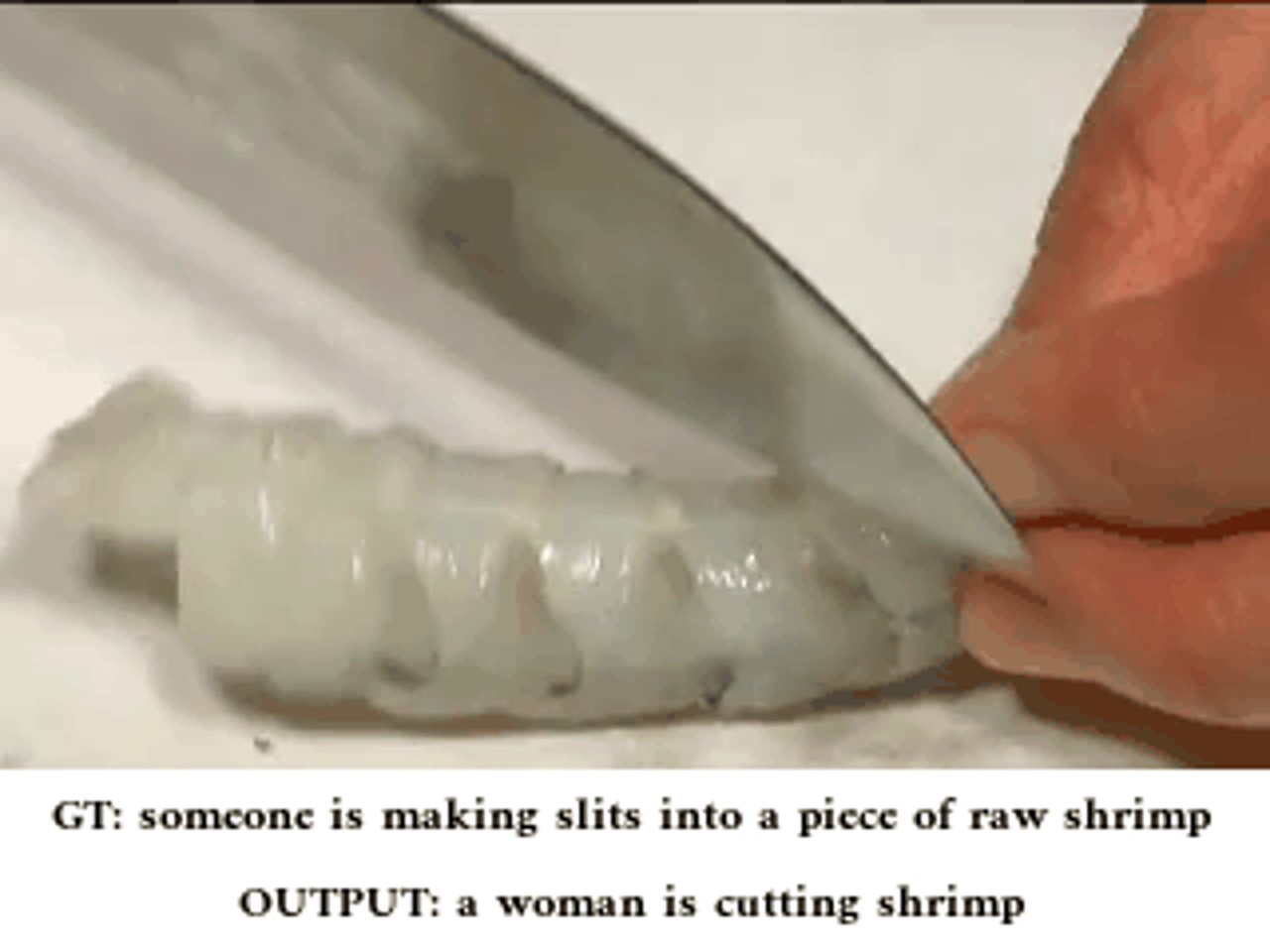}
    \includegraphics[width=0.31\linewidth,height=3cm]{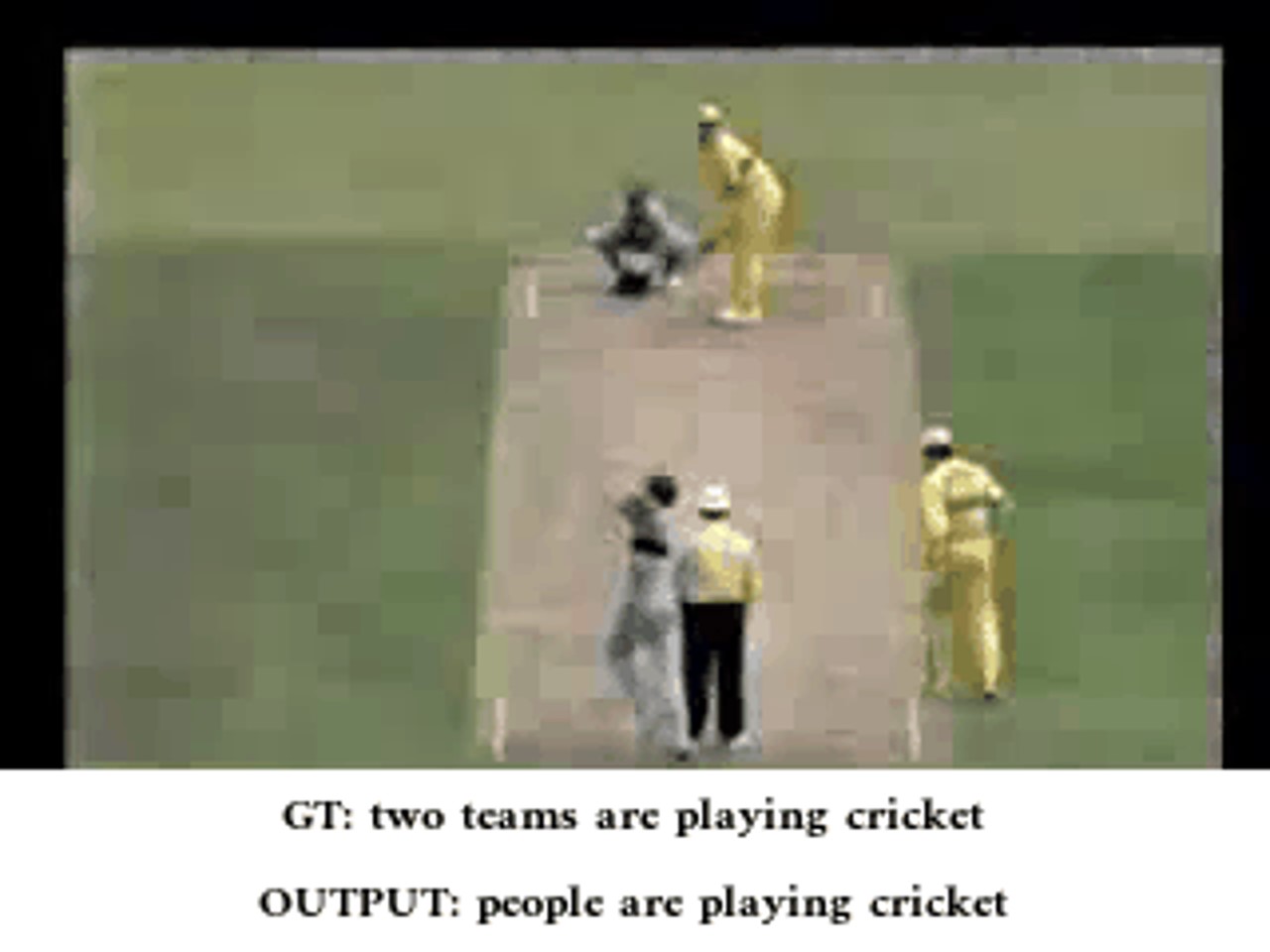}
    \includegraphics[width=0.31\linewidth,height=3cm]{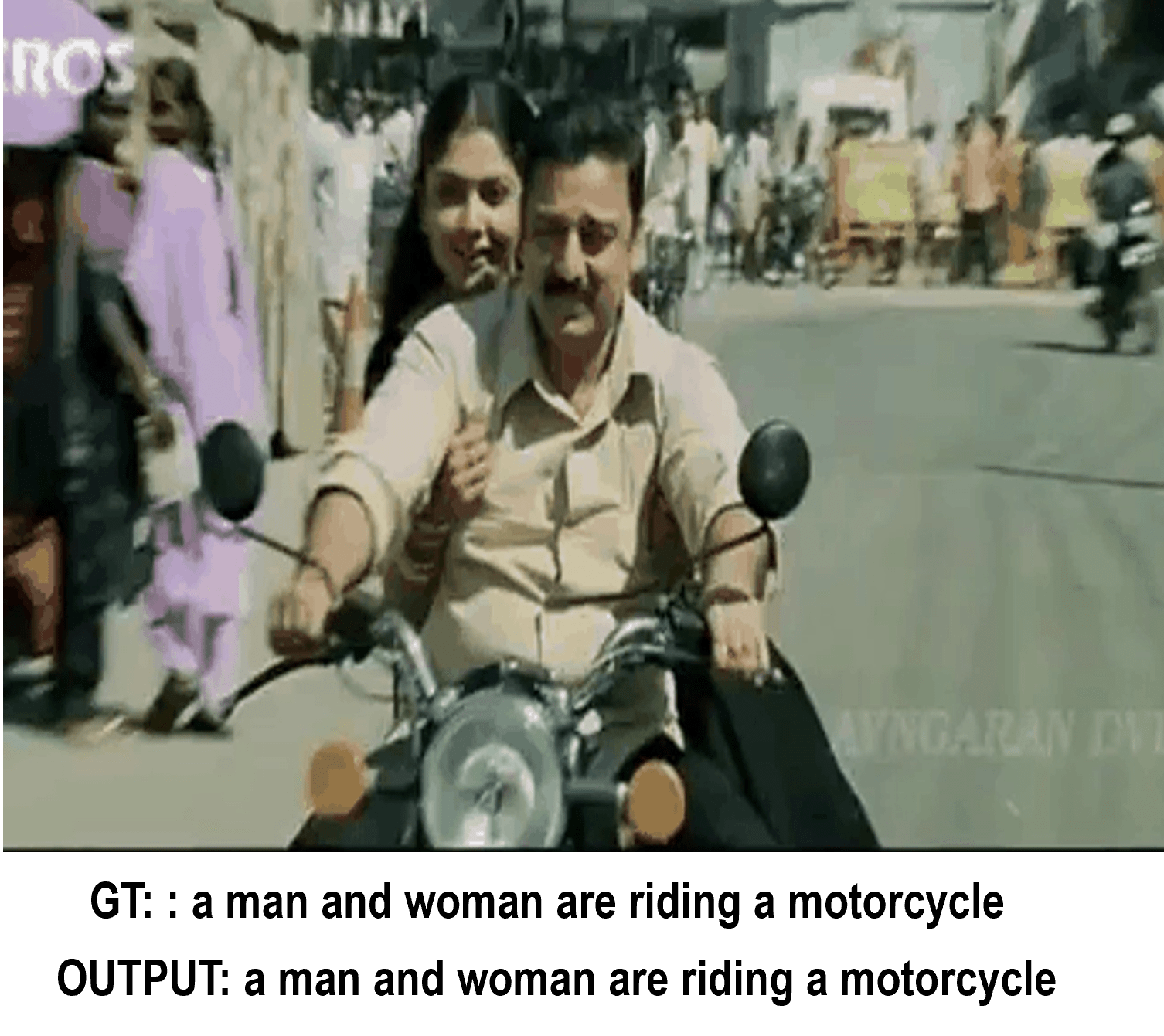}
    \\
    \includegraphics[width=0.31\linewidth,height=3cm]{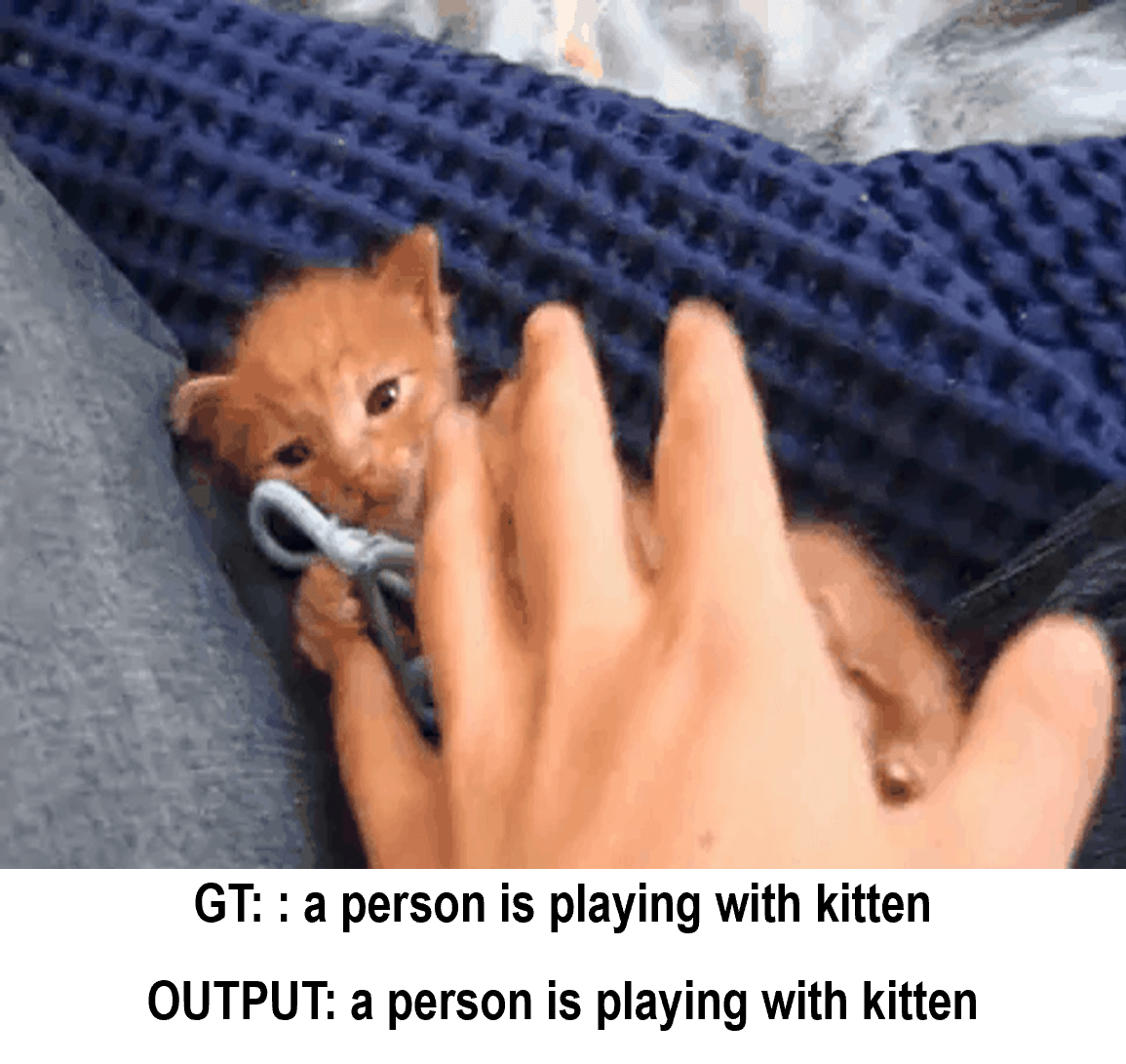}
    \includegraphics[width=0.31\linewidth,height=3cm]{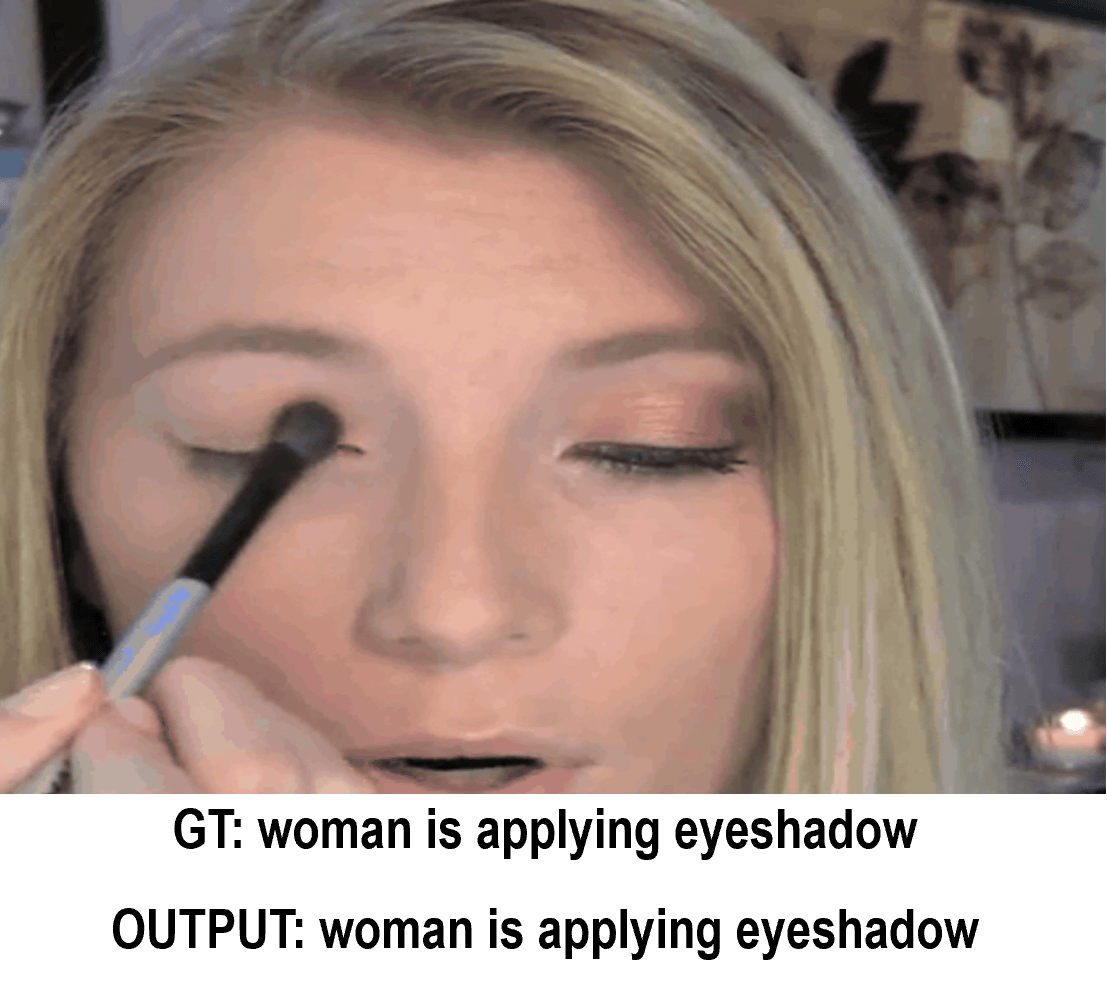}
    \includegraphics[width=0.31\linewidth,height=3cm]{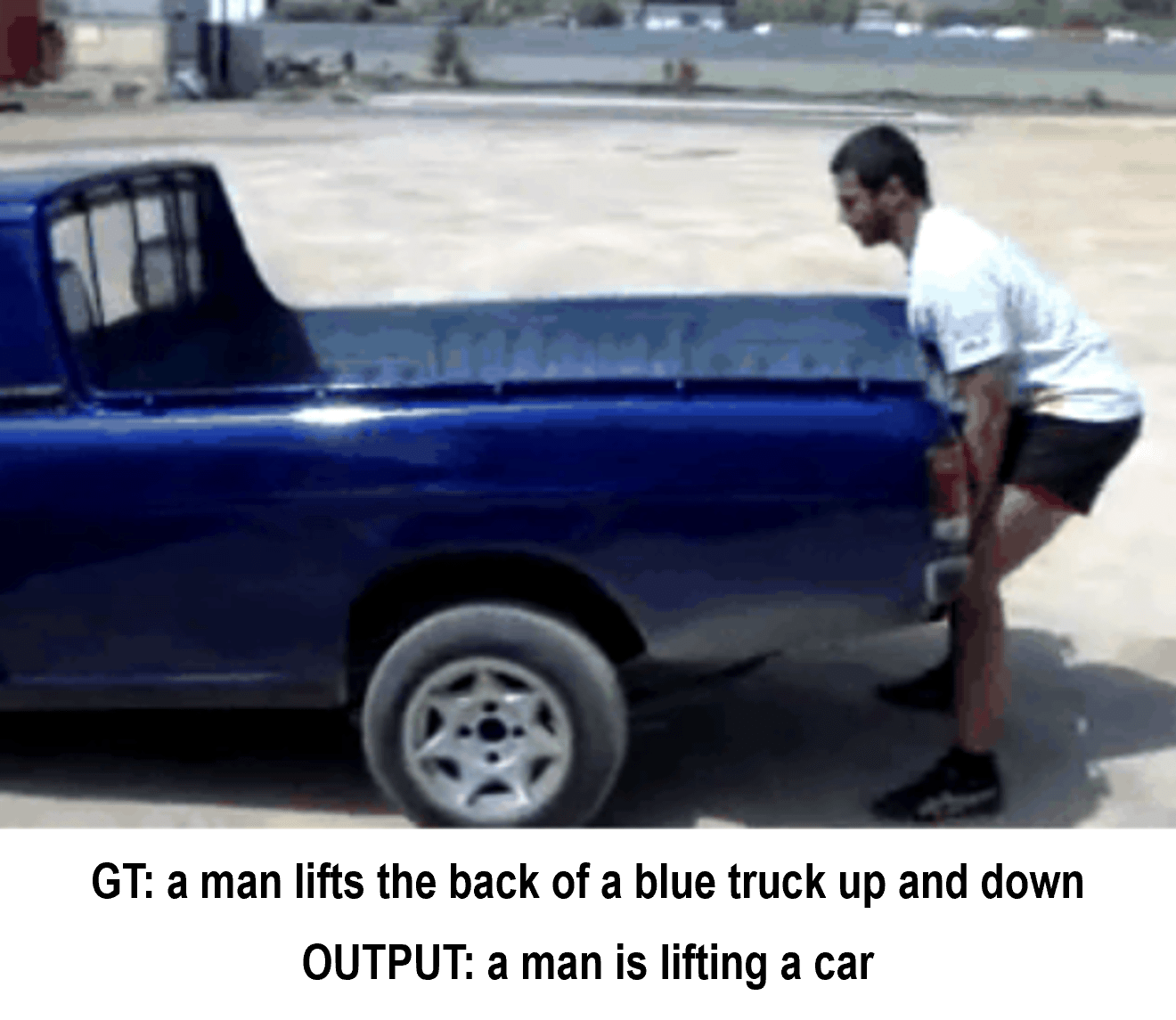}
    \\
      
    \caption{Based on METEOR and BLEU-4 we present a selection of the 10 percent best-performing videos in the  MSVD data set. GT refers to the ground truth reference.
}
    \label{fig:best-outmsvd}
\end{figure*}

Additional top-performing videos based on METEOR and BLUE-4 and their corresponding generated captions are demonstrated in~\Cref{fig:best-out} and~\Cref{fig:best-outmsvd}.

\begin{figure*}
    \centering

    {\centering 
     \small{\textbf{Video xxHx6s-DbUo-162-165}} \\
    \includegraphics[width=\linewidth]{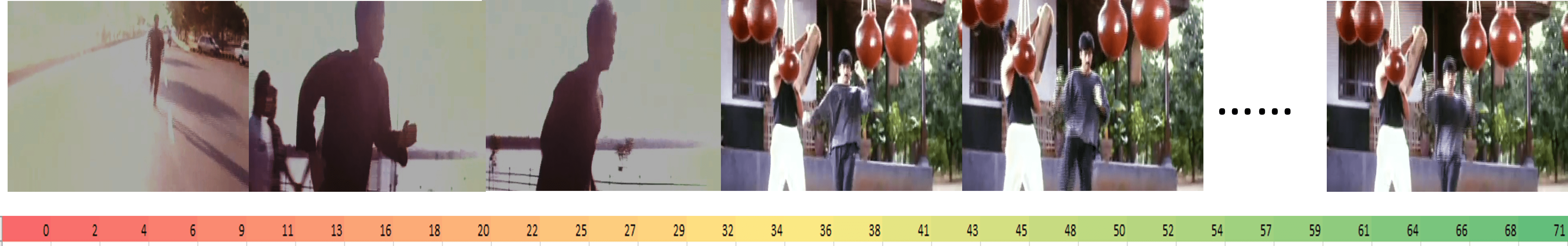} 
    }
    \small{ 
        \textbf{Uniform sampling}: a man is kicking a soccer ball CIDEr=5.41} \\[1.5ex]
        
    {\centering 
    \includegraphics[width=\linewidth]{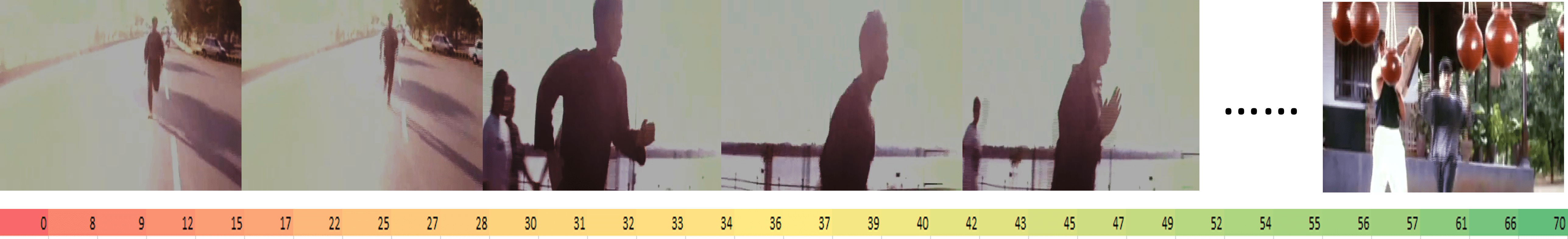} 
    }
    \small{ 
        \textbf{Adaptive sampling}: a man is running CIDEr=465.9\\
        \textbf{Reference}: a man is running on the road } \\[2ex]
        
    \vspace{1ex}

    {\centering 
    \small{\textbf{Video 7153}}\\
    \includegraphics[width=\linewidth]{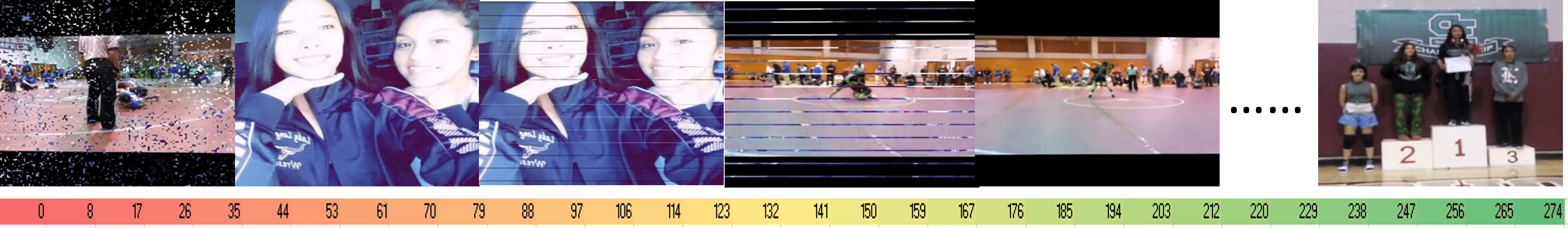} 
    }
    \small{ 
        \textbf{Uniform sampling}: a group of people are playing sports CIDEr=9.24} \\[1.5ex]
     {\centering 
    \includegraphics[width=\linewidth]{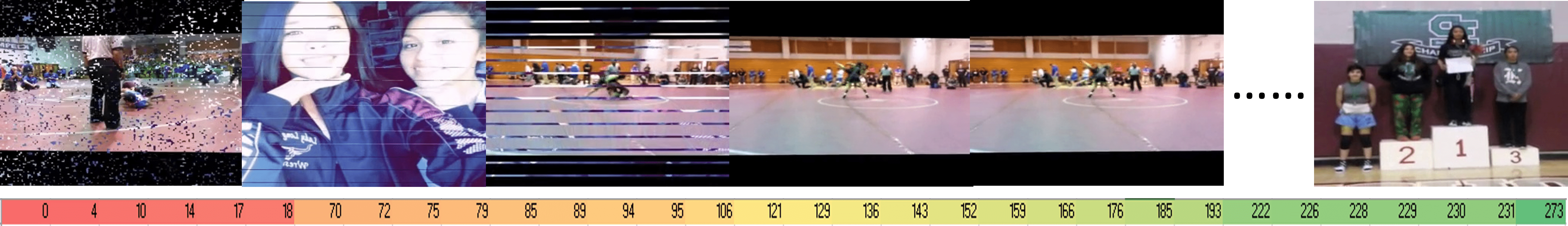}
    }
    \small{ 
        \textbf{Adaptive sampling}: a man is talking about a wrestling match CIDEr=78.88 \\
        \textbf{Reference}: two guys are wrestling in a competition \\
    }
    \caption{Effect of the Adaptive Frame Selection. In these example videos uniform sampling (top) wastes some of the 32 input frames for repetitive non-informative content. Our adaptive frame selection prefers those frames with strong differences to the previous one. Often, more diverse frames are selected helping generate better captions.}%
    \label{fig:AFSSupp}
\end{figure*}

\subsubsection{AFS Examples}
In addition, \Cref{fig:AFSSupp} shows representative frames in two videos of MSVD and MSR-VTT where the adaptive frame selection makes a major difference.
\textit{Video xxHx6s-DbUo-162-165} is about a man running on the road. Adaptive frame sampling is able to pick up the informative frames about the running movement. Thus, it generates a proper caption.
Also, in \textit{Video 7153} AFS selects frames, which contain the wrestling movement, that are key to correctly describing the \enquote{wrestling [\ldots] competition} instead of making a broad statement about \enquote{playing sports}.

\subsection{Caption Diversity}%
\begin{figure*}
    \centering
    {
    \hfill \includegraphics[width=0.38\linewidth]{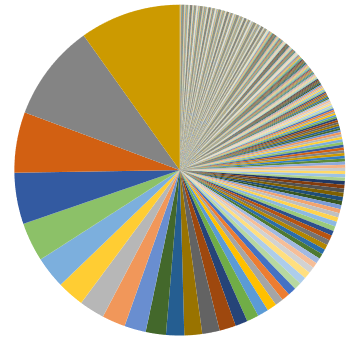} \hfill
    \hfill  \includegraphics[width=.38\linewidth]{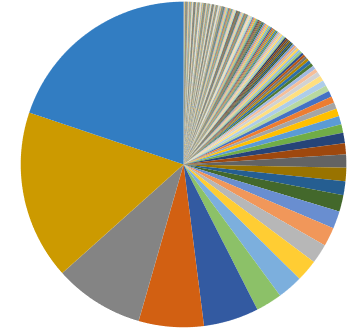} \hfill
    \\
    \parbox{0.55\linewidth}{\centering
    \small{\modelName{} (ours): }
    }
    \parbox{0.40\linewidth}{\centering
    \small{SemSynAN~\cite{perez2021improving}}
    }
    }

    \caption{Frequency of different POS structures on MSR-VTT Data set. More different segments indicate a higher diversity in captions. Note, that SemSynAN only uses 4 different POS structures for more than 50\% of its generated captions.}%
    \label{fig:pos_msr}
\end{figure*}

\begin{figure*}
    {
        \hfill \includegraphics[width=0.33\linewidth]{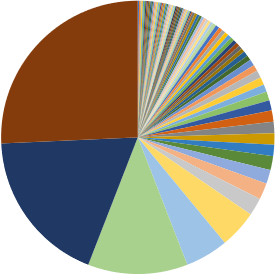} \hfill
        \hfill \includegraphics[width=.33\linewidth]{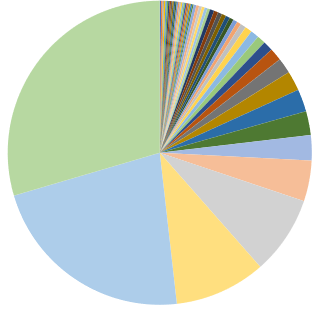} \hfill
        \\
        \parbox{0.60\linewidth}{\centering
        \small{\modelName{} (ours)}
        }
        \parbox{0.40\linewidth}{\centering
        \small{SemSynAN~\cite{perez2021improving}}
        }
    }
      
    \caption{Frequency of POS structures on MSVD Data set. More different segments indicate a higher diversity in captions. Note, that SemSynAN only uses 6 different POS structures for more than 75\% of its generated captions. }%
    \label{fig:pos_msvd}
\end{figure*}

\label{sec:diversity}
Following the discussion on caption diversity in Section 4.4, we visualize the distribution of generated Part-of-Speech tagging (POS) structures for the captions by our model and by SemSynAN~\cite{perez2021improving} in \Cref{fig:pos_msr} for the MSR-VTT and \Cref{fig:pos_msvd} for the MSVD data set.
While some sentence structures are more frequent than others it can be seen that our approach generates more diverse sentences.

As explained in Section 4.2 of our paper some related works operate on a limited subset of the data sets. As we show in \Cref{tab:vocab-sizes} working on a subset limits the diversity of the sentences and the size of the vocabulary.
Our vocabulary on MSVD is about twelve-thousand words. In contrast, the vocabulary in SemSynAN~\cite{perez2021improving} has only half the size, which will also restrict the diversity of the captions generated by the SemSynAN model.

\begin{table}
    \centering{
    \begin{tabular}{c c c}
    \toprule
        Data set & \modelName{} (Ours) & SemSynAN~\cite{perez2021improving} \\
        \midrule
        MSVD    & 9636           & $\sim$6K\\
        MSR-VTT & 23081          & $\sim$12K \\
        \bottomrule
    \end{tabular}
    }
    \caption{Vocabulary sizes on MSVD and MSR-VTT for our model and SemSynAN~\cite{perez2021improving} showing that our model operates on a much larger set of words.}%
    \label{tab:vocab-sizes}
\end{table}


\subsection{Training Details}%
\label{tab:training}

We initialize our encoder with the weights of a Swin-B network weight with layer number $= \{ 2,2,18,2 \}$~\cite{liu2021swin}, which was trained on the Kinetics400 data set~\cite{kay2017kinetics}.
For all videos, we employ AFS to select $N=32$ frames and follow the default Swin-B normalization, cropping and resizing operations to transform them to a dimension of $3 \times 224 \times 224$. The input to the Swin network is $T \times W \times H \times 3$ (in our case $(32,224,224,3)$), which is divided to $3 \times \frac{T}{2} \times \frac{W}{4} \times \frac{H}{4}$ 3D tokens. These 3D tokens are hierarchically merged to analyse their temporal relations.
The output of our encoder is the last hidden layer of the Swin-B architecture variant. It has the size of $\text{Size}=(\frac{T}{2},\frac{W}{32},\frac{H}{32},1024)$. After average 2D-pooling on $(W,H)$, followed by a single linear layer, we produce sixteen output tokens of size $768$, encoding the visual state to be fed to the decoder. The same output is also taken to produce a semantic context vector.

Our decoder backbone is initialized by pre-trained BERT~\cite{devlin2018bert} on which we add a language modelling head.
During training and inference we employ causal masks to prevent attention to future tokens.
Following prior work~\cite{chen2021motion,chen2018tvt,zhang2020object}, sentences longer than 20 words are truncated. All words are converted to lower case and the decoder predicts lower case tokens. Its dropout probability is set to 0.3.
A two-layer MLP with RELU activation is trained for $K=768$ concepts (layer sizes $(1024, 2048, 768)$).
In a first step, we train the semantic concept MLP network using the pre-trained Swin-B and BCE loss.

We train our model end-to-end network to minimize the overall loss $\mathcal{L}$ where the training loss for decoder is the cross-entropy to the randomly selected training caption and the loss for the full network including the semantic vector MLP is given by 
\begin{equation}
\mathcal{L}=\mathcal{L}_{CE} + \lambda \cdot \mathcal{L}_{BCE}, \;\;\text{with} \;\; \lambda=0.1.
\end{equation}

To train our models, we use the AdamW~\cite{loshchilov2017decoupled} optimizer with default settings, with a learning rate of 0.00001 and an effective batch size of 8. Additionally, we use gradient clipping of 0.05.
The final model is selected based on the best harmonic mean of the METEOR~\cite{banerjee2005meteor} and BLEU-4~\cite{papineni2002bleu} scores to avoid optimizing for a single metric.

\end{document}